\documentclass{article}

\usepackage{arxiv}

\usepackage{rotating}
\usepackage{listings}
\usepackage{pdflscape}
\usepackage{longtable}
\usepackage{lscape}
\usepackage{multicol}
\usepackage{algpseudocode}
\usepackage{subcaption}
\usepackage{url}
\usepackage{changepage}
\usepackage{upgreek}
\usepackage{xcolor}
\usepackage{array}
\usepackage{makecell}
\usepackage{siunitx}
\usepackage{xr-hyper}
\usepackage{amsmath,amssymb,amsfonts}
\usepackage[numbers, sort&compress]{natbib}
\usepackage{textcomp}
\usepackage{wrapfig}
\usepackage[super]{nth}
\usepackage{caption}
\usepackage{algorithm}
\usepackage{setspace}
\usepackage[utf8]{inputenc}
\usepackage{fourier}
\usepackage{soul}
\usepackage{float}
\usepackage{ragged2e}
\usepackage[T1]{fontenc}
\usepackage{xurl}
\usepackage{hyperref}      
\usepackage{booktabs}    
\usepackage{nicefrac}     
\usepackage{microtype}     
\usepackage{lipsum}
\usepackage{fancyhdr}    
\usepackage{graphicx}
\usepackage{tabularx}
\usepackage{multirow}

\newlength{\fulllength}
\title{Large Language Models and 3D Vision for Intelligent Robotic Perception and Autonomy
\thanks{\textit{\underline{Citation}}: 
\textbf{Mehta V, Sharma C, Thiyagarajan K. Large Language Models and 3D Vision for Intelligent Robotic Perception and Autonomy. Sensors. 2025; 25(20):6394. https://doi.org/10.3390/s25206394.}} 
}

\author{
  Vinit Mehta \\
  Machine Learning Lab \\
  IIIT Hyderabad \\
  Hyderabad, India \\
  \texttt{vinit.mehta@research.iiit.ac.in} \\
    \And
  Charu Sharma \\
  Machine Learning Lab \\
  IIIT Hyderabad \\
  Hyderabad, India \\
  \texttt{charu.sharma@iiit.ac.in} \\
   \And
  Karthick Thiyagarajan \\
  SensR Lab \\
  Western Sydney University \\
  Penrith, Australia \\
  \texttt{K.Thiyagarajan@westernsydney.edu.au} \\
}

\begin{document}
\maketitle

\begin{abstract}
With the rapid advancement of artificial intelligence and robotics, the integration of Large Language Models (LLMs) with 3D vision is emerging as a transformative approach to enhancing robotic sensing technologies. This convergence enables machines to perceive, reason and interact with complex environments through natural language and spatial understanding, bridging the gap between linguistic intelligence and spatial perception. This review provides a comprehensive analysis of state-of-the-art methodologies, applications and challenges at the intersection of LLMs and 3D vision, with a focus on next-generation robotic sensing technologies. We first introduce the foundational principles of LLMs and 3D data representations, followed by an in-depth examination of 3D sensing technologies critical for robotics. The review then explores key advancements in scene understanding, text-to-3D generation, object grounding and embodied agents, highlighting cutting-edge techniques such as zero-shot 3D segmentation, dynamic scene synthesis and language-guided manipulation. Furthermore, we discuss multimodal LLMs that integrate 3D data with touch, auditory and thermal inputs, enhancing environmental comprehension and robotic decision-making. To support future research, we catalog benchmark datasets and evaluation metrics tailored for 3D-language and vision tasks. Finally, we identify key challenges and future research directions, including adaptive model architectures, enhanced cross-modal alignment and real-time processing capabilities, which pave the way for more intelligent, context-aware and autonomous robotic sensing systems.
\end{abstract}

\section{Introduction}\label{sec:introduction}

Robot sensing is at the forefront of a future where humanity coexists with highly advanced, human-like robots capable of independent reasoning, natural interaction and deep environmental understanding. These next-generation robots will seamlessly handle tasks ranging from household chores to managing companies, learning and adapting like human children, but starting with the vast wealth of accumulated human knowledge. The integration of Large Language Models (LLMs) and 3D vision enables robots to perceive their surroundings in three dimensions, comprehend complex instructions and interact naturally with humans and objects. This synergy is driving advancements in robotic manipulation, autonomous navigation and environmental interaction, bringing us closer to a future where robots are intelligent, integral partners in our daily lives.

\begin{figure}
    \centering
    \includegraphics[width=0.99\textwidth]{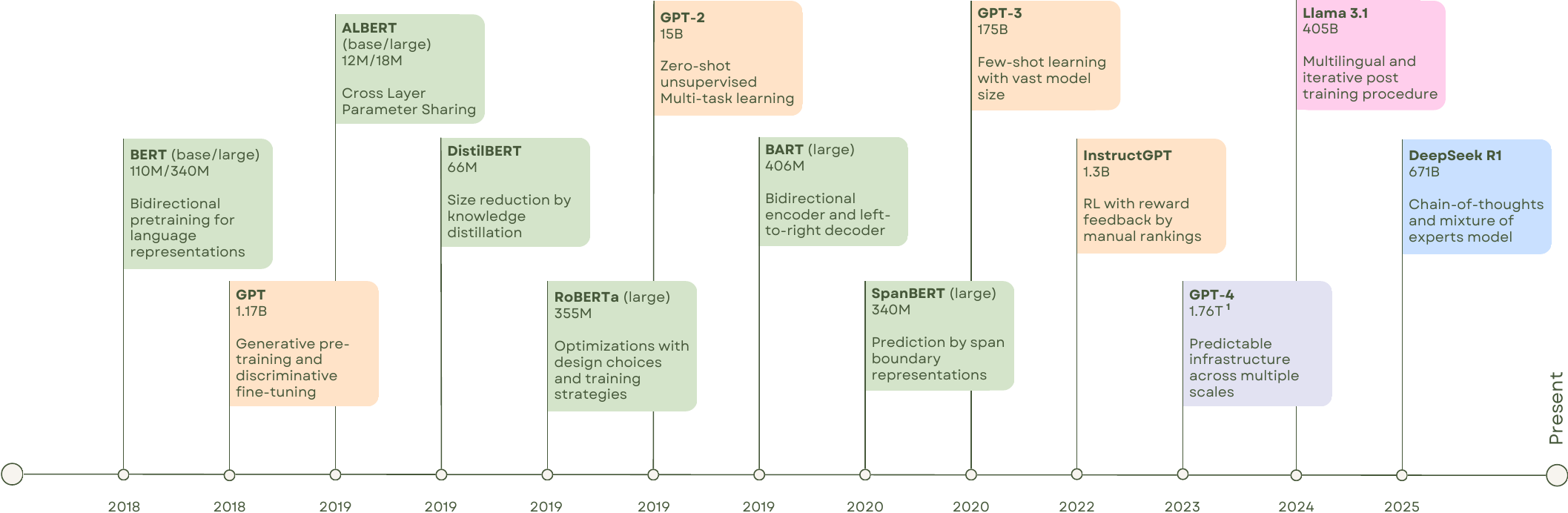}
    \caption{Major LLM Development: Various different LLM models (BERT \citep{devlin2018bert}, \mbox{ALBERT \citep{lan2019albert},} \mbox{DistilBERT \citep{sanh2019distilbert},} RoBERTa \citep{liu2019roberta}, SpanBERT \citep{joshi2020spanbert}, GPT \citep{Radford2018ImprovingLU}, GPT-2 \citep{Radford2019LanguageMA}, GPT-3 \citep{brown2020language}, GPT-4 \citep{achiam2023gpt}, \mbox{Llama 3.1 \citep{Llama3.1},} InstructGPT \citep{InstructGPT}, BART \citep{BART}, DeepSeek R1 \citep{DeepSeekR1}), along with their parameter count and key feature of improvement. \textsuperscript{1} GPT-4 parameters count source: \citep{GPT4Wikipedia}.} 
    \label{fig:llm_timeline}
\end{figure}

In recent years, LLMs such as ChatGPT \citep{achiam2023gpt}, LLaMA \citep{touvron2023llama}, Vicuna \citep{vicuna2023} MiniGPT \citep{zhu2023minigpt} and DeepSeek R1 \citep{DeepSeekR1} have emerged as transformative tools in natural language processing (NLP), see Figure  \ref{fig:llm_timeline}. These models exhibit exceptional capabilities in language understanding, text generation and translation. LLMs follow a two-phase process: pre-training and fine-tuning. During pre-training, models are exposed to vast and diverse text corpora, enabling them to learn intricate language patterns. Fine-tuning then adapts this pre-trained knowledge to specific tasks or domains, enhancing task-specific performance.

The integration of LLMs into 3D vision tasks has led to significant advancements in fields such as scene understanding, visual question answering (VQA) and robotic manipulation. Traditionally, methods in these domains have relied heavily on 2D visual representations, as seen in foundational works \citep{NovelTrailDetection, DDRInSmartphone, AutonomousDrivingWithSceneUnderstanding, FastPanopticSegmentation}. The advent of LLMs, including models like GPT and its variants, has introduced unprecedented improvements in common-sense reasoning and contextual understanding. However, applying these models to 3D data presents unique challenges and opportunities.

While recent efforts have aligned images with LLMs through Vision-Language Models (VLMs), such as  \citep{FOREWARN, CLIP, DeepSeekVLM, ALIGN, ViLT}, grounding LLMs in 3D data remains a relatively unexplored area. This is a critical limitation, as 3D data inherently provides richer spatial and contextual information compared to 2D images. Despite rapid progress in multimodal LLMs \citep{EMLMs, zhang2024mm, yin2023survey, Kamath2024, wang2024exploring}, the scarcity of comprehensive 3D datasets and the complexities of aligning dense 3D visual data with textual embeddings pose significant challenges.

This review article aims to provide a comprehensive overview of the current state-of-the-art (SOTA) research at the intersection of 3D vision and LLMs. We examine recent breakthroughs in LLMs that excel at tasks requiring common-sense reasoning and contextual comprehension, alongside the challenges of encoding and integrating 3D data within these models. Additionally, we explore applications such as text-to-3D generation, 3D scene understanding, VQA and autonomous driving.

Recent research on the integration of LLMs into 3D vision can be broadly categorized into three main methodologies: direct embedding, 2D-to-3D mapping and pre-alignment. These approaches provide foundational insights into this emerging field.

The remainder of this article is organized as follows. Section \ref{sec:overview_of_llms} presents an overview of LLMs, their architectures and recent advancements. This is followed by Section \ref{sec:fundamentals_of_3d_vision}, which introduces the foundational concepts of 3D vision, covering various types of 3D data and key sensing technologies and Section \ref{sec:motivation} presents the motivation behind combining LLMs and 3D vision. Section \ref{sec:sensing_technologies} explores 3D vision and its applications in robotic sensing technologies. In Section \ref{sec:applications}, we review key advancements and applications of LLMs integrated with 3D vision. We begin by discussing localization and grounding (Section \ref{sec:localization_and_grounding}), a foundational capability for robots to interpret their surroundings and connect linguistic descriptions to physical entities. Subsequently, Section \ref{sec:dynamic_scenes} explores the understanding of dynamic scenes, including critical tasks such as human action recognition, essential for robots operating in human-centric environments. The distinct challenges and methodologies pertinent to indoor and outdoor scene understanding are detailed in Section \ref{sec:indoor_and_outdoor_scene_understanding}. A significant stride towards more adaptable systems, open vocabulary understanding and pretraining, which enables generalization to unseen objects and scenarios, is covered in Section \ref{sec:open_vocabulary_understanding_and_pretraining}. Section \ref{sec:text_to_3d} delves into techniques for generating detailed and customizable 3D environments from textual inputs, crucial for training and testing robotic systems. Furthermore, the importance of integrating diverse multimodal inputs, such as touch, audio and thermal data, to achieve a more holistic and robust scene understanding is emphasized in Section \ref{sec:multimodality}. Finally, Section \ref{sec:embodied_agent} explores the advancements in embodied agents, intelligent systems that perceive, interact with and act upon their 3D environments by leveraging these integrated capabilities. Sections \ref{sec:datasets} and \ref{sec:evaluation_metrics} provide a comprehensive review of existing datasets and evaluation metrics for training and assessing models. Finally, Section \ref{sec:challenges} presents an analysis of current research limitations and future directions in 3D vision and LLM-based robotic sensing technologies, leading to the conclusions outlined in Section \ref{sec:conclusion}.

\section{Background}\label{sec:background}

This section provides an overview of the foundational technologies of LLMs and 3D vision, examining their core principles, key advancements and applications. We start by defining the basic terminologies and concepts in these fields, illustrating how LLMs have revolutionized natural language understanding and generation, while 3D vision has empowered machines to perceive and interpret three-dimensional environments. Finally, we discuss the rationale for integrating these two domains, highlighting their potential to tackle complex challenges and enable novel applications at their intersection for next-generation multimodal robotic sensing technologies.

\subsection{Overview of LLMs}\label{sec:overview_of_llms}

LLMs have had a transformative impact on the AI community and beyond. Open-source LLMs, in particular, have driven significant advancements, making a substantial contribution to the field's development. These models offer a promising approach for encoding and projecting human knowledge and experiences into a latent space.

Initially developed for text-based translation tasks, LLMs have now found applications far beyond their original scope. These models exhibit exceptional capabilities in tackling complex tasks that were once the exclusive domain of specialized algorithms or human expertise. Their ability to reason from ambiguous human language and formulate precise responses has led to extensive usage in embodied control and scene understanding via images and other multimodal inputs \citep{DoAsICanNotAsISay, InnerMonologue, SocraticModels}.

Typically trained on extensive textual data, LLMs constitute foundation models, also known as base models. They employ self-supervised objectives such as next-token \mbox{prediction \citep{brown2020language, Radford2019LanguageMA}} or masked token reconstruction \citep{devlin2018bert, ExploringTheLimitsOfTransferLearning}. Built on deep learning architectures, with transformers playing a central role, pre-training forms the foundation of LLM development. This process enables them to learn intricate language patterns and acquire broad knowledge encompassing grammar, syntax, semantics and general \mbox{world understanding.}

According to \citep{ExploringTheLimitsOfTransferLearning}, an effective LLM should possess four main qualities: understanding the meaning of words and context in natural language, generating text that sounds human-like, comprehending specialized contexts and following instructions accurately.

Fine-tuning further tailors these models for specific applications such as summarization, translation or question answering. LLMs like GPT-4 \citep{achiam2023gpt}, LLaMA \citep{touvron2023llama} and MiniGPT \citep{zhu2023minigpt} have demonstrated remarkable generalization capabilities across diverse tasks and languages. The transformer architecture underlying LLMs emphasizes attention mechanisms, which are instrumental in focusing on relevant parts of the input text. These mechanisms enable the capture of long-range dependencies and contextual relationships, enhancing their versatility across NLP and multimodal applications.

\begin{figure}
    \centering
    \includegraphics[width=0.99\textwidth]{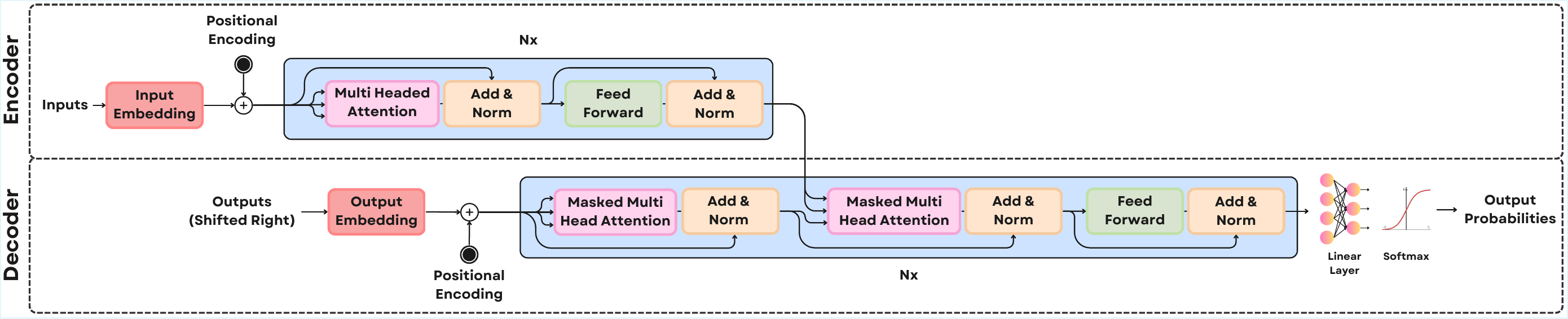}
    \caption{Transformer Architecture: Constructed on an encoder-decoder framework, the architecture employs multiple linear, normalization, embedding, feed-forward, self-attention and cross-attention blocks to generate outputs efficiently \citep{AttentionIsAllYouNeed}.}
    \label{fig:transformer_architecture}
\end{figure}

As depicted in Figure \ref{fig:transformer_architecture}, the core innovation lies in its attention mechanism, which allows the model to dynamically weight the importance of different input elements when producing an output sequence. Unlike traditional models reliant on recurrence or convolutions, Transformers achieve this by employing self-attention to capture dependencies between tokens irrespective of their positional distance. This mechanism computes a contextually aware representation by calculating weighted sums based on learned relevance scores derived from Query, Key and Value projections. The success of this architecture in natural language processing spurred its exploration in other modalities.

\begin{equation}
    \text{Attention}(Q, K, V) = \text{softmax}\left( \frac{QK^\top}{\sqrt{d_k}} \right)V
    \label{eq:attention_equation}
\end{equation}

\vspace{-6pt}

The Vision Transformer (ViT) \citep{ViT} extended Transformer models to computer vision by treating images as sequences of patches, achieving strong performance and often surpassing traditional CNNs on image recognition benchmarks. To further improve attention quality in visual tasks, the Differential Transformer \citep{DiffTransformer} introduced a novel approach called differential attention. This method addresses the issue that standard attention mechanisms often highlight not only important features but also irrelevant background noise. To mitigate this, the Key and Query matrices are each split into two sub-parts. Attention maps are computed separately for these sub-parts and the final attention map is obtained by subtracting one from the other. This subtraction helps cancel out common noise patterns, inspired by noise cancellation principles, resulting in cleaner, more focused attention distributions. This refinement enables the model to attend more precisely to salient visual information, thereby contributing to enhanced performance on visual tasks.

\begin{equation}
    \begin{aligned}
        \text{DiffAttention}(Q_1, K_1, Q_2, K_2, V) =\ & \left( \text{softmax}\left( \frac{Q_1 K_1^\top}{\sqrt{d}} \right) \right. \left. - \lambda \ \text{softmax}\left( \frac{Q_2 K_2^\top}{\sqrt{d}} \right) \right) V
    \end{aligned}
    \label{eq:diff_attention}
\end{equation}

\subsection{Fundamentals of 3D Vision}\label{sec:fundamentals_of_3d_vision}

Three-dimensional vision encompasses the understanding and interpretation of spatial and structural information from three-dimensional data. Unlike 2D vision, which relies on flat image representations, 3D vision captures depth, geometry and spatial relationships, providing a richer context for perception and interaction. Core components of 3D vision include point clouds, meshes and volumetric data, which represent objects and scenes in three-dimensional space.

Techniques such as depth estimation, 3D reconstruction and object recognition form the foundation of 3D vision systems. These systems have applications in diverse fields, including autonomous driving, robotics, augmented reality (AR) and medical imaging. Despite its potential, 3D vision faces challenges such as high computational requirements, sparse and noisy data and the need for specialized hardware, including LiDAR and depth cameras, for capturing 3D data.

\begin{figure}
    \centering
    \includegraphics[width=0.99\columnwidth]{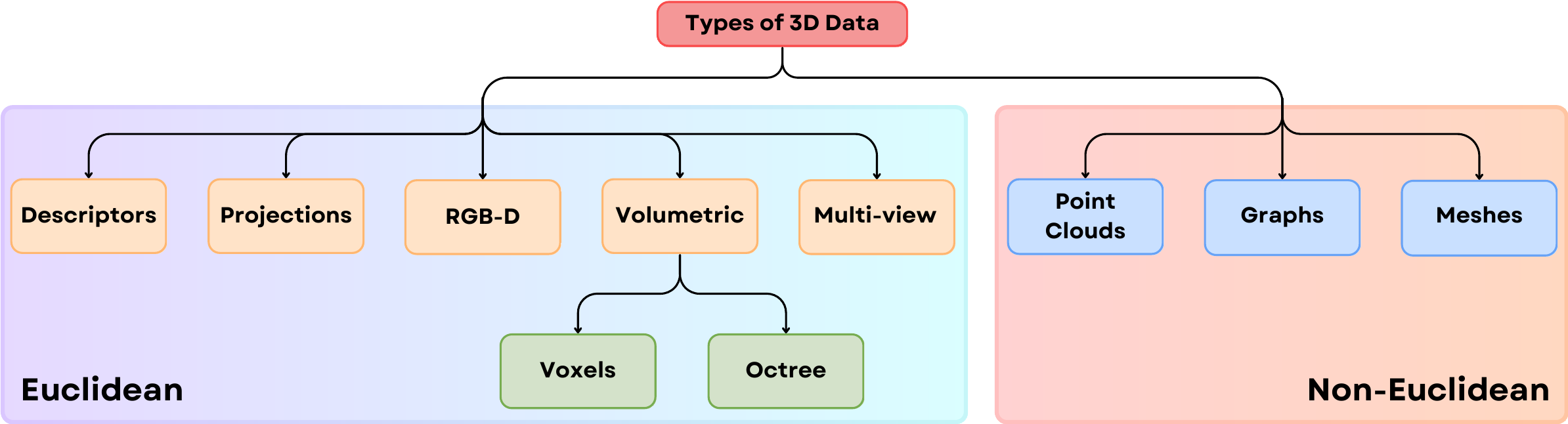}
    \caption{Categories of 3D Data Representations: 3D data can be classified into two primary categories: Euclidean (including Descriptors, Projections, RGB-D, Volumetric and Multi-view) and Non-Euclidean (comprising Point Clouds, Graphs and Meshes) \citep{3dDataRepresentationsSurvey}.}
    \label{fig:3d_data_representations}
\end{figure}

Three-dimensional data can be represented in various forms, with structural and geometric properties varying across representations. As illustrated in Figure \ref{fig:3d_data_representations}, 3D data can be broadly categorized into 3D Euclidean Data and 3D Non-Euclidean Data \citep{3dDataRepresentationsSurvey}. Euclidean data, such as voxel grids and multi-view images, comprises a grid structure, facilitating the extension of 2D Deep Learning (DL) methods to the 3D domain, where networks still utilize convolution operations. In contrast, non-Euclidean 3D data, such as point clouds and meshes, lack a grid array structure, posing challenges for extending existing 2D DL methods to the 3D domain \citep{GenerativeAIMeets3d}.

\vspace{-4pt}

Below is a brief overview of each type of 3D data representation:

\begin{itemize}
    \item \textbf{Descriptors}: These are compact mathematical or statistical summaries of 3D shapes, used for tasks like recognition or matching. Examples include shape histograms and curvature descriptors.
    \item \textbf{Projections}: 3D objects are represented through their 2D projections from multiple viewpoints, simplifying the representation while retaining key spatial information.
    \item \textbf{RGB-D}: Combines RGB color data with depth information captured by devices like Kinect. This format is widely used in robotics and augmented reality applications.
    \item \textbf{Voxel Grid}: A 3D grid cell that represents a portion of space, storing attributes like object presence, color or density. It provides detailed 3D representations but can \mbox{be memory-intensive.}
    \item \textbf{Octree}: A hierarchical structure that divides 3D space into smaller cubes only where needed, reducing memory usage while maintaining detail in occupied regions.
    \item \textbf{Multi-view}: Uses multiple 2D images of a 3D object captured from various angles, allowing algorithms to infer 3D structure.
    \item \textbf{Point Clouds}: A collection of points in 3D space, each representing a specific position on an object's surface. Used in lidar scanning and autonomous vehicles. For capturing point cloud data Matterport pro camera, LiDAR, Microsoft Kinect, etc., can be used.
    \item \textbf{Graphs}: Represent 3D data as nodes and edges, where nodes correspond to object elements and edges represent relationships. Common in structural analysis and \mbox{mesh processing.}
    \item \textbf{Meshes}: Define 3D shapes using vertices, edges and faces, forming a network of polygons (typically triangles). They are widely used in computer graphics.
\end{itemize}

{The distinction between Euclidean (grid-like) and Non-Euclidean (irregular) 3D data provides a helpful starting point for choosing deep learning models, but it's not a strict rule. The main issue is that this classification confuses an object's true physical shape with the digital format we use to represent it. A perfect example of this ambiguity is RGB-D data, which is stored in a grid-like format, similar to a regular image (Euclidean), but contains depth information that describes a 3D scene (Non-Euclidean). To solve this, the community often refers to this as "2.5D" data and resolves the ambiguity based on the specific goal or task. For a task like semantic segmentation, where a pixel-by-pixel map is needed, the RGB-D data is treated as a 2.5D image and processed with CNNs. However, for tasks like robotic grasping that require precise 3D geometry, the data is converted into a point cloud (a non-Euclidean format) and processed with geometric deep learning models \citep{GoingBeyondEuclideanData}. Therefore, the most rigorous approach, supported by surveys such as \citep{3dDataRepresentationsSurvey}, is to view this taxonomy not as a rigid binary choice, but as a flexible, task-dependent framework where the final application determines how the data is interpreted and used.}

\subsection{Motivation Behind Combining LLMs and 3D Vision} \label{sec:motivation}

Recent advancements in LLMs and their potential applications in the 3D Vision field, see Table  \ref{tab:3d_models}, present an opportunity to address critical challenges for embodied agents. Motivated by the integration of LLMs and 3D vision for next-generation multimodal robotic sensing technologies, this research investigates three pivotal research questions (RQ):

\begin{itemize}
    \item \textbf{RQ1:} What are the dominant architectural paradigms and cross-modal alignment strategies for integrating the symbolic, semantic reasoning of Large Language Models with the raw geometric and spatial data from diverse 3D sensors (e.g., LiDAR, RGB-D) to achieve robust spatial grounding and object referencing in robotic systems?
    \item \textbf{RQ2:} What architectural frameworks and semantic grounding techniques enable Large Language Models to interpret and reason over heterogeneous, non-visual sensor data, such as tactile force distributions, thermal signatures and acoustic cues, thereby enriching a robot's 3D world model to enhance situational awareness and allow for more nuanced physical interaction under conditions where visual data is ambiguous or unreliable?
    \item \textbf{RQ3:} Given the inherent challenges of 3D data scarcity and the modality gap between unstructured point clouds and structured language, what emerging methodologies, spanning open-vocabulary pre-training, procedural text-to-3D generation and the fusion of non-visual sensory inputs (tactile, thermal, auditory), are being employed to create more generalizable and robust robotic perception systems?
\end{itemize}

\vspace{-6pt}
\begin{table}
\caption{Summary of 3D data input modalities utilized by various models. The Table categorizes the input types into Euclidean and Non-Euclidean formats, highlighting the specific data representation employed by each study in their methodologies. A \checkmark symbol indicates instances where the model can directly accept the given 3D data representation as input or with minimal adjustments, such as bypassing or modifying specific sub-modules.}
\label{tab:3d_models}
\begin{center}
\begin{tabular}{lcccccc} 
    \toprule
    \multirow{2}{*}{\textbf{Models}} &
    \multicolumn{3}{c}{\textbf{Euclidean}} &
    \multicolumn{3}{c}{\textbf{Non-Euclidean}}\\
    \cmidrule(lr){2-4} \cmidrule(lr){5-7}
    & \textbf{RGB-D} & \textbf{Voxel} & \textbf{Multi-View} & \textbf{Point Cloud} & \textbf{Graph} & \textbf{Mesh} \\ \midrule
    WildRefer \citep{WildRefer} & \checkmark & - & - & \checkmark & - & - \\
    CrossGLG \textsuperscript{1} \citep{CrossGLG} & \checkmark & - & - & - & - & - \\
    3DMIT \citep{3DMIT} & - & - & - & \checkmark & - & - \\
    LiDAR-LLM \citep{LiDARLLM} & - & - & - & \checkmark & - & - \\
    SceneVerse \citep{SceneVerse} & - & - & - & \checkmark & - & - \\
    Agent3D-Zero \citep{Agent3DZero} & - & - & \checkmark & \checkmark & - & - \\
    PointLLM \citep{PointLLM} & - & - & - & \checkmark & - & - \\
    QueSTMaps \citep{QueSTMaps} & \checkmark & - & - & \checkmark & - & - \\
    Chat-3D \citep{Chat3D} & - & - & - & \checkmark & - & - \\
    ConceptFusion \citep{ConceptFusion} & \checkmark & - & - & \checkmark & - & - \\
    RREx-BoT \citep{RRExBoT} & - & - & - & \checkmark & - & - \\
    PLA \citep{PLA} & \checkmark & - & - & \checkmark & - & - \\
    OpenScene \citep{OpenScene} & - & - & \checkmark & \checkmark & - & - \\
    ULIP \citep{ULIP} & - & - & \checkmark & \checkmark & - & - \\
    PolarNet \citep{PolarNet} & \checkmark & - & \checkmark & \checkmark & - & - \\
    \bottomrule
\end{tabular}
\end{center}
\noindent{\footnotesize{\textsuperscript{1} Temporal Input.}}
\end{table}

\begin{figure}
    \centering
    \includegraphics[width=0.9\linewidth]{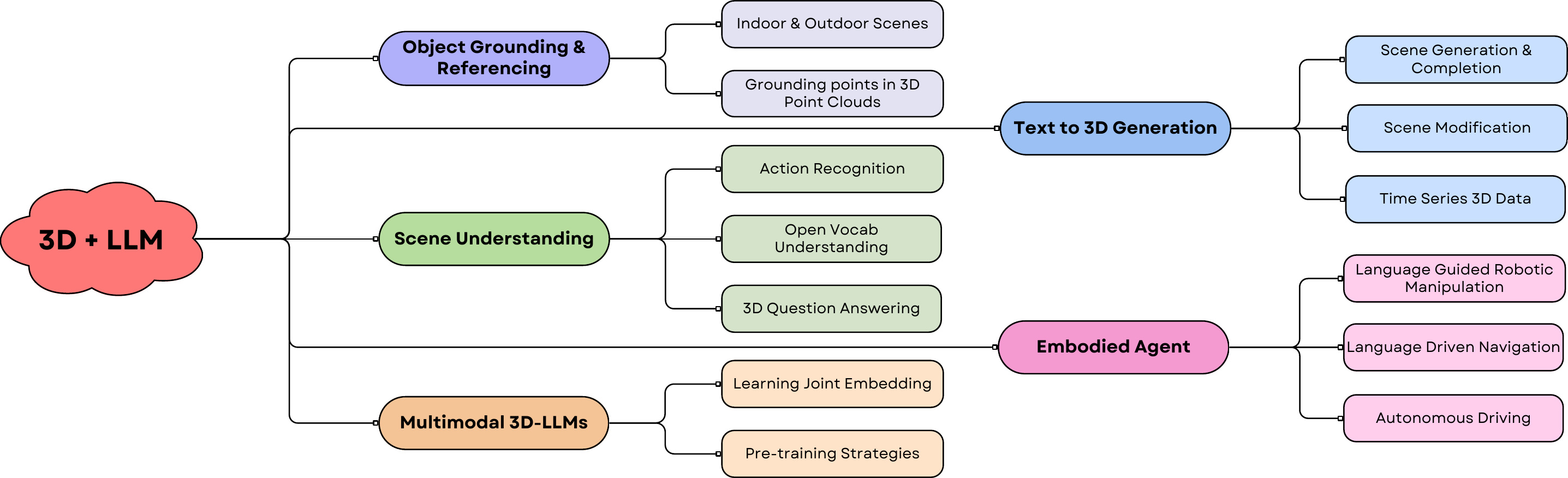}
    \caption{Research areas at the intersection of 3D and LLMs: Advancing capabilities in 3D scene understanding, generation and modification; object grounding and referencing; developing embodied agents; and integrating 3D data with multimodal language models.}
    \label{fig:sub_topics_flowchart}
\end{figure}

\vspace{-4pt}
This article explores these questions, reviewing recent advancements in the field and identifying future research directions. The integration of LLMs with 3D vision leverages the complementary strengths of these technologies, allowing for enhanced capabilities. LLMs excel in reasoning, language understanding and contextual interpretation, while 3D vision provides detailed spatial and geometric information. Combining these capabilities facilitates the development of more intelligent and context-aware systems. Figure \ref{fig:sub_topics_flowchart} illustrates the various research areas and applications within the 3D + LLM domain. It is important to note that the boundaries between these categories are fluid, resulting in significant overlap in the technologies and architectures used to address each problem.

In robotics, for example, understanding 3D scenes through visual data can be enhanced by LLMs' ability to interpret and generate instructions in natural language. Similarly, in applications such as autonomous vehicles and augmented reality (AR), LLMs can provide semantic understanding and decision-making capabilities, complementing the spatial awareness afforded by 3D vision. The convergence of these domains holds promise for innovative solutions to complex real-world problems.

{Practical applications of integrating 3D vision with large language models (LLMs) span crucial areas like robotic grasping and autonomous driving, where spatial intelligence and contextual reasoning are tightly interwoven. For robot grasping, frameworks such as ORACLE-Grasp \citep{OracleGrasp} leverage multimodal large models to semantically interpret 3D environments and zero-shot select task-appropriate grasps on previously unseen objects, avoiding reliance on exhaustive labeled datasets or retraining. Leveraging video-based 3D geometry encoders, Multimodal LLMs can learn to reason about 3D scenes directly from video streams, enhancing scene understanding and spatial inference without explicit 3D data inputs \citep{EnhancingMLLMsWith3DVision}. Such capabilities empower robots to manage unstructured \mbox{environments robustly.}}

In the autonomous driving domain, vision-language models enrich 3D perception, planning and decision-making by incorporating language-guided reasoning, supported by datasets like OmniDrive \citep{OmniDrive} that embed counterfactual reasoning for better trajectory planning. Novel 3D tokenization methods, such as Atlas \citep{Atlas}, incorporate the physical priors of 3D space to improve reliability in end-to-end planning tasks. Large-scale 3D-grounded datasets like 3D-GRAND \citep{3DGrand} provide densely grounded 3D text pairs that significantly reduce hallucinations while improving grounding accuracy in 3D-LLMs, facilitating more reliable scene interpretation and interaction. These advancements enable practical tasks, including grounded object reference, scene captioning and spatial question answering, essential for effective embodied agents and autonomous systems operating in real-world 3D environments.

As shown in Table \ref{tab:3dllm_vs_traditional_methods} the integration of Large Language Models (LLMs) into 3D pipelines introduces a substantial computational gap compared to traditional methods, motivating a deeper investigation into their practical viability. While established 3D processing frameworks often operate efficiently within a range of 10–200 GigaFLOPs (GFLOPs), consume a modest 20–80 Watts on edge hardware and achieve low latencies between \mbox{10–100 ms \citep{PointNext,FrameMining},} emerging LLM–3D systems present a starkly different profile. These advanced pipelines, exemplified by architectures like OmniDrive \citep{OmniDrive} and \mbox{Lidar-LLM \citep{LiDARLLM},} escalate computational demands to the 1–5+ TeraFLOPs (TFLOPs) scale, require over 250 Watts of power typically from high-performance GPUs and exhibit significantly higher latencies from 200 ms to several seconds. This profound increase in resource requirements, consistently observed across various 3D-language \mbox{tasks \citep{3DGPT,PolarNet},} creates \mbox{a critical} bottleneck for their deployment in real-time, power-constrained applications such as autonomous robotics and embodied agents, thus underscoring the urgent need to benchmark and analyze the computational efficiency of these powerful new models.

While recent comprehensive surveys provide expansive overviews of multimodal Large Language Models (LLMs), often covering model architectures and applications across general domains with a primary focus on 2D vision-language fusion \citep{yin2023survey,MultimodalFoundationModels}, this review uniquely advances the discourse by concentrating specifically on the critical intersection of LLMs and 3D vision, framed entirely within the context of intelligent robotic perception and autonomy. We move beyond 2D representations to deeply analyze varied 3D data types, including point clouds, voxels and meshes, and further distinguish our contribution by systematically examining the integration of non-visual sensory inputs such as tactile, thermal and auditory data, which are pivotal for robust embodied interaction \cite{MultimodalFusionAndVLMs}. Consequently, our distinct robotics-specific contribution is to provide a targeted roadmap that bridges high-level semantic reasoning with low-level sensing hardware, addressing key challenges in spatial grounding, open-vocabulary scene understanding and real-time deployment bottlenecks on resource-constrained platforms. By cataloging robotics-centric datasets and evaluation metrics, this review serves as a specialized guide for developing the next generation of embodied agents capable of complex perception, reasoning and interaction in the physical world \cite{MultimodalAlignmentAndFusionSurvey}.

\begin{table}
\caption{{Comparison between LLM–3D pipelines and traditional methods.\label{tab:3dllm_vs_traditional_methods}}}
\begin{center}
\begin{tabular}{ccc} 
    \toprule
    \textbf{{Metric}} & \textbf{{Traditional 3D Methods}} & \textbf{{LLM-3D Pipeline}}\\
    \midrule
    {FLOPs} & {10–200 GFLOPs per inference} & {1–5+ TFLOPs per inference}\\
    {Power (Watts)} & {20–80 W (embedded GPUs, edge devices)} & {250+ W (high-performance GPUs/clusters)}\\
    {Latency} & {10–100 ms} & {200 ms}\\
    \bottomrule
\end{tabular}
\end{center}
\end{table}
\unskip

\section{Sensing Technologies for Robotic Perception}\label{sec:sensing_technologies}

In robotics, vision systems are critical as they enable robots to interact intelligently with their environment by providing essential feedback about object locations and robot poses. This information is vital for controlling robot motion. Visual serving techniques are often applied for precise control, ranging from positioning robotic arms \citep{DSPBasedVisualServoing} and helping humanoid robots navigate environments \citep{VisionBasedLocalization} to estimating object poses for \mbox{recognition \citep{FullPoseEstimation, CornerBased3DObjectPoseEstimation}} and assisting with robot motion estimation \citep{KalmanFilterForRobotEgoMotionEstimation, RobotHandPoseEstimation}. Additionally, these vision systems contribute to sophisticated tasks like object grasping \citep{DirectPerceptionObjectGrasping, ConstructionOfA3DObject}.

Beyond merely detecting objects, robotic vision aims to help robots understand their representations, which is key for efficient decision-making and task execution. Effective representation methods should be independent of changes in object orientation or viewpoint. As discussed earlier Section  \ref{sec:fundamentals_of_3d_vision}, 3D data comes in various formats, as in Figure \ref{fig:3d_data_representations}, each needing to be mapped into latent spaces that facilitate comprehensive object understanding.

Modern advancements in stereo vision and image acquisition systems have further cemented the importance of 3D vision for mobile robots \citep{Gunatilake2021}.
Stereo Vision uses two or more cameras to infer depth via triangulation from disparities between views and produces dense or semi-dense depth maps, often registered with RGB images (RGB-D). They are commonly used for close-range manipulation \citep{CloseRangeManipulationStereoVision}, obstacle avoidance \citep{ObstacleAvoidanceStereoVision1, ObstacleAvoidanceStereoVision2}, visual odometry \citep{OdometryStereoVision1, OdometryStereoVision2} and 3D reconstruction \citep{3dReconstructionStereoVision} in textured environments. But they are sensitive to textureless surfaces, baseline limits range/accuracy and computationally intensive matching.

\begin{figure}
    \centering
    \includegraphics[width=0.8\linewidth]{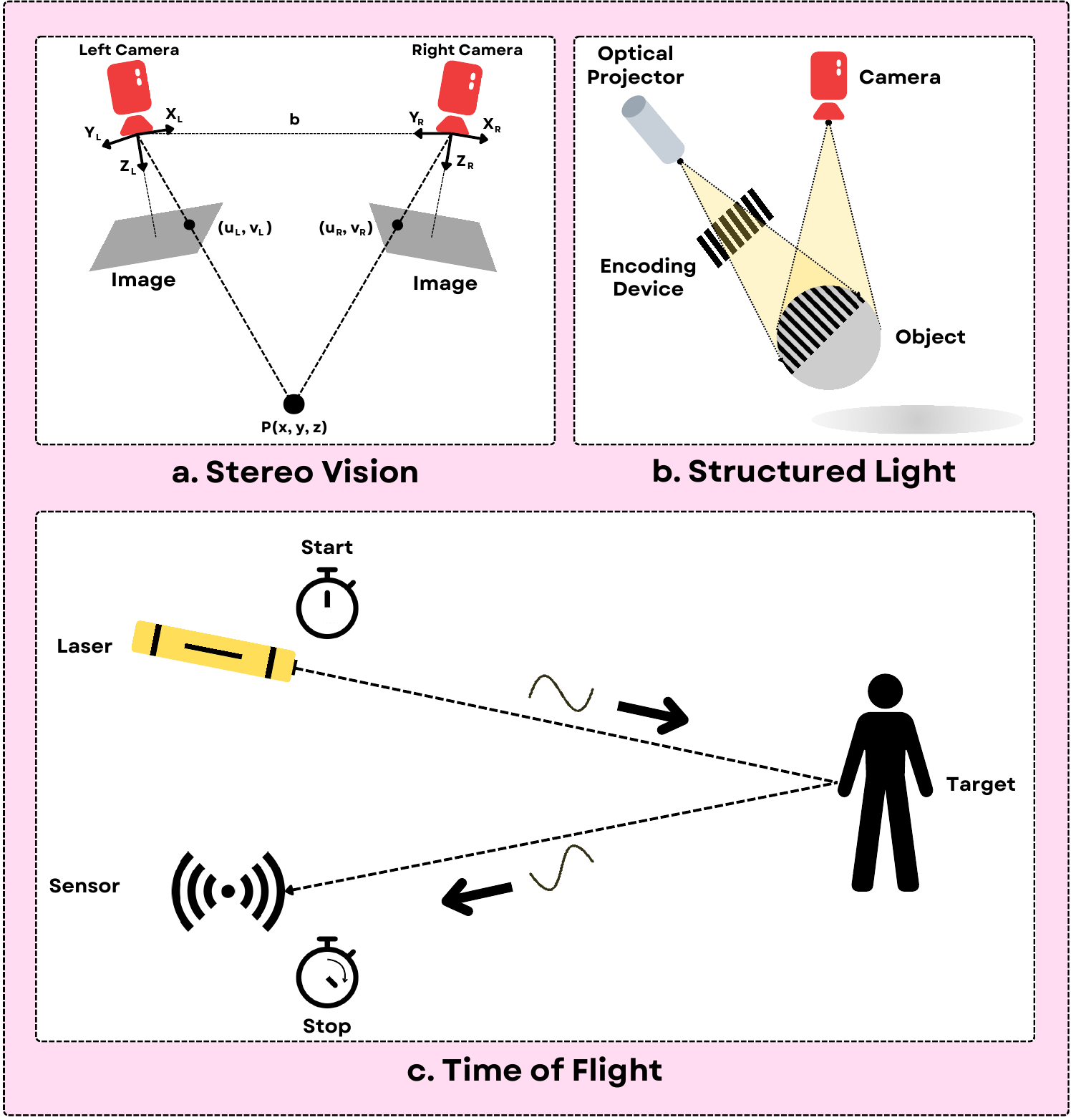}
    \caption{Overview 
 of different 3D vision sensing techniques work: stereo vision (\textbf{a}),structured \mbox{light (\textbf{b}) and} time of flight (\textbf{c}), illustrating their working principles and components for capturing \mbox{spatial information.}}
    \label{fig:3d_sensing_techniques}
\end{figure}

Visual sensors in robotic systems are typically categorized into three types \citep{AdvancesInSensingAndProcessing}: linear transducers, such as single laser radar systems \citep{3DMeasurementsFromImagingLaserRadars}; 2D sensor arrays, such as embedded cameras \citep{CMOSCameraForIndoorAndOutdoor}; and 3D depth sensors utilizing specialized light-based cameras \citep{ActiveOpticalRangeImagingSensors}. Current technologies for 3D depth sensing rely on monocular structured light methods \citep{BayesianSensor}, binocular structured light systems \citep{DesignOfA3DInfrared} and time-of-flight (TOF) approaches \citep{3DShapeScanning}, see Figure \ref{fig:3d_sensing_techniques}.

3D sensing methods are categorized into active and passive techniques, as summarised in Table \ref{tab:3d_sensing}. Passive methods determine an object's shape by using its reflectance and the scene's illumination, which means they don't need active devices. These methods generate range data, representing the distances between the sensor and the object's surface. However, Shape from Shading (SFS) is not suitable for mirrored surfaces, as it relies on the shadows cast by the surface under different lighting conditions to create a 3D model and has limited applicability in robotics due to its requirement for accurate light source parameters. This requirement renders SFS ineffective in complex lighting conditions, such as outdoor scenes. Shape from Texture (SFT) also has limitations, as it requires high-quality texture data and precise modeling of projection distortions, making it challenging to use in novel robotic environments.

\begin{table}
\caption{Comparison of various 3D Sensing Methods \citep{AdvancesInSensingAndProcessing, SOTA3DImagingSensors}.\label{tab:3d_sensing}}
\begin{center}
\begin{tabular}{cccc} 
    \toprule
    \textbf{Method} & \textbf{Principle} & \textbf{Modality} & \textbf{Type}\\
    \midrule
    Stereo Vision & Triangulation & Passive & Direct\\
    Structured Lighting & Triangulation & Active & Direct\\
    Shape from Shading & Monocular Images & Passive & Indirect\\
    Shape from Texture & Monocular Images & Passive & Indirect\\
    Time of Flight & Time Delay & Active & Direct\\
    Interferometry & Time Delay & Active & Direct\\
    LiDAR & Time Delay/Phase Shift & Active & Direct\\
    \bottomrule
\end{tabular}
\end{center}
\end{table}
\unskip

Structured lighting projects known light patterns (e.g., stripes, grids) onto the scene and observes their deformation using a camera to calculate depth via triangulation. It produces dense RGB-D point clouds or depth maps. This method is widely used because of its simplicity and precision in applications like depth perception \citep{ColorEncodedStructuredLight, VisionProcessingForRealtime}. In mobile robotics, structured lighting facilitates navigation and obstacle detection \citep{IndoorMobileRobotObstacle, HelpmateAutonomousMobileRobot}, scene \mbox{understanding \citep{IndoorSceneSegmentation},} and 3D reconstruction \citep{StructuredLightUnderwaterSurface, HighResolutionUnderwater}. It is also applied in shape \mbox{acquisition \citep{RapidShapeAcquisition},} object modeling \citep{DualBeamStructuredLight} and 3D hand-eye robot vision systems \citep{3DHandEyeRobotVisionSystem}. \mbox{Table \ref{tab:combined_3d_sensors}} contains the basic information about various sensors for different techniques available in the market. But it's sensitivity to ambient light (especially sunlight), reflective surfaces and struggles with inter-sensor interference, leading to major limitations and range is also typically limited.

\begin{table}
\footnotesize
\caption{Common 3D Sensor Examples for Robotic Perception Grouped by Technology \citep{AdvancesInSensingAndProcessing,SOTA3DImagingSensors,LidarComparison1, LidarComparison2}.\label{tab:combined_3d_sensors}}
\begin{center}
	\newcolumntype{C}{>{\centering\arraybackslash}X}
    \begin{tabularx}{\fulllength}{ccCcc}
        \toprule
        \textbf{Technology}	& \textbf{Sensor Example} & \textbf{Key Specs (Resolution/Rate/Points)} & \textbf{Typical Range (m)} & \textbf{Integrated IMU}\\
        \midrule
        \multirow[m]{5}{*}{Stereo Vision}
        & Bumblebee 2 & \makecell{648 $\times$ 488 @ 48 fps} & 0.1--20 & $\times$\\
        & Bumblebee XB3 & \makecell{1280 $\times$ 960 @ 16 fps} & 0.1--20 & $\times$\\
        & Nerian SP1 & \makecell{1440 $\times$ 1440 @ 40 fps} & - & $\times$\\
        & DUO3D Stereo Camera & \makecell{640 $\times$ 480 @ 30 fps} & - & \checkmark\\
        & OrSens3D Camera & \makecell{640 $\times$ 640 @ 15 fps} & - & $\times$\\
        \midrule
        \multirow[m]{7}{*}{Structured Light}
        & Microsoft Kinect v1 & \makecell{320 $\times$ 240 @ 30 fps} & 1.2--3.5 & $\times$\\
        & PrimeSense Carmine & \makecell{640 $\times$ 480 @ 30 fps} & 0.35--3 & $\times$\\
        & Orbbec Astra Pro & \makecell{640 $\times$ 480 @ 30 fps} & 0.6--8 & $\times$\\
        & Intel RealSense D435 & \makecell{1280 $\times$ 720 @ 90 fps} & 0.1--10+ & $\times$\\
        & Intel RealSense R200 & \makecell{640 $\times$ 480 @ 60 fps} & 3--4+ & $\times$\\
        & Intel RealSense R399 & \makecell{640 $\times$ 480 @ 60 fps} & 0.1--1.2 & $\times$\\
        & Intel RealSense ZR300 & \makecell{480 $\times$ 360 @ 60 fps} & 0.5--2.8 & \checkmark\\
        \midrule
        \multirow[m]{9}{*}{Time-of-Flight}
        & Microsoft Azure Kinect & \makecell{1024 $\times$ 1024 @ 30 fps} & 0.5--3.86 (NFOV) & \checkmark\\
        & Microsoft Kinect v2 & \makecell{512 $\times$ 484 @ 30 fps} & 0.5--4.5 & $\times$\\
        & SICK Visionary-T & \makecell{144 $\times$ 176 @ 30 fps} & 1--7 & $\times$\\
        & Basler ToF ES Camera & \makecell{640 $\times$ 480 @ 20 fps} & 0--13 & $\times$\\
        & MESA SR4000 & \makecell{176 $\times$ 144 @ 54 fps} & 5/10 & $\times$\\
        & MESA SR4500 & \makecell{176 $\times$ 144 @ 30 fps} & 0.8--9 & $\times$\\
        & Argos3D P100 & \makecell{160 $\times$ 120 @ 160 fps} & 3 & $\times$\\
        & Argos3D P330 & \makecell{352 $\times$ 287 @ 40 fps} & 0.1--10 & $\times$\\
        & Sentis3D M520 & \makecell{160 $\times$ 120 @ 160 fps} & 0.1--5 & $\times$\\
        \midrule
        \multirow[m]{4}{*}{LiDAR}
        & Velodyne VLP-16 (Puck) & \makecell{16 Ch $\sim$300k pts/s} & $\sim$100 & Optional/Ext.\\
        & Ouster OS1-64 & \makecell{64 Ch $\sim$1.3M pts/s} & $\sim$100--120 & \checkmark\\
        & Hesai PandarXT-32 & \makecell{32 Ch $\sim$640k pts/s} & $\sim$120 & \checkmark\\
        & Livox Mid-70 & \makecell{NRS $\sim$100k pts/s} & $\sim$70 & \checkmark\\ 
        \bottomrule
    \end{tabularx}
    \end{center}
	\noindent{\footnotesize{Ch: Channels; NRS: Non-repetitive Scan; 3D: three-dimensional; IMU: inertial measurement unit; ToF: time of flight; fps: Frames per second; \checkmark: Present; $\times$: Absent.}}
\end{table}

Time-of-Flight (ToF) measures the round-trip time of emitted light (often infrared) reflecting off surfaces. Indirect ToF measures phase shift, while direct ToF measures pulse delay, eventually both of them producing depth maps, often with an associated \mbox{amplitude/intensity} image. They can also provide RGB-D when combined with an RGB sensor. They are good for medium-range depth sensing, less affected by texture than \mbox{stereo/SL} and are used in gesture recognition \citep{GestureRecognitionToF}, indoor navigation \citep{IndoorNavigationToF1, IndoorNavigationToF2ObstacleDetectionToF} and obstacle detection \citep{IndoorNavigationToF2ObstacleDetectionToF}. As a limitation, they can suffer from multi-path interference, lower spatial resolution compared to SL and accuracy can depend on surface reflectivity \citep{MultiPathInterferenceToF, LowSpatialResolutionToF}. {While Time-of-Flight (ToF) sensors are powerful, their deployment in robotics has historically been hampered by multipath interference (MPI), where signals reflecting off multiple surfaces create erroneous depth measurements that can compromise navigation and manipulation, especially in cluttered indoor environments with corners or reflective objects. To address this, recent advancements have focused on two complementary strategies: innovative multi-tap pixel architectures and sophisticated deep learning-based denoising. On the hardware front, multi-tap pixel designs allow the sensor to capture the incoming light at multiple distinct time intervals or gates, which enables it to better separate the direct light signal from the delayed, indirect signals causing the interference \citep{ToFSensorUsingFourTapLockInPixels}. This hardware-level solution provides more accurate depth information from the source, significantly improving reliability in scenes with challenging materials like glass. Concurrently, software-based approaches utilizing deep learning have demonstrated remarkable success in mitigating the remaining distortions. These methods use neural networks, such as CNNs, that are trained on vast datasets to recognize the specific patterns of MPI noise and effectively remove them from the depth images, resulting in smoother and more physically accurate 3D maps \citep{DLForMultiPathErrorRemovalInToFSensors, LearningToRemoveMultipathDistortions, DeepEndToEndToFImaging, MultipathInterferenceSuppression}. The synergy of these advanced multi-tap sensors and powerful denoising algorithms has a direct and substantial impact on robotic deployment by reducing depth errors, leading to more robust 3D mapping and obstacle detection and ultimately enabling safer and more reliable navigation and manipulation in complex, real-world scenarios \citep{DLForTransientImageReconstruction}. This fosters greater autonomy by reducing the computational burden for downstream tasks, such as scene understanding and path planning, thereby expanding the operational capabilities of robots in dynamic, human-centric environments.}

LiDAR (Light Detection and Ranging) emits laser pulses and measures the time/phase of reflected signals to determine distance. Mechanical LiDARs spin to cover 360°, while Solid-State LiDARs use MEMS mirrors or optical phased arrays. They generate sparse or dense 3D point clouds, often with intensity values. They are standard for autonomous driving (long-range perception, mapping), outdoor/large-scale mapping \citep{OutdoorMappingLidar} and mobile robot navigation \citep{MobileRobotNavigationLidar}. But for the limitations, typically they are more expensive, data is often sparser than camera-based methods (especially vertically), can struggle with certain atmospheric conditions (fog, rain) or specific materials (black, highly reflective) \citep{AtmosphericInterferenceLidar, MaterialChallengesLidar, OutdoorMappingLidar}. There is a significant, underexploited opportunity in integrating the nuanced physical characteristics of advanced LiDAR systems into Large Language Model (LLM) pipelines. Physics-aware LiDAR variants, such as Frequency-Modulated Continuous-Wave (FMCW) and those operating at different wavelengths (905 nm vs. 1550 nm), offer richer data that can enhance robotic reasoning. FMCW LiDAR, for instance, provides simultaneous range and velocity measurements for every point, granting it superior resilience to interference and robust performance in adverse weather. Integrating this 4D data into LLM frameworks requires specialized encoders that process not just geometry, but also physical metadata like velocity and reflectance, a concept well-suited for the hierarchical embeddings in transformer architectures as explored in recent bio-inspired systems \citep{LiDARLLM}. Similarly, the choice of wavelength critically impacts scene understanding; 1550 nm systems provide better eye safety and longer range, whereas 905 nm systems are more cost-effective and perform better in rain or fog \citep{Why905and1550nm, LidarSystems}. A physics-aware LLM could leverage these wavelength-specific features, such as atmospheric attenuation and material reflectance, to improve its reasoning about the environment and its own sensor limitations. To achieve this integration, a clear strategy is necessary: first, feature vectors must be engineered to explicitly include physical data, such as velocity or phase. Next, these features must be aligned with language embeddings using cross-modal transformer models, similar to the approaches in recent LiDAR-LLM frameworks \citep{LiDARLLM}. This, however, necessitates the creation of new training datasets containing rich text descriptions linked to this physical data. Successfully implementing this would significantly enhance applications like autonomous driving in dynamic or challenging weather conditions. Nonetheless, major challenges remain, including the scarcity of annotated physics-aware LiDAR datasets, the computational cost of processing such complex data in real-time and the need for adaptive fusion models that can intelligently leverage different sensor modalities based on context, a key challenge in the broader field of robotic perception and autonomy.

The field of robotic vision benefits from the continuous development of new sensors. Obtaining high-quality initial data directly can significantly reduce computational demands in subsequent processes. A major challenge is the high cost of 3D acquisition equipment, especially for small and medium-sized enterprises (SMEs), though increased market competition and technological advancements are gradually lowering prices. The complexity of these systems, which necessitates skilled operators, also hinders their widespread use in industrial applications.

While 3D vision provides essential geometric understanding, achieving truly robust and nuanced interaction with the world often requires robots to leverage other senses. Vision alone can be ambiguous or insufficient, especially during physical contact, in occluded scenarios or when non-visual cues are critical. Integrating modalities like touch, hearing and thermal perception, often fused and interpreted with the aid of advanced models like LLMs, allows robots to build a more complete and actionable understanding of their surroundings and interactions. Refer to table  \ref{tab:other_sensors_concise} for an overview of tactile, audio and thermal sensor examples.

\begin{table}
\small
\caption{Concise Examples of Tactile, Auditory and Thermal Sensors in Robotics \citep{TactileSensorsComparison,BioTac,GelSightMini}.\label{tab:other_sensors_concise}}
\begin{center}
    \begin{tabularx}{\textwidth}{l XXXXX}
        \toprule
        \textbf{Type} & \textbf{Sensor} & \textbf{Principle/Specs} & \textbf{Output} & \textbf{Applications} & \textbf{Limitations}\\
        \midrule
        \multirow{3}{*}{Tactile} 
        & GelSight & Optical cam; elastomer def. & Hi-res surface, texture, force map & Dexterous manipulation, object ID, slip detection & Bulky, wear, calibration, high processing\\
        \cmidrule{2-6}
        & BioTac/Weiss & Biomimetic; piezoresistive array & Force, vibration, temperature & Grasp stability, material ID, HRI & Wiring, calibration drift, signal interpretation\\
        \cmidrule{2-6}
        & FlexiForce & Thin-film piezoresistive & Single-point force & Contact detection, basic force sensing & Low spatial info, shear wear\\
        \midrule
        \multirow{3}{*}{Audio}
        & ReSpeaker & MEMS mic array (4–8) & Multi-channel audio, DOA, text & Voice command, speaker localization, ambient sound monitoring & Noise handling, source separation, non-speech interpretation\\
        \cmidrule{2-6}
        & Built-in Mic Arrays & Custom array config. & Similar to above & Same as above & Proprietary processing, limited flexibility\\
        \cmidrule{2-6}
        & Single Mic & Condenser/MEMS & Mono audio & Sound recording, basic event detection & No directional info, noise sensitivity\\
        \midrule
        \multirow{3}{*}{Thermal}
        & FLIR Lepton & LWIR, 160 × 120 px & Low-res thermal image & Presence detection, human detection & Low resolution and frame rate, emissivity dependence\\
        \cmidrule{2-6}
        & FLIR Boson & LWIR, 320 × 640 px & Medium/high-res thermal image & Surveillance, machine monitoring, HRI & High cost, lower resolution than visual\\
        \cmidrule{2-6}
        & Seek CompactPRO & LWIR, 320 × 240 px & Medium-res thermal image & Easy integration, human detection, diagnostics & Varying performance, requires adaptation\\
        \bottomrule
    \end{tabularx}
\end{center}
\end{table}

Tactile Sensing provides the crucial sense of touch, essential for dexterous manipulation and safe physical interaction, capabilities often limited by vision, particularly when a target object is occluded by the robot's own gripper. Robots equipped with tactile sensors can confirm stable grasps, actively identify objects based on perceived texture, shape or hardness, detect slip during manipulation and adjust grip force accordingly and gather vital safety feedback during physical human-robot interaction (HRI) to avoid exerting excessive force \citep{ShadowRobot}. Sensor technologies range from simple binary contact switches to complex arrays mounted on robot fingertips or palms (e.g., on platforms like the Shadow Dexterous Hand \citep{ShadowRobot} or integrated into grippers like those from Robotiq or Weiss Robotics). These sensors rely on various physical principles, including piezoresistive, capacitive, piezoelectric or optical methods. The data generated can include force distributions across a contact area, pressure maps, high-resolution contact images capturing micro-geometry and \mbox{texture \citep{ProgressInTemperatureTactileSensors},} vibration patterns and even thermal contact properties (like with the BioTac sensor). Despite their benefits, significant challenges remain in tactile sensing, including the complexity of integrating wiring and electronics into compact robotic hands, sensor durability under repeated contact, reliable calibration and the computational difficulty of interpreting the rich, high-dimensional tactile data streams in real-time \citep{ProgressInTemperatureTactileSensors}. {Formally connecting haptic feelings, like “slippery” or “rough,” with natural language is achieved by creating a latent space, where both the data from a robot’s touch sensors and the words used to describe that feeling share the same feature vector. The primary technique involves using a multimodal learning framework, which combines specialized tactile encoders, often adapted from audio processing models because of their similar data structures, with powerful large language models. Through a process called contrastive learning, the system is trained to pull the vector for a specific touch signal (e.g., the high-frequency vibrations from a rough surface) close to the vector for the corresponding word (``rough'') within this shared space. This ensures that the robot’s sensory experience and human language are quantitatively and qualitatively aligned. More advanced models enhance this process by incorporating material priors, which help the system make better guesses about new textures it encounters, as demonstrated by frameworks like RETRO \citep{RETRO}. For these systems to work effectively, they must be trained on large, specifically annotated multimodal datasets, such as HapticCap \citep{Hapticcap}, Touch-Vision-Language sets \citep{TVL} and others that show success in tasks like zero-shot texture recognition \citep{MultimodalZeroShotLearningForTactileTextureRecognition}. The ultimate goal, as explored in foundational reviews of latent and embedding spaces, is to create a robust connection that allows for two-way communication; a robot can not only generate a linguistic description of a new surface it touches but can also interpret a verbal command, like “apply a gentle, smooth motion,” and translate it into a specific physical action, a capability explored in research that embeds force profiles and language together \citep{CrossModalityForceAndLanguageEmbeddings}. The success of this language-to-touch alignment is then measured by analyzing how closely related sensory signals and their word descriptors cluster together in the latent space, confirming that the system has built a meaningful bridge between feeling and speaking, see Figure \ref{fig:multimodal_robot_perception}.

\begin{figure}
    \centering
    \includegraphics[width=0.99\textwidth]{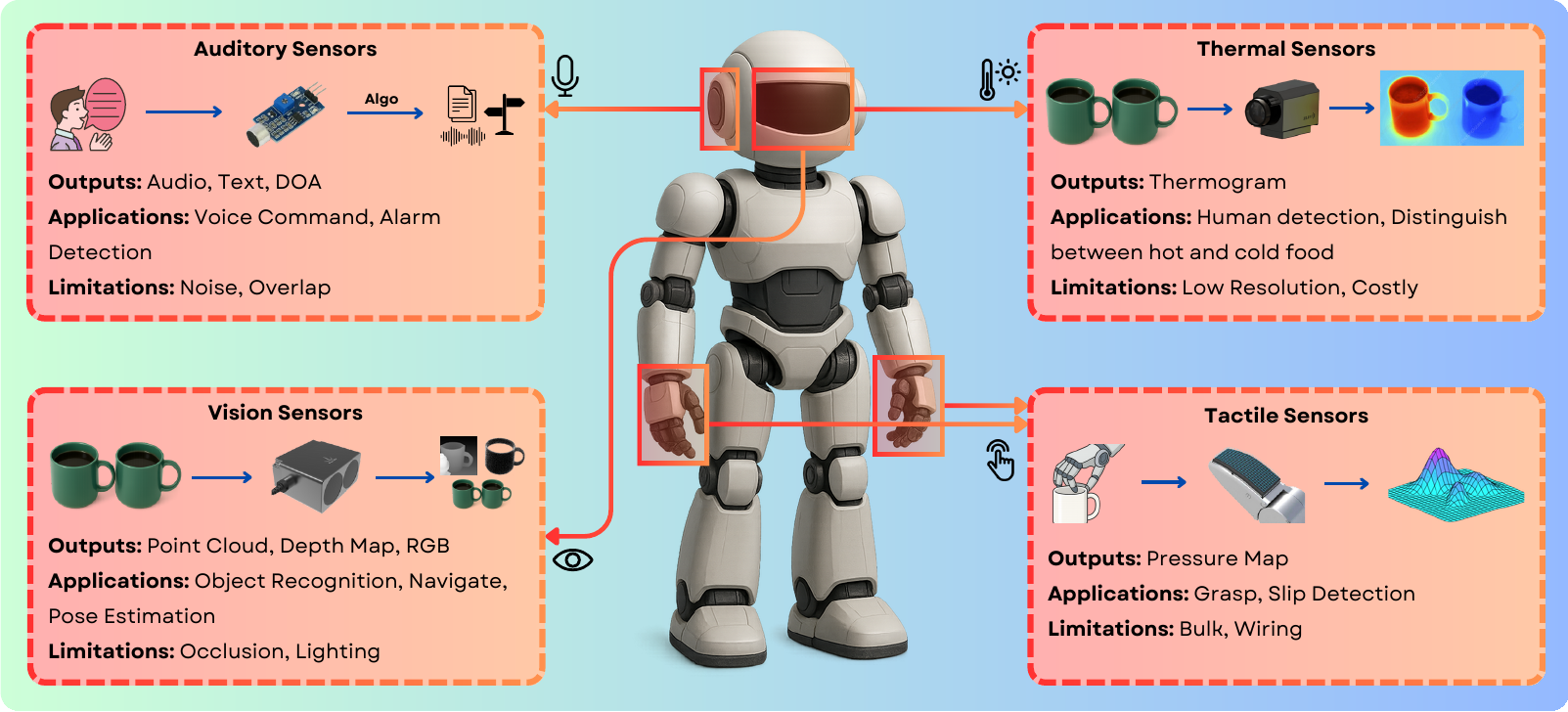}
    \caption{Multimodal robot perception: A robot equipped with visual, tactile, auditory and thermal sensors, illustrating their respective applications and limitations. Visual sensors aid navigation but are hindered by occlusion; tactile sensors enable manipulation but are often bulky; auditory sensors process voice commands but struggle with noise; and thermal sensors detect heat for safety but are constrained by low resolution and cost \textit{(Image of central robot generated via ChatGPT-4o).}}
    \label{fig:multimodal_robot_perception}
\end{figure}

Auditory Sensing equips robots with the ability to perceive sound events and understand spoken language, enabling richer interaction and broader environmental awareness. This is critical for applications ranging from service robots responding to voice commands or calls for help, to industrial robots detecting machine faults through characteristic sounds or mobile robots navigating based on auditory cues. Robots need to detect important environmental sounds (alarms, impacts, breaking glass), robustly understand human speech commands even in noisy conditions, localize sound sources to orient towards a speaker or event and maintain general environmental awareness through background sound analysis. While single microphones can be used, microphone arrays (often in circular or linear configurations) are commonly used in robotics, as they enable sound source localization through techniques like beamforming \citep{SoundSourceLocalization}. The raw data consists of audio waveforms, which are often processed into spectrograms (time-frequency representations) or specific acoustic features (e.g., Mel-Frequency Cepstral Coefficients, MFCCs) for tasks like sound \mbox{classification \citep{SpeechRecognitionMFCC}.} For speech interaction, output includes the recognized text and potentially the estimated direction of the speaker. Key challenges include effectively filtering background noise, separating overlapping sound sources (the ``cocktail party \mbox{problem'') \citep{CocktailPartyProblem},} and accurately interpreting the semantic meaning of diverse non-speech sounds within the robot's operational context. {Transformer-based temporal models have become foundational for aligning acoustic sequences with textual reasoning, enhancing robotic perception in complex environments. These models utilize self-attention to capture long-range dependencies within sequential acoustic inputs, such as waveforms and spectrograms, while cross-attention mechanisms synchronize these features with textual representations for robust semantic grounding and reasoning \citep{CLIP}. Architectures such as AudioCLIP exemplify this by creating a shared latent space for audio and text, enabling advanced tasks including acoustic event recognition, voice command interpretation and audio-driven robotic control. The integration of such temporal models is now a core component of modern Multimodal Large Language Models (MM-LLMs), which unify acoustic sensing with other modalities to perform sophisticated reasoning in dynamic settings \citep{zhang2024mm, Kamath2024}. This synergy between acoustic data and language processing is critical for improving robotic autonomy, supporting applications from speech-driven manipulation to acoustic scene classification in embodied agents.}

Thermal Sensing allows robots to perceive the environment based on temperature distributions, offering unique advantages independent of visible light conditions. This is particularly useful for detecting humans or animals reliably in darkness, fog, smoke or cluttered environments (e.g., in search and rescue scenarios or for human-aware \mbox{navigation) \citep{SearchAndRescueThermal, PlanningAndSearch},} monitoring machinery temperature for predictive maintenance or detecting overheating failures \citep{SearchAndRescueThermal}, identifying active heat sources or thermal leaks and generally enhancing safety in HRI by detecting human presence when vision might fail. The primary sensors are thermal cameras operating typically in the Long-Wave Infrared (LWIR) spectrum (roughly 8--14 $\upmu$m), capturing emitted rather than reflected radiation. They produce thermal images (thermograms), where pixel intensity values correspond directly to estimated surface temperatures. While powerful, thermal sensing in robotics faces challenges: thermal cameras often have lower spatial resolution and slower frame rates compared to visual cameras \citep{LowSpatialResolutionThermal}, interpreting thermal signatures requires contextual knowledge (e.g., emissivity of different materials affects readings) and high-performance thermal cameras can be costly, although smaller, lower-cost modules (like the FLIR Lepton series) are becoming increasingly common for robotic integration \citep{SearchAndRescueThermal}. {Beyond basic heat detection, a significant emerging trend involves leveraging Large Language Model (LLM) priors to dramatically enhance the interpretability of low-resolution thermal imagery, a process known as semantic super-resolution. This approach enables a robot to infer rich, detailed semantic context that goes far beyond what the raw thermal pixel data can provide on its own. By fusing sparse thermal signals with other sensor modalities, such as RGB cameras or LiDAR and integrating the extensive world knowledge embedded in LLMs, robots can effectively "upsample" the meaning of a scene, allowing for robust object recognition and human presence detection even in visually degraded conditions, such as complete darkness or thick fog. For example, pioneering frameworks like RTFNet have demonstrated effective fusion of RGB and thermal data for segmentation \citep{RTFNet}, while newer methods extend this concept by using LLM-guided attention to achieve superior performance in safety-critical domains like autonomous driving, as explored in benchmarks like OpenRSS \citep{OpenRSS}. This fusion capability, central to modern systems like ConceptFusion, which creates open-set 3D maps from multiple sensors \citep{ConceptFusion}, promises to turn thermal cameras from simple heat sensors into powerful tools for semantic scene understanding. However, this advancement carries a critical safety caveat that must be addressed. The primary risk is that an LLM might ``hallucinate'' or confidently ``fill in'' details that are not actually present, potentially obscuring small or partially occluded real-world hazards that the low-resolution thermal sensor was incapable of detecting in the first place. This issue of reliably grounding language-based inference to sparse physical data remains a significant challenge, as noted in broader discussions on universal 3D scene \mbox{dialogue \citep{Chat3D}.} Furthermore, the reliability of these models is constrained by the current scarcity of large-scale, annotated thermal-language datasets needed for robust training and validation. Therefore, for any safety-critical application, it is imperative that the implementation of semantic super-resolution is balanced with rigorous uncertainty quantification and explicit fail-safe mechanisms to ensure that inferred semantic details never override the detection of critical safety cues from raw sensor data. Future research must prioritize the development of multimodal fusion architectures that not only incorporate LLM-driven insights but also explicitly model uncertainty and are validated against curated datasets designed for thermal-language alignment.}

\section{LLMs and 3D Vision Advancements and Applications}\label{sec:applications}

With the advent of advanced 3D data acquisition methods, the quality of 3D data has improved significantly compared to earlier, noisier and lower-quality datasets. Today, high-performance 3D sensors (see Table \ref{tab:combined_3d_sensors}) are available at increasingly affordable prices, making high-quality data collection accessible to a much broader audience. However, to fully harness the potential of this richer data, several challenges and bottlenecks still need to be addressed. This section highlights some of the key advancements that have been made in this area.

\subsection{Localization and Grounding}\label{sec:localization_and_grounding}

The foundational step in scene understanding for embodied agents is localization, followed by classification. After knowing where it is and what objects are present, a robot can begin to interpret its surroundings. This initial challenge has been extensively addressed, with numerous fast and accurate models now available for segmenting images (including RGBD and thermal data) and 3D point clouds from LiDAR sensors \citep{Yolo, PointNet, PointNet++, PointLLM}. Once objects are localized and classified, the next hurdle is holistic scene understanding, which goes beyond individual objects to the challenge of object grounding. Here, the robot must not only recognize objects but also connect linguistic descriptions to specific physical entities in its environment. This requires integrating language understanding with visual and spatial context to precisely identify the referenced object.

Figure \ref{fig:object_grounding_bifurcation} illustrates two main strategies for object grounding: pre-explored and self-exploring environments. The pre-explored approach leverages detailed prior information, such as annotated maps or datasets, to quickly and efficiently locate objects-making it ideal for static environments with minimal changes, as it offers predictable performance and lower computational demands. However, it can struggle to adapt when faced with dynamic or unfamiliar settings. In contrast, the self-exploration approach enables real-time adaptation and learning, using techniques like Simultaneous Localization and Mapping (SLAM) or reinforcement learning to navigate and understand complex, changing environments. While this method is more flexible and robust, it typically requires more computational resources and may be less accurate in highly unstructured scenes.

\begin{figure}
    \centering
    \includegraphics[width=0.97\linewidth]{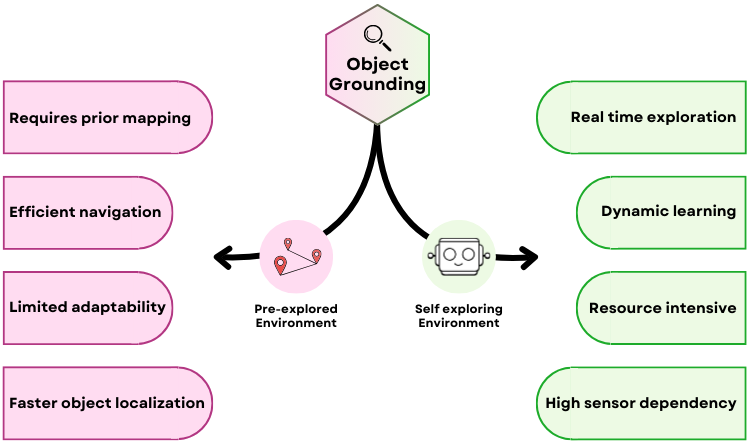}
    \caption{Bifurcation of Object Grounding Approaches: A visual comparison of pre-explored and self-exploring environment strategies, highlighting their unique characteristics, methodologies and application scenarios.}
    \label{fig:object_grounding_bifurcation}
\end{figure}

Enforcing long-term temporal consistency when fusing Simultaneous Localization and Mapping (SLAM) state with Large Language Model (LLM) memory is critical for robust 3D object grounding in dynamic environments. A primary strategy involves incremental memory updates, where the SLAM state, encompassing robot pose, map features and object locations, is periodically integrated into the LLM’s memory while reconciling environmental changes, such as loop closures. This fusion is often managed through probabilistic filtering techniques to handle uncertainty and drift, alongside cross-modal alignment, which synchronizes geometric SLAM data with the LLM's semantic embeddings. Advanced memory architectures, leveraging episodic memory or temporal attention mechanisms as explored in foundational works on models like Palm-e \citep{PALME} and CrossGLG \citep{CrossGLG}, enable coherent reasoning across extended time horizons. State-of-the-art frameworks such as ConceptFusion \citep{ConceptFusion} and OmniDrive \citep{OmniDrive} exemplify this by implementing semantic map fusion, which allows an LLM to perform temporally informed spatial queries on updated maps. While significant progress has been made, as also seen in efforts to empower LLMs with temporal understanding of point clouds \citep{PointLLM}, key challenges persist in managing the computational overhead and high-dimensional state representations for real-time performance, pointing towards future research in efficient memory distillation and continual learning paradigms.

Together, these methodologies provide a robust framework for tackling the diverse challenges of object grounding. Advances in this area allow AI systems to link natural language to specific objects and spatial locations, empowering robots to interpret, manipulate and interact with their environment based on verbal instructions. Recent research has introduced a range of innovative approaches. For example, Ref.\citep{RRExBoT} frames object grounding as an information retrieval problem, they fine-tune a pretrained VL backbone \mbox{(ViLBERT\citep{ViLBERT})} with 3D positional encodings for object proposals derived from RGBD observations. To handle the challenge of scoring hundreds of thousands of 3D region proposals, they introduce a “bag of tricks,” which includes: augmenting training with both positive and negative viewpoints, adding contextual region features, limiting exploration range using a path-length-based threshold and grouping region proposals by viewpoints for inference-time scoring. During inference, the agent performs local exploration using a frontier-based strategy, samples region proposals at each viewpoint and selects the highest-scoring match to the language query. Similarly, $\text{D}^3\text{Net}$ \citep{D3Net} system processes 3D point clouds using a detector (PointGroup \citep{Pointgroup}) to generate object proposals, which are then described by a speaker module that produces natural language captions. These captions are matched to objects via a transformer-based listener that localizes referred objects. Critically, the model is trained with a self-critical reinforcement learning objective that uses listener feedback (grounding accuracy and classification) to refine the speaker’s caption generation, encouraging more discriminative and spatially-aware descriptions as described by the architecture in the Figure \ref{fig:d3net_arch}. This integration allows the model to leverage partially annotated data, enabling semi-supervised learning. At the intersection of 3D vision and language, large language models are employed not only to generate coherent and context-sensitive captions but also to interpret and ground them within complex 3D scenes, enhancing both understanding and localization of objects in natural language-based robotic applications.

\begin{figure}
    \centering
    \includegraphics[width=0.99\linewidth]{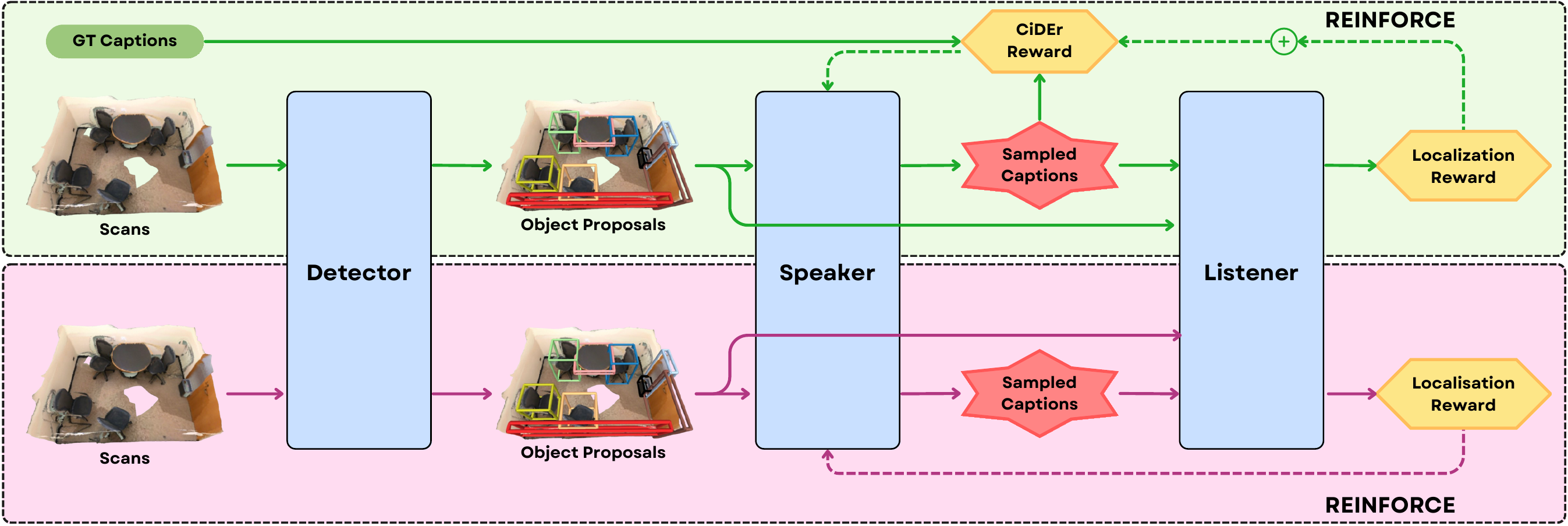}
    \caption{$\text{D}^3\text{Net}$ processes point clouds to generate object proposals, which the speaker captions. The listener matches captions to objects and REINFORCE propagates rewards from captioning and localization. It also supports end-to-end training without ground truth descriptions (bottom pink block).\citep{D3Net}.}
    \label{fig:d3net_arch}
\end{figure}

Building on these, the Transcrib3D \citep{Transcrib3D} methodology begins by applying a 3D object detector (Mask3D \citep{Mask3D}) to a colored point cloud, which is obtained via a LiDAR sensor. This process extracts each object's semantic category, position, size and color and then transcribes them into a structured textual scene description. To focus reasoning, the system filters this list to retain only objects relevant to the referring expression, reducing distraction and computational load. As can be seen in the Figure \ref{fig:transcrib3d_arch}, the core innovation lies in leveraging LLMs as reasoning engines: the filtered textual scene and the referring expression are combined into a prompt for the LLM, which iteratively generates and refines Python code to perform necessary spatial calculations, interprets the results and updates its reasoning until it identifies the correct object. This process is enhanced by principle-guided zero-shot prompting-embedding spatial heuristics into the LLM’s instructions- and a self-correction fine-tuning mechanism that enables smaller models to approach the performance of large ones. By using text as the bridge between visual perception and language reasoning, Transcrib3D sidesteps the need for massive multimodal datasets.

Hybrid grounding systems offer a compelling solution by balancing the reliability of pre-explored priors with the adaptability of real-time SLAM-based exploration, addressing the limitations inherent in using either approach alone \citep{LookNoDeeper}. These systems create a synergistic framework by utilizing static, pre-annotated maps to constrain the SLAM search space, thereby enhancing localization efficiency and reducing odometric drift. Simultaneously, they integrate live SLAM data, allowing the robot to adapt to dynamic or unmapped environmental changes. The integration of Large Language Models (LLMs) further empowers these hybrid systems by providing advanced contextual reasoning; LLMs can interpret and reconcile information from both the static priors and dynamic sensory inputs to achieve more robust semantic grounding and nuanced spatial understanding, as demonstrated by recent methodologies like Transcrib3D and $\text{D}^3\text{Net}$ \citep{Transcrib3D, D3Net}. Looking forward, a key research direction involves developing adaptive architectures that dynamically weigh the influence of prior knowledge against real-time observations, potentially guided by LLM-driven logic, to optimize computational efficiency and grounding accuracy for next-generation robotic perception in diverse and evolving environments.

\begin{figure}
    \centering
    \includegraphics[width=0.99\linewidth]{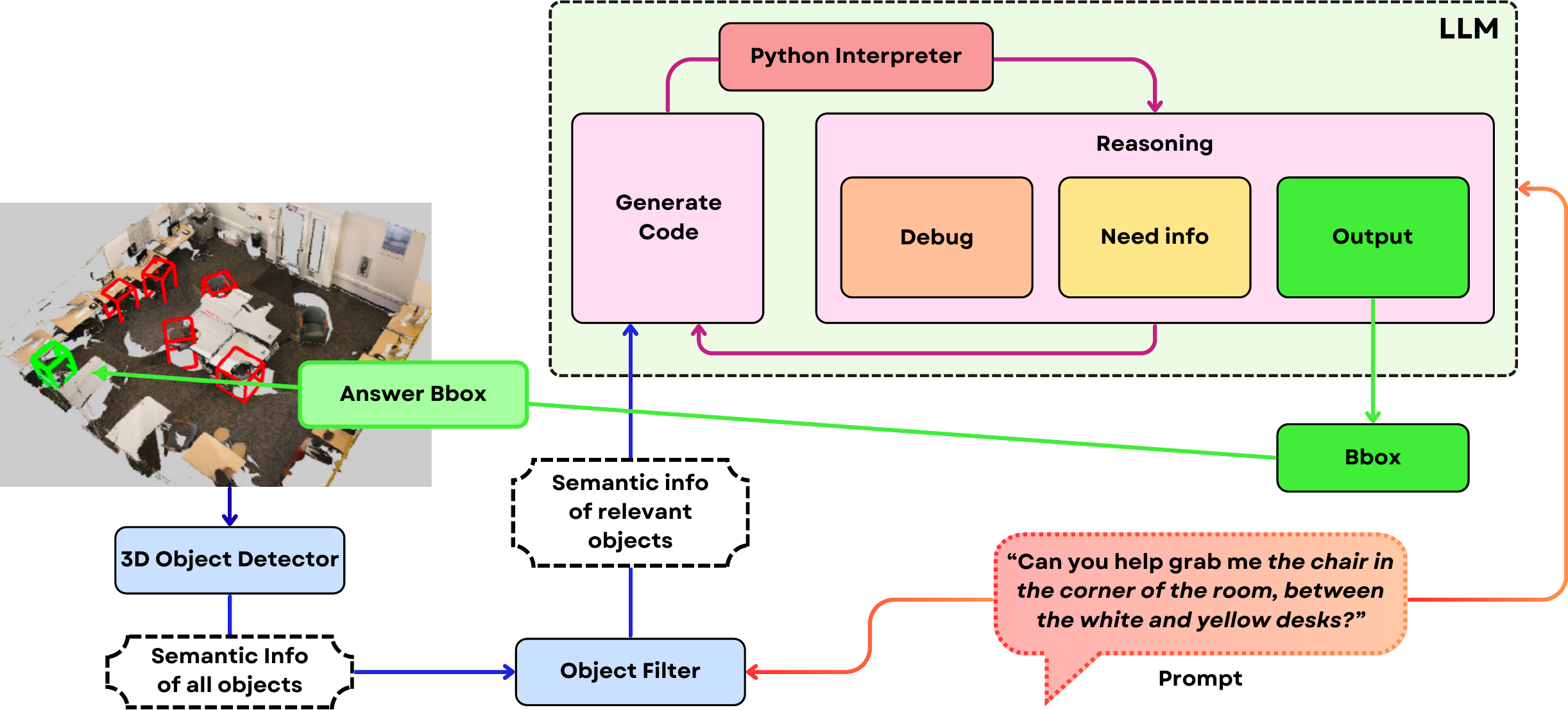}
    \caption{Transcrib3D converts 3D spatial data into text to enable object grounding through language models, using code generation and reasoning. It supports zero-shot and fine-tuned modes \citep{Transcrib3D}.}
    \label{fig:transcrib3d_arch}
\end{figure}

\subsection{Dynamic Scenes}\label{sec:dynamic_scenes}

As robots move from static to dynamic scene understanding, a key challenge is enabling them to interpret and predict changes in their environment over time-especially human motion. This capability is critical for robots that will assist people in homes or factories, as it allows them to safely and effectively respond to human activities. Human-centric dynamic scene understanding involves not only recognizing and categorizing movements, but also interpreting interactions within a shared space \citep{PointLightMotion, CrossGLG, SceneVerse, Agent3DZero, ConceptGraphs}. Research in neuroscience has shown that humans recognize motion by focusing on a few key body joints, a principle adopted in robotics through point-light motion capture systems. These systems use reflective markers on major joints and infrared cameras to triangulate 3D positions over time, producing stick-figure animations that capture the essence of movement \citep{PointLightMotion}. Commercial systems like Vicon and OptiTrack are widely used for \mbox{this purpose.}

Inspired by these biological and technological insights, early robotics approaches relied on low-level joint data for action recognition. However, more advanced methods now integrate high-level semantic information from large language models (LLMs) to improve action understanding. For instance, CrossGLG \citep{CrossGLG} centers around leveraging LLMs to enrich vision-based one-shot 3D action recognition. It's dual-branch framework combines high-level semantic guidance from LLM-generated textual descriptions with low-level skeletal motion data. During training, global textual prompts are used to identify key joints relevant to an action (global-to-local) and then joint-level textual descriptions interact with joint features to form a holistic action representation (local-to-global) as can be understood from Figure \ref{fig:cross_glg_arch}. This cross-modal guidance enables the skeleton encoder to focus on semantically critical motion cues, improving generalization. Importantly, LLMs act as semantic translators, injecting human-level understanding into the vision model without being needed at inference time. This is particularly important in applications like autonomous vehicles, where predicting human actions, such as a pedestrian’s next move and that too in real time, is vital for safety.

\begin{figure}
    \centering
    \includegraphics[width=0.99\linewidth]{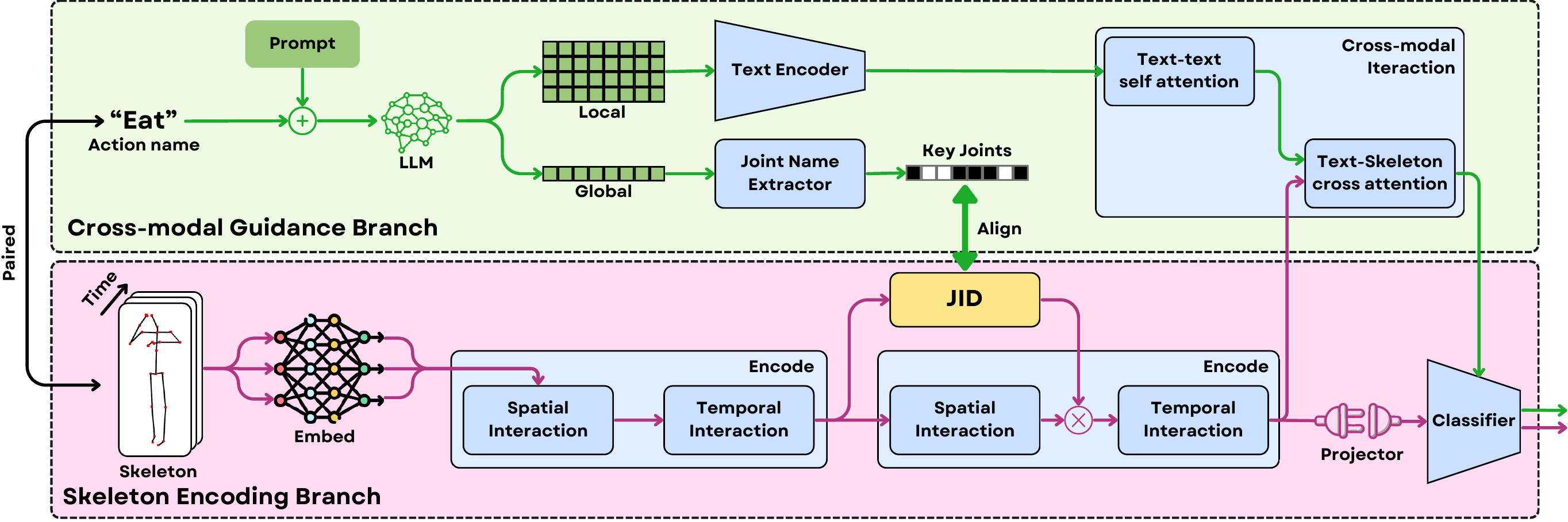}
    \caption{CrossGLG includes a Joint Importance Discrimination (JID) module to assess joint importance from skeleton features. A cross-modal guidance branch aids skeleton feature learning, but during novel class inference, only the skeleton encoding branch is used without text \citep{CrossGLG}.}
    \label{fig:cross_glg_arch}
\end{figure}

Recent advances have extended beyond action recognition to encompass general scene understanding in dynamic environments. The Grounded Pre-training for Scenes (GPS) framework demonstrates that large-scale, open-vocabulary datasets can significantly improve the generalization of 3D vision-language models, achieving state-of-the-art results in 3D scene understanding tasks \citep{SceneVerse}. Pushing this further, Agent3D-Zero \citep{Agent3DZero} method mimics human-like scene comprehension by analyzing images captured from strategically selected viewpoints. The process starts with a bird’s-eye view (BEV) image enhanced by the Set-of-Line Prompting (SoLP) technique, which overlays grid lines and coordinates to guide the VLM in selecting informative camera angles. Crucially, the VLM serves not just as a passive interpreter but as an active agent, planning which views to observe and reasoning over them using its embedded world knowledge. This intersection of vision and language allows Agent3D-Zero to bypass the need for extensive 3D annotations.

\textls[-15]{To address the complexity of real-world environments, ConceptGraphs \citep{ConceptGraphs} constructs a dynamic, scene graph as the robot explores. It first employs class-agnostic segmentation and vision-language foundation models to extract and fuse object-level features from RGB-D images into a compact 3D scene graph. Each object node is semantically enriched using LVLMs (e.g., LLaVA \citep{LLaVA}) for multi-view captioning, which is then refined using LLMs \mbox{(e.g., GPT-4 \citep{achiam2023gpt})} to provide coherent object descriptions. Spatial relationships between objects are inferred through LLMs by reasoning over object captions and 3D geometry, resulting in semantically meaningful graph edges. This structured representation enables natural language interaction with robots, allowing them to interpret complex \mbox{open-ended queries.}}

Altogether, these developments are equipping embodied robots with the tools needed to understand and operate in dynamic, human-centered environments-enabling safer, more adaptive and more intelligent robotic assistance.

\subsection{Indoor and Outdoor Scene Understanding}\label{sec:indoor_and_outdoor_scene_understanding}

One notable advantage of indoor environments is their clear division into purpose-defined rooms, which allows robots to significantly narrow down their search space and focus their computational resources. For example, if a robot is asked to fetch a toothbrush and finds itself in the living room, it can infer that the probability of locating the toothbrush there is extremely low. Instead of searching the entire living room, the robot can prioritize moving toward the bathroom, where the item is most likely to be found. This targeted approach reduces unnecessary computation. To facilitate this, methods like \mbox{QueSTMaps \citep{QueSTMaps}} have been developed to generate queryable topological and semantic representations of indoor 3D environments. The approach begins by reconstructing a 3D point cloud from posed RGB-D images and introduces a novel multi-channel occupancy representation, combining top-down density, floor and ceiling occupancy maps, to enhance room and transition region segmentation via a Mask R-CNN \citep{MaskRCNN} instance segmentation network. For semantic labeling, object instances within each segmented room are identified using a 3D semantic mapping pipeline, where each object is embedded with CLIP features. These object-level CLIP embeddings are aggregated for each room and then processed through a self-attention transformer, aligning the room representations with natural language descriptions of the rooms. This alignment empowers the system to interpret and respond to open-ended, room-level queries, leveraging the reasoning and open-vocabulary capabilities of LLMs to bridge visual scene understanding and flexible language-based interaction, achieving superior performance in room segmentation and classification on datasets such as \mbox{Matterport3D \citep{Matterport3D}.} Such scene classification and semantic mapping approaches are crucial for efficient indoor robot operation, as they allow robots to leverage contextual cues and object-level information to infer room categories and functional areas \citep{Matterport3D}.

Transitioning from indoor to outdoor environments introduces a host of new challenges. Outdoor spaces are generally less structured, combining natural and artificial elements and are subject to widely varying lighting conditions due to weather, time of day and shadows. Unlike the controlled lighting and confined spaces of indoor settings, outdoor environments are vast and dynamic, requiring robots to use long-range sensors such as LiDAR and GPS. The complexity and scale of outdoor scenes demand robust solutions for real-time understanding and localization. The task of 3D Visual Grounding in the Wild (3DVGW), as introduced in \citep{WildRefer}. Their approach addresses the challenge of grounding natural language descriptions in real-world 3D scenes by utilizing multimodal inputs, namely, synchronized LiDAR point clouds, RGB images and free-form linguistic queries. To bridge the gap between these heterogeneous modalities, the authors design a Triple-modal Feature Interaction (TFI) module that fuses semantic information from a pretrained RoBERTa language model, geometric features from point clouds (via \mbox{PointNet++ \citep{PointNet++}),} and visual features from images (via ResNet34 \citep{Resnet}). Moreover, to handle temporal dynamics, crucial in human-centric, changing scenes, they introduce a Dynamic Visual Encoder (DVE) employing a transformer-based attention mechanism to track motion across frames. This enables the model to align motion-related textual cues (e.g., ``a boy running with a red balloon'') with dynamic visual evidence. The fused representation is passed through a DETR-style transformer decoder \citep{DETR} to localize the target object directly in 3D space. By integrating language-guided attention into a vision pipeline, WildRefer demonstrates how LLMs can semantically anchor object detection in dynamic 3D environments, paving the way for intelligent human-robot interaction in open-world settings.

Other key subdomains of general scene understanding includes (a) Shape \linebreak \mbox{Correspondence \citep{ZeroShot3DShapeCorrespondence},} enabling alignment and matching of 3D models, (b) Outdoor Scene Understanding \citep{WildRefer}, focused on object and context recognition in open environments, \mbox{(c) Scene} Question Answering \citep{ScanQA}, which facilitates contextual comprehension of visual scenes through question answering, (d) Action Recognition \citep{SkeletonBasedActionRecognition}, identifying human actions using spatial-temporal cues, (e) Indoor Scene Understanding \citep{OpenScene}, capturing room layouts and object placements and (f) 3D Object Classification \citep{PointLLM}, categorizing objects from geometric data.

\subsection{Open Vocabulary Understanding and Pretraining}\label{sec:open_vocabulary_understanding_and_pretraining}

Despite significant progress, a persistent challenge in robotics is that most models are trained in a fully supervised manner on fixed datasets. This approach restricts their ability to generalize to unseen objects or new environments, often requiring fine-tuning or even retraining for each novel scenario. Collecting new datasets for every unfamiliar object or context is not only labor-intensive but also limits scalability. To address these issues, the field is moving toward open-vocabulary understanding, which includes both open-vocabulary grounding and classification, enabling robots to handle a broader range of queries and objects without explicit retraining \citep{VisualProgramming}.

Visual Programming \citep{VisualProgramming} introduces an innovative solution by leveraging large language models (LLMs) to generate structured programs for open-vocabulary object localization. Instead of relying on traditional supervised learning with extensive object-text annotations, the authors employ LLMs to interpret natural language queries and generate modular visual programs that decompose complex spatial reasoning tasks. These visual programs, comprising view-independent, view-dependent and functional modules, are automatically translated into executable Python code that interacts with 3D scene data (such as point clouds and detected objects). The LLMs are crucial for parsing free-form descriptions (Figure \ref{fig:visual_programming_arch}), identifying relevant objects and spatial relations and orchestrating the logic needed for localization. A key technical innovation is the language-object correlation (LOC) module, which fuses geometric features from 3D point clouds with appearance cues from 2D images, enabling open-vocabulary object recognition and precise grounding. This approach leverages LLMs' reasoning and planning abilities to bridge the gap between language and 3D vision, allowing the system to generalize to unseen categories and spatial relations without additional training or annotations.

\begin{figure}
    \centering
    \includegraphics[width=0.9\linewidth]{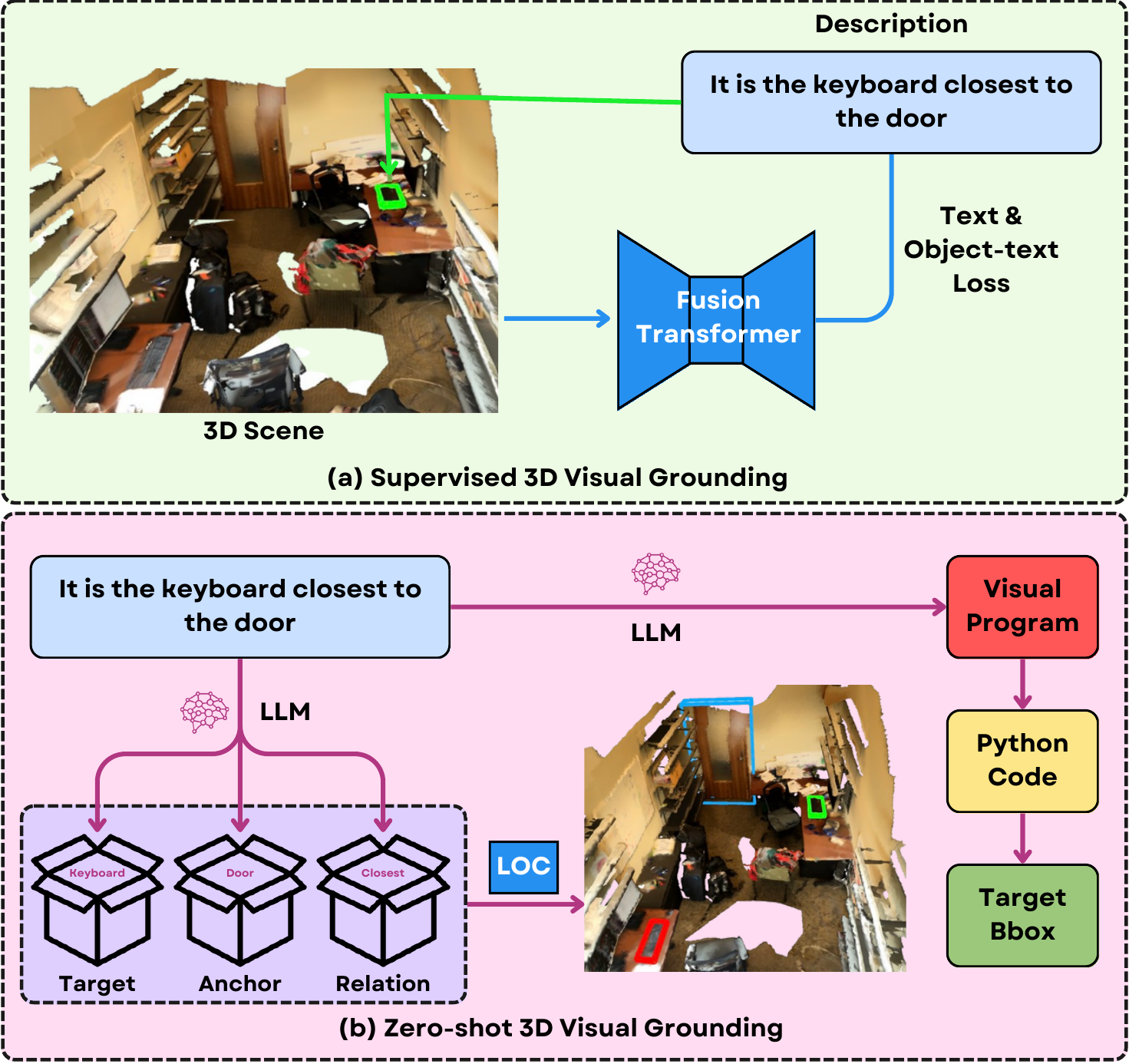}
    \caption{(\textbf{a}) Supervised 3DVG uses 3D scans and text queries with object-text annotations. \mbox{(\textbf{b}) Zero-shot} 3DVG relies on LLM-generated programs (target, anchor, relation) to locate objects, excelling at spatial reasoning without annotations \citep{VisualProgramming}.}
    \label{fig:visual_programming_arch}
\end{figure}

A notable approach in this space is Context-Aware Alignment and Mutual Masking for 3D-Language Pre-Training, which addresses challenges such as sparse point clouds and relational ambiguities. Its Context-Aware Spatial-Semantic Alignment (CSA) aligns spatial relationships in 3D with textual semantics to reduce ambiguity, while Mutual 3D-Language Masked Modeling (M3LM) jointly masks 3D proposals and language tokens, boosting cross-modal learning. This framework has demonstrated strong generalization across tasks such as 3D visual grounding and question answering, consistently outperforming baselines on datasets like ScanRefer \citep{ScanRefer} and ScanQA \citep{ScanQA,ContextAwareAlignmentAndMutualMaskingFor3DLanguagePreTraining}.

Similarly, ULIP \citep{ULIP} aims to learn a unified representation for images, text and 3D point cloud data. It constructs triplets from CAD model meshes, using randomized surface sampling for point clouds, multi-view imaging and text annotations for descriptions. Through contrastive learning, the 3D encoder is trained to align with the CLIP \citep{CLIP} feature space, making ULIP agnostic to the underlying 3D backbone architecture (e.g., \citep{PointNet, PointNet++, PointNext}). Pre-training on large-scale datasets like ShapeNet55 allows ULIP to achieve state-of-the-art results in both standard and zero-shot 3D classification tasks, as demonstrated on benchmarks such as ModelNet40 \citep{3DShapeNets} and ScanObjectNN \citep{ScanObjectNN}.

{Large Language Models (LLMs) effectively handle ambiguous open-vocabulary segmentation queries and mitigate semantic drift by integrating advanced reasoning with spatial-semantic constraints and programmatic decomposition. To overcome the broad interpretations and potential for drift inherent in underspecified language, frameworks such as Transcrib3D \citep{Transcrib3D} convert 3D spatial data into structured text, allowing the LLM to iteratively generate and refine executable code guided by spatial heuristics that constrain the query's scope and prevent semantic deviation. Similarly, visual programming approaches translate ambiguous natural language into modular, executable programs that precisely aggregate spatial and semantic cues, ensuring robust object grounding \citep{VisualProgramming}. This capability is further strengthened by context-aware alignment models like CSA and M3LM, which anchor textual descriptions to 3D spatial relations, thereby reducing relational ambiguity and improving semantic consistency across modalities \citep{ContextAwareAlignmentAndMutualMaskingFor3DLanguagePreTraining}. The fusion of language with multimodal data, demonstrated in works like $\text{D}^3\text{Net}$ and WildRefer and the use of structured dynamic scene understanding, as seen in ConceptGraphs, also play a crucial role by using established spatial relationships to disambiguate queries \citep{D3Net, WildRefer, ConceptGraphs}. Collectively, these methods showcase how LLMs, when combined with programmatic reasoning and tight multimodal integration, can maintain semantic fidelity and perform accurate segmentation without drifting towards irrelevant interpretations.}

Collectively, these open-vocabulary approaches are paving the way for more flexible, adaptive and scalable robotic systems that can operate in diverse, real-world settings without requiring exhaustive retraining or manual data collection.

\subsection{Text to 3D}\label{sec:text_to_3d}

Most of the models discussed so far rely on fully supervised training, which demands large annotated datasets to achieve reasonable performance. This becomes especially challenging in the context of 3D scene understanding, where the diversity of possible outputs, ranging from classification and localization to planning, makes comprehensive data collection both time-consuming and expensive. Unlike 2D computer vision, where abundant image data is readily available online, 3D datasets are scarce and difficult to generate, even with synthetic modeling, which itself is a slow and resource-intensive process. To address this bottleneck, data generation models, particularly those leveraging text-to-3D technology, have emerged as a promising solution. These models can synthesize detailed and semantically rich 3D environments from natural language descriptions, thus providing valuable resources for training and evaluating robotic vision systems.

Text-to-3D models originally developed for applications in movies and games, are now being adapted for 3D computer vision research. For instance, DreamFusion \citep{DreamFusion} utilizes Score Distillation Sampling (SDS) to optimize Neural Radiance Fields (NeRFs) \citep{NeRF} using pretrained 2D diffusion models. In this process, a NeRF is randomly initialized and iteratively refined using multiview images under the guidance of a 2D diffusion model, resulting in a 3D representation that can render the object from arbitrary viewpoints or be exported as a 3D model. While this approach effectively bridges the gap between 2D and 3D data without requiring large-scale 3D datasets and offers high-fidelity reconstructions ideal for detailed scene understanding, it has limitations. Specifically, DreamFusion requires training a separate NeRF for each object or scene and these extensive per-scene training requirements limit its generalizability and real-time applicability in dynamic robotic environments. In contrast to NeRFs, other methodologies present distinct advantages for robotics. Gaussian splatting, seen in models like GALA3D \citep{GALA3D}, provides a compelling balance between rendering quality and computational efficiency, which is critical for rapid scene synthesis. Diffusion models, such as GPT4Point \citep{GPT4Point}, excel at generating diverse 3D point clouds with fine-grained control, benefiting robotic manipulation and object recognition. However, procedural generation, exemplified by 3D-GPT \citep{3DGPT}, which uses large language models to programmatically create and edit 3D content, currently represents a highly relevant paradigm for robotics due to its unparalleled controllability, interpretability and seamless integration with existing robotic simulation and planning toolchains. Thus, while Gaussian splatting and diffusion models show promise for real-time applications and NeRFs remain valuable for offline modeling, the procedural approach's alignment with task-specific, iterative environment modeling makes it exceptionally suitable for current robotic systems.

Other approaches, such as GPT4Point \citep{GPT4Point}, directly align features from text to 3D point cloud space. GPT4Point employs a two-stage architecture: first, it uses the QFormer \citep{QFormer} framework for feature alignment, then applies point diffusion models for controllable text-to-3D generation. While DreamFusion excels at realistic texture rendering, GPT4Point is particularly well-suited for tasks that require precise point-cloud recognition.

Beyond object-level generation, some models focus on designing entire 3D scenes from natural language \citep{Chat2Layout, GALA3D, Uni3DLLM}, with interactive capabilities that allow users to iteratively modify the scene during generation \citep{Chat2Layout}. For example, 3D-GPT \citep{3DGPT} decomposes the complex task of 3D scene synthesis into a planning–reasoning–execution pipeline powered by LLMs. Three specialized agents, task dispatch, conceptualization and modeling, work in unison (Figure \ref{fig:3dgpt_arch}): the task dispatch agent identifies the relevant procedural functions needed based on user instructions; the conceptualization agent enriches vague scene descriptions into detailed visual specifications; and the modeling agent translates these into executable Python code for Blender using the Infinigen library. Rather than generating raw 3D geometry directly, LLMs are tasked with understanding the semantics of user input and leveraging rule-based procedural generation, which enables fine-grained control over parameters such as shape, texture, lighting and animation. This synergy between language understanding and vision tool manipulation demonstrates how LLMs can act as intelligent controllers within traditional 3D software ecosystems.

The interplay between methodologies, such as NeRFs in DreamFusion \citep{DreamFusion}, Gaussian splatting in GALA3D \citep{GALA3D} and procedural generation in 3D-GPT \citep{3DGPT}, demonstrates the rapid advancement and increasing sophistication of text-to-3D research. Looking ahead, future work will likely focus on improving model resolution and efficiency, leveraging higher-capacity language models and exploring new applications in interactive robotics and augmented reality.

\begin{figure}
    \centering
    \includegraphics[width=0.7\linewidth]{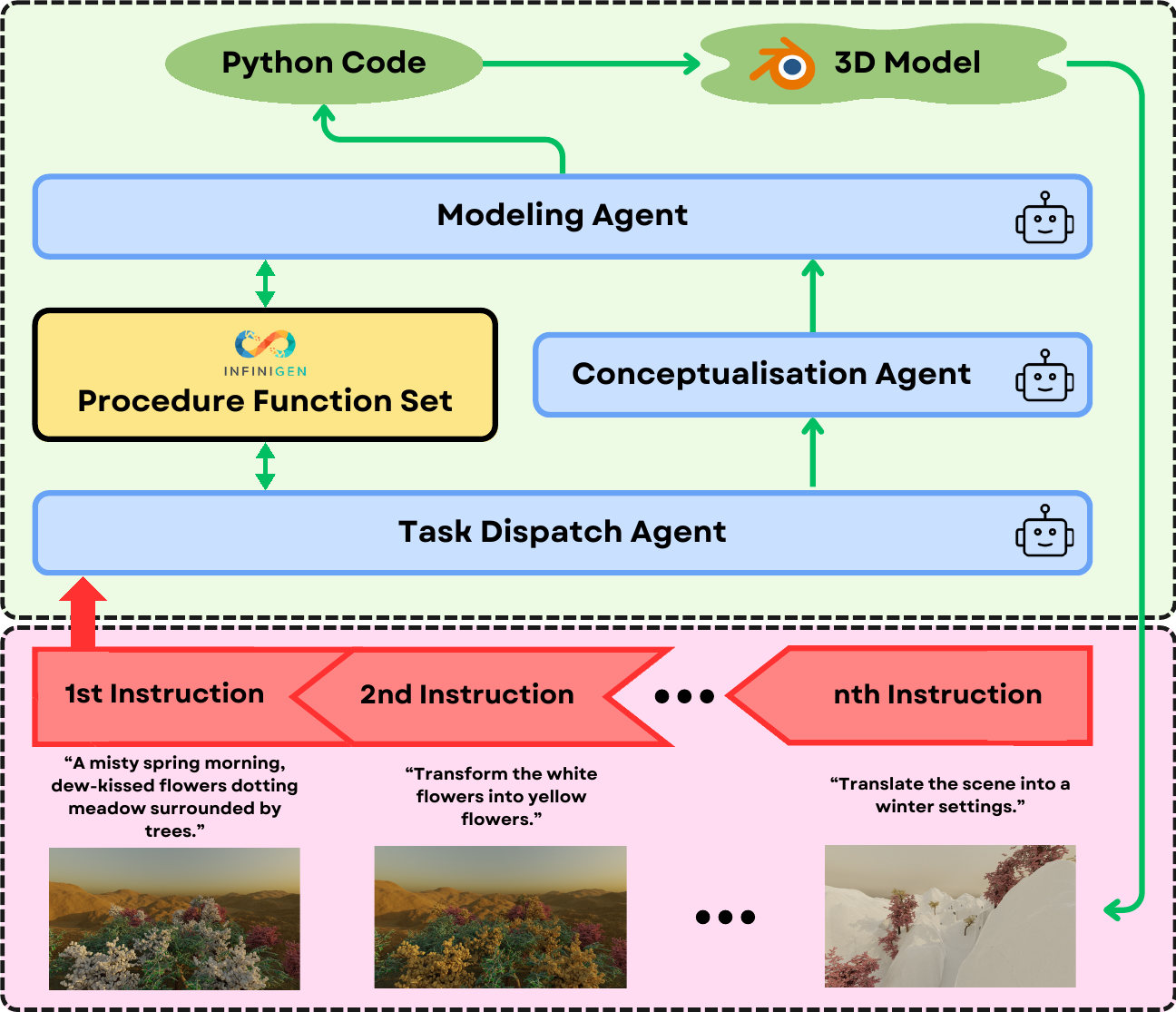}
    \caption{3D-GPT uses large language models (LLMs) within a multi-agent framework, where three cooperative agents work together to generate 3D content procedurally. These agents reference documentation from the procedural generator, determine the necessary function parameters and generate Python code. This code interacts with Blender's API to create and render 3D models \citep{3DGPT}.}
    \label{fig:3dgpt_arch}
\end{figure}

\subsection{Multimodality}\label{sec:multimodality}

Why limit ourselves to just vision, when humans naturally rely on a rich combination of senses, like touch, temperature and sound, to fully understand their surroundings? The same philosophy is now being adopted in robotics, leading to the development of multimodal systems that integrate multiple types of sensory input for a more robust and adaptable perception of the world \citep{LiDARLLM, PointLLM, Chat3D, 3DMIT, ConceptFusion}. By combining data from different modalities, robots can overcome the limitations of vision alone and extend their capabilities far beyond simply “seeing” a scene.

For example, in foggy conditions where RGBD cameras struggle, thermal imaging (using infrared rays) can penetrate the fog and provide a clear thermogram, allowing the robot to perceive its environment when vision fails. Similarly, if a robot is asked to fetch hot food, thermal vision can help it distinguish hot items from cold ones-something that would be difficult with vision alone. There are numerous scenarios where alternative sensory data, such as tactile, auditory or thermal input, can significantly impact perception and decision-making.

As shown in Figure \ref{fig:general_architecture_of_mllm} Modern multimodal approaches typically use dedicated encoders for each input type (e.g., LiDAR, RGB, thermal, tactile), followed by adapters that map the encoded data into a shared latent space for reasoning within a large language model (LLM) \citep{PointLLM, Chat3D}. LiDAR-LLM, for instance, uses a view-aware transformer and \mbox{a three-stage} training process to make raw LiDAR data more interpretable, which is particularly valuable for autonomous vehicles and robotic navigation \citep{LiDARLLM}. Alternatively, 3DMIT bypasses explicit alignment by directly embedding 3D scene information into language prompts, enabling multi-task scene understanding-such, such as 3D visual question answering, captioning and grounding, without the need for complex cross-modal mapping. This direct approach has shown strong performance, even surpassing models that require explicit 3D-language alignment \citep{3DMIT}.

\begin{figure}
    \centering
    \includegraphics[width=0.99\textwidth]{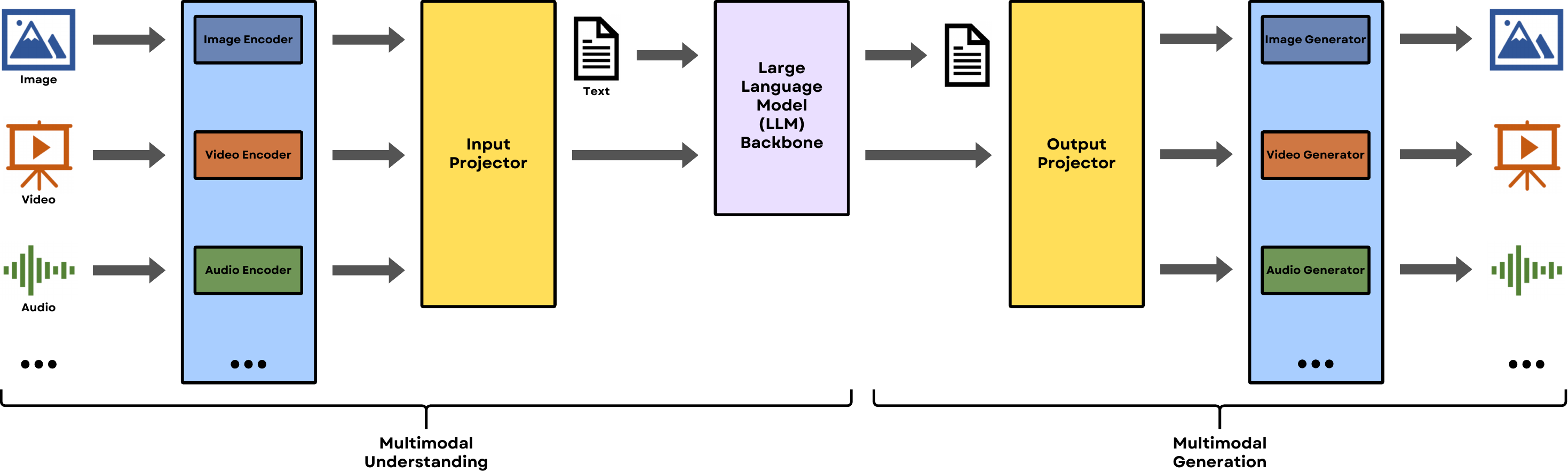}
    \caption{General Architecture of an MLLM: A framework where inputs like images, video, audio and 3D point clouds are processed by specialized encoders (Visual: \citep{CLIP, ViT, BEiT, Vcoder}, \mbox{Audio: \citep{CFormer, HuBERT, Whisper, CLAP},} 3D: \citep{PointBERT, ULIP}). These encoded features are aligned into a unified latent space via projectors \mbox{(e.g., MLP,} Cross-attention) and then fed into an LLM backbone (e.g., \citep{vicuna2023, touvron2023llama}). This enables multimodal understanding and generation tasks like image synthesis or cross-modal reasoning through its \mbox{input/output projectors.}}
    \label{fig:general_architecture_of_mllm}
\end{figure}

ConceptFusion \citep{ConceptFusion} presents a novel methodology that tightly integrates large vision-language foundation models (like CLIP \citep{CLIP}, DINO \citep{DINO} and AudioCLIP \citep{AudioCLIP}) with dense 3D mapping systems to build open-set, multimodal 3D scene representations. At its core, the method fuses pixel-aligned semantic features, extracted in a zero-shot fashion from pre-trained foundation models, directly into 3D maps constructed using traditional SLAM techniques. These maps can be queried across modalities, text, images, audio and clicks, enabling powerful open-set recognition and spatial reasoning. To bridge the gap between global (image-level) and local (region-level) context, ConceptFusion introduces a feature fusion mechanism that combines CLIP’s global embeddings with region-specific embeddings derived from class-agnostic object proposals. This enables robust localization of long-tailed and fine-grained concepts without any finetuning. Furthermore, LLMs are employed not just for parsing complex natural language queries but also for composing symbolic 3D spatial functions (e.g., “how far is X from Y?”) over the semantic map, demonstrating a powerful intersection of LLMs and 3D visual perception for downstream embodied reasoning and robotics tasks.

{While Multimodal Large Language Models (MLLMs) integrated with diverse sensory inputs hold promise for robust robotic perception, their resilience under adversarial sensor corruption presents a critical challenge for safe deployment. Phenomena such as LiDAR spoofing, where malicious actors inject false spatiotemporal signals and audio jamming, which disrupts environmental sound recognition, can severely degrade the reliability of perception modules feeding into the MLLM's reasoning pipeline. Although multimodal fusion theoretically offers robustness through cross-modal consistency checks, empirical vulnerabilities persist, driving research into countermeasures like adversarial training and data augmentation. For example, recent frameworks such as LiDAR-LLM \citep{LiDARLLM} show improved noise tolerance through advanced transformer-based encoding, while systems like ConceptFusion \citep{ConceptFusion} and OmniDrive \citep{OmniDrive} leverage sensor redundancy to detect inconsistent signals. However, progress is hindered by significant open challenges, including the scarcity of annotated adversarial datasets, the difficulty of real-time anomaly detection in complex sensor streams and the opaque reasoning pathways of LLMs under corrupted conditions. Therefore, promising future research trajectories involve creating adaptive fusion strategies that dynamically weigh sensor trustworthiness, developing explainable anomaly detection mechanisms and establishing standardized benchmarks for evaluating the adversarial robustness of embodied agents.}

Ultimately, integrating multiple sensory modalities-much like humans do-enables robots to operate more reliably in complex, dynamic environments, adapting to challenges that would confound single-modality systems. This multimodal perception is key for the next generation of intelligent, human-assistive robots.

\subsection{Embodied Agent}\label{sec:embodied_agent}

In recent years, the integration of large language models (LLMs) with 3D vision systems has driven remarkable progress in robotics, especially for tasks related to perception, manipulation and navigation \citep{3DLLM, LearningFromUnlabeled3DEnvironments}. Figure \ref{fig:robotic_perception_pipeline} describes the basic pipeline for all embodied agent robots which perceive their environment, process the information and take action based on that. By harnessing the reasoning and language understanding abilities of LLMs, robots can now interpret complex instructions, recognize and localize objects and perform sophisticated actions within dynamic, real-world 3D \mbox{environments \citep{3DLLM,ConceptGraphs,OmniDrive}.} This fusion of technologies has led to the development of fully autonomous embodied agents-intelligent systems with a physical or simulated body that can perceive, interact with and act upon their environment \citep{FrameMining,ToolFlowNet,DexPoint}. Unlike traditional software agents that operate solely in digital domains, embodied agents exist in spatial worlds, integrating sensors and actuators to perceive and manipulate their surroundings \citep{ConceptGraphs,OmniDrive}. In observed environments, object localization is often simplified to selecting from a list of detected objects and the ability of these systems to generalize is rigorously tested in unobserved \mbox{settings \citep{LearningFromUnlabeled3DEnvironments}.}. Autonomous vehicles are a standout application area, where the integration of LLMs and 3D vision has enabled significant advances in perception and decision-making \citep{OmniDrive}.

\begin{figure}
    \centering
    \includegraphics[width=0.7\linewidth]{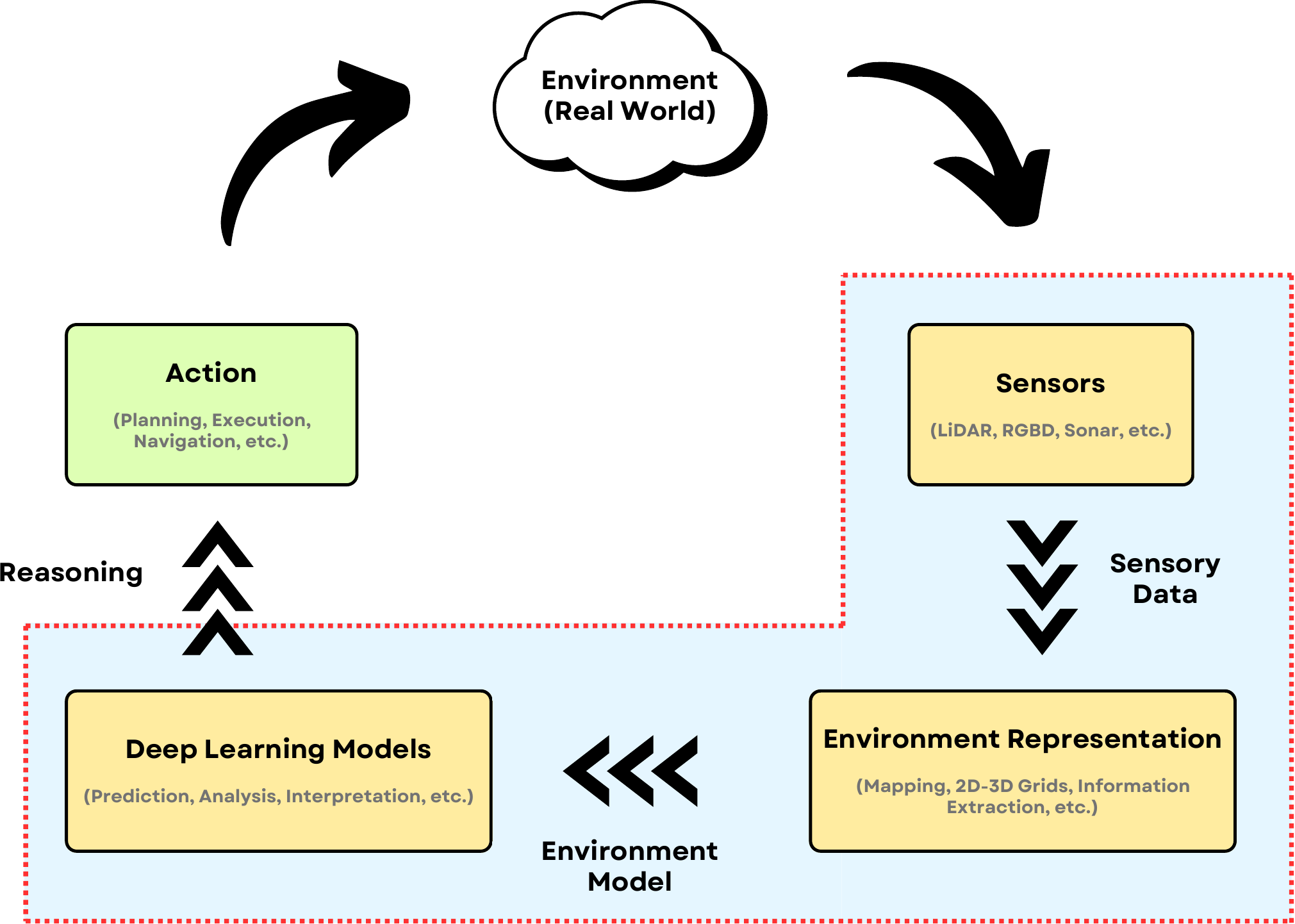}
    \caption{Generic Robotic Perception Pipeline: A typical perception loop where a robot gathers sensor data from its environment, which is then fused into a structured representation (e.g., a 3D model or panoptic map). Deep learning models process this representation to gain environmental understanding and make decisions, creating a continuous cycle that repeats until the task is complete.}
    \label{fig:robotic_perception_pipeline}
\end{figure}

A notable example is PolarNet \citep{PolarNet}, which advances language-guided robotic manipulation by leveraging point cloud representations to encode 3D spatial information more effectively than traditional 2D image-based methods \citep{GATO,PALME,HiveFormer,BCZ,AutoLambda}. PolarNet reconstructs point clouds from multi-view RGB-D images and utilizes the Open3D toolkit for downsampling and filtering. The system encodes point clouds using PointNext \citep{PointNext} and processes language instructions with a CLIP encoder \citep{CLIP}. These representations are then fused using self-attention and cross-attention networks \citep{AttentionIsAllYouNeed} and decoded through an MLP to predict robot actions in 7 degrees of freedom (7-DoF): Cartesian coordinates $\mathbf{a}_t^{xyz} \in \mathbb{R}^3$, rotations as quaternions $\mathbf{a}_t^q \in \mathbb{R}^4$ and gripper open/close states $\mathbf{a}_t^o \in {0, 1}$. This comprehensive action space enables robots to execute complex manipulation tasks with high precision. The generalization ability of PolarNet is further supported by studies on point cloud normalization and sim-to-real transfer \citep{FrameMining,ToolFlowNet,DexPoint}. Meanwhile, works such as \mbox{ConceptGraphs \citep{ConceptGraphs}} and OmniDrive \citep{OmniDrive} highlight how LLMs can enhance both reasoning and perception, with ConceptGraphs focusing on object manipulation and OmniDrive on navigation and decision-making in autonomous driving. Datasets like HM3D-AutoVLN \citep{LearningFromUnlabeled3DEnvironments} and systems like PolarNet \citep{PolarNet} underscore the importance of large-scale data generation and 3D spatial encoding for tackling challenges of scalability and precision in embodied AI.

While integrating 3D data and LLMs has advanced autonomous driving, real-world outdoor driving presents unique challenges that go beyond simply perceiving proximity and making reactive decisions. To safely navigate, vehicles often need a broader, long-range view to anticipate and respond to distant events, but this introduces new issues such as occlusions, fog and smog, which can severely limit the effectiveness of conventional vision and depth sensors \citep{OpenRSS,RTFNet}. To overcome these limitations, sensor fusion-combining multiple sensing modalities-becomes essential.

Thermal imaging, for example, is increasingly recognized as a critical component in autonomous driving systems. Unlike visible light, infrared radiation has a longer wavelength that allows it to penetrate fog, smog and darkness, providing clear detection of objects, pedestrians and animals even in low-visibility conditions \citep{OpenRSS,RTFNet}. This capability greatly enhances safety and situational awareness, as thermal sensors can reliably identify hazards that RGB cameras might miss \citep{OpenRSS,RTFNet}. Recent models such as \mbox{OpenRSS \citep{OpenRSS}} and \mbox{RTFNet \citep{RTFNet}} demonstrate the effectiveness of fusing RGB and thermal data, achieving robust semantic segmentation and object detection in challenging environments by leveraging lightweight cross-modal fusion techniques and efficient encoder-decoder architectures.

Beyond visual data, tactile sensing is gaining traction for its ability to provide vehicles with a ``sense of touch.'' Tactile sensors can detect differences in road texture, surface grip and even tire conditions, offering insights that vision alone cannot provide \citep{MultiPLY}. This information is particularly valuable when visually similar surfaces have vastly different physical properties, such as wet versus dry pavement. By integrating tactile data, autonomous vehicles can better assess road safety and adapt their driving strategies accordingly \citep{MultiPLY}.

Audio data also plays a complementary role, especially when objects are occluded from view. Acoustic signals can help identify activities or hazards, such as approaching emergency vehicles or construction, enabling vehicles to react even without direct visual confirmation \citep{MultiPLY}. Similarly, RGB-D images, while useful, are highly dependent on ambient lighting; thermal and other non-visual modalities can fill in these gaps during nighttime or adverse weather \citep{OpenRSS}.

Recent research has explored aligning 3D point cloud representations with natural language and some models even use 2D images as a bridge for this \mbox{alignment \citep{CLIP2Point,PointBindPointLLM,JM3D}.} However, the integration of additional modalities is proving vital for addressing the limitations of vision-only systems. For instance, OpenRSS \citep{OpenRSS} introduces a Thermal Information Prompt to inject thermal data into vision encoders, while RTFNet \citep{RTFNet} employs dual encoders for RGB and thermal fusion, achieving state-of-the-art segmentation in urban driving scenarios.

The MultiPLY model \citep{MultiPLY} exemplifies the next step in multi-sensory embodied AI that actively explores and interacts with its surroundings, rather than passively receiving data, using a large-scale dataset that includes tactile and temperature information alongside visual data. The system constructs object-centric representations of the 3D world, leveraging pretrained LLMs (e.g., LLaVA \citep{LLaVA}) to encode and reason over both abstract visual features and real-time sensory feedback. By introducing structured action tokens (such as <TOUCH> or <HIT>) and state tokens (e.g., <TEMPERATURE>, <IMPACT SOUND>), the model enables bidirectional communication between language understanding and physical interaction, allowing the LLM to guide exploration and update its internal state based on multisensory observations.

Autonomous vehicles predominantly rely on LiDAR sensors to perceive their surroundings, primarily because LiDAR provides highly accurate and reliable depth estimates, which are crucial for ensuring safety in self-driving scenarios where even a single error can have serious consequences \citep{OmniDrive}. Compared to RGB-D cameras, LiDAR systems are less affected by changing lighting conditions and can deliver precise range data over longer distances, making them particularly well-suited for outdoor environments and high-speed navigation. While RGB-D sensors offer dense texture and are effective in controlled indoor settings, their depth accuracy and performance degrade significantly in bright sunlight or at greater distances, limiting their utility for outdoor autonomous driving. As a result, most research and practical deployments in autonomous driving leverage LiDAR as the primary input modality.

A notable example is OmniDrive \citep{OmniDrive}, which enters by integrating advanced vision technologies with LLMs to enable robust 3D perception, reasoning and planning for autonomous driving. The core approach employs a novel 3D multimodal LLM architecture inspired by Q-Former \citep{QFormer}, where multi-view, high-resolution images are processed by a shared visual encoder and then compressed into a set of sparse queries augmented with 3D positional encoding. These queries jointly encode dynamic objects and static map elements, effectively creating a condensed and spatially-aware world model. This representation is then aligned with an LLM, which uses its strong reasoning and language capabilities to perform complex tasks such as scene description, traffic regulation understanding, 3D grounding, counterfactual reasoning and motion planning. The system is trained in two stages: first, 2D pretraining leverages large-scale image-text pairs to align visual and language features, followed by 3D finetuning that enhances spatial localization and temporal modeling for video input Looking ahead, the developers of OmniDrive plan to validate its robustness and scalability on larger datasets such as nuPlan \citep{nuPlan}, which offers extensive real-world driving data and a comprehensive evaluation framework for autonomous vehicle planning.

{For safety-critical open-vocabulary understanding applications of embodied agents, it is necessary to achieve sub-100 ms reasoning latencies, but it remains a significant challenge. The primary bottleneck stems from the formidable computational demands of processing high-dimensional, multimodal inputs, such as 3D point clouds and executing complex reasoning within large language models (LLMs). Current systems often rely on multi-stage pipelines for perception and planning, which introduce significant latency overheads that are incompatible with hard real-time constraints (Section \ref{sec:challenges}). Although research into end-to-end architectures that directly embed 3D scene information into lightweight prompts Section \ref{sec:open_vocabulary_understanding_and_pretraining} efficient fusion techniques like Vision-and-Language Transformers (VILT) \cite{ViLT} or adapter modules show promise, but these approaches have yet to consistently meet the stringent sub-100 ms threshold. Future progress towards bridging this gap will likely depend on innovations in real-time adaptive architectures, policy steering via latent \mbox{alignment \citep{FOREWARN},} and aggressive hardware-software co-design tailored for accelerated, high-fidelity reasoning (Section \ref{sec:challenges}).}

Overall, advancing autonomous driving requires moving beyond traditional vision and depth sensing to embrace a broader spectrum of modalities, including thermal, tactile, audio and more. These innovations collectively enhance perception, safety and adaptability, ensuring autonomous systems can operate reliably in the complex, unpredictable conditions of real-world driving environments \citep{OpenRSS,RTFNet,MultiPLY,CLIP2Point,PointBindPointLLM,JM3D}.

Collectively, these models demonstrate the rapidly expanding capabilities of LLMs and related technologies in advancing 3D scene understanding across various domains. By integrating spatial data with language, recent frameworks have achieved notable improvements in accurately interpreting and reasoning about complex 3D environments. This synergy not only boosts the precision and adaptability of scene understanding but also broadens the scope of practical applications, ranging from robotics and autonomous vehicles to urban planning and smart environments. As the field progresses, future research will likely focus on creating more robust and versatile autonomous agents capable of seamlessly transitioning between tasks such as manipulation, navigation and driving, dynamically adjusting to the needs of their environment. Hardware platforms like \mbox{TurtleBot \citep{RRExBoT},} LoCoBot and Fetch exemplify the kinds of robots that can embody these advanced models, serving as practical testbeds for developing and evaluating next-generation 3D scene understanding and embodied intelligence.

\section{Datasets} \label{sec:datasets}

Datasets are building blocks of any machine learning or deep learning model. They are like the alphabet, using which the model learns to spell words and sentences. Having high-quality and big datasets is therefore crucial for training any neural network architecture. The better the quality of our training dataset the better the quality of our model will be. In this section, we explore the various datasets that are available out there, along with some of their statistics (number of scenes) and attributes (kind of scenes) for training robots for various tasks of scene understanding, grounding and visual question answering for \mbox{an embodied agent.} Table \ref{tab:3d_datasets} gives statistics for various available 3D datasets.

\begin{table}[htbp]
\small
\caption{Summary of key statistics and features of various benchmark 3D datasets, covering applications like grounding, scene reconstruction, semantic annotations and embodied AI training.\label{tab:3d_datasets}}
\begin{center}
\newcolumntype{L}{>{\raggedright\arraybackslash}X}
    \begin{tabularx}{\fulllength}{lL}
        \toprule
        \textbf{Datasets} & \textbf{Remarks}\\
        \midrule
        ScanNet \citep{ScanNet} & 2.5M views in 1513 scenes with 3D poses, reconstructions and semantic annotations.\\
        Sr3D \textsuperscript{1} \citep{ReferIt3D} & 83,572 synthetic utterances. Built on top of ScanNet \cite{ScanNet}.\\
        Sr3D+ \textsuperscript{1} \citep{ReferIt3D} & 83,572 synthetic utterances. Enhanced with fewer-distractor utterances.\\
        Nr3D \textsuperscript{1} \citep{ReferIt3D} & 41,503 human utterances. Built on top of ScanNet \cite{ScanNet}.\\
        ScanRefer \textsuperscript{1} \citep{ScanRefer} & 51,583 descriptions of 11,046 objects from 800 ScanNet scenes \cite{ScanNet}.\\
        3RScan \citep{3RScan} & 1482 RGB-D scans of 478 environments with 6DoF mappings and temporal changes.\\
        ScanScribe \textsuperscript{1} \citep{ScanScribe} & 2995 RGB-D scans of 1185 scenes with 278K paired descriptions.\\
        KITTI360Pose \textsuperscript{1} \citep{KITTI360Pose} & 150K+ images, 1B 3D points with coherent 2D-3D semantic annotations.\\
        ScanNet++ \citep{ScanNet++} & 460 scenes, 280,000 captured DSLR images and over 3.7M iPhone RGBD frames.\\
        ARKitScenes \citep{ARKitScenes} & 5047 captures of 1661 unique scenes with oriented Bounding Box of room-defining objects.\\
        HM3D \citep{HM3D} & 1000 3D reconstructions of multi-floor residences and private indoor spaces.\\
        MultiScan \citep{MultiScan} & 273 scans of 117 scenes with 10,957 objects, part-level semantics and mobility annotations.\\
        Structured3D \textsuperscript{2} \citep{Structured3D} & Comprises 3500 scenes, 21,835 rooms, 196K renderings, with ``primitive + relationship'' structures.\\
        ProcTHOR \textsuperscript{2} \citep{ProcTHOR} & 10,000 procedurally generated 3D scenes for training embodied AI agents.\\
        Matterport3D \citep{Matterport3D} & 10,800 panoramas, 194,400 RGB-D images of 90 scenes with 2D/3D semantics.\\
        ModelNet \textsuperscript{2} \citep{3DShapeNets} & 151,128 3D CAD models belonging to 660 unique object categories.\\
        Kaist \citep{Kaist} & 95K color-thermal image pairs with 103K annotations and 1182 unique pedestrians.\\
        ScanObjectNN \citep{ScanObjectNN} & 15,000 \textsuperscript{3} in 15 categories with point-based attributes and semantic labels.\\
        ScanQA \citep{ScanQA} & 40K question-answer pairs from 800 indoor scenes drawn from the ScanNet \cite{ScanNet} dataset.\\
        SQA3D \citep{SQA3D} & 650 scenes \cite{ScanNet} with 6.8K situations, 20.4K descriptions and 33.4K reasoning questions.\\
        SUNCG \textsuperscript{2} \citep{SUNCG} & 45,622 scenes, 49,884 floors, 404,058 rooms and 5.7M object instances across 84 categories.\\
        REVERIE \citep{REVERIE} & 10,318 panoramas of 86 buildings with 4140 objects and 21,702 crowd-sourced instructions.\\
        FAO \citep{SOON} & 1500 aligned IR-visible image pairs across 14 classes, covering harsh conditions.\\
        FMB \citep{FMB} & 4K sets of annotated instructions with 40K trajectories.\\
        nuPlan \citep{nuPlan} & 1282 h of driving scenarios with auto-labeled object tracks and traffic light data.\\
        NTU RGB+D 120 \citep{liu2019ntu} & 114K+ RGB+D human action samples, 120 action classes and 8M+ frames.\\
        STCrowd \citep{STCrowd} & Synchronized LiDAR and camera data with 219K pedestrians, 20 persons per frame.\\
        FLIR ADAS \citep{FlirADAS} & RGB + Thermal frames, for various weather conditions for ADAS.\\
        3D-Grand \citep{3DGrand} & 40,087 household scenes paired with 6.2 million densely grounded scene-language instruction.\\
        EmbodiedScan \citep{EmbodiedScan} & Over 5k scans, 1M egocentric RGB-D views, 1M language prompts, 160k 3D-oriented boxes for \mbox{760 categories.}\\
        M3DBench \citep{M3DBench} & Over 320K language pairs with 700 scenes with a special prompting that interweaves language with visual cues.\\
        SceneVerse \citep{SceneVerse} & 68K scenes with 2.5M vision-language pairs generated using human annotations and \mbox{scene-graph-based approach.}\\
        PhraseRefer \cite{PhraseRefer} & 227K phrase-level annotations from 88K sentences across the Nr3D, Sr3D and ScanRefer \cite{ReferIt3D} datasets.\\
        CLEVR3D \cite{CLEVR3D} & 171K questions about object attributes and spatial relationships, generated from \mbox{8771 3D scenes.}\\
        \bottomrule
    \end{tabularx}
    \end{center}
	\noindent{\footnotesize{\textsuperscript{1} Generally used for grounding; \textsuperscript{2} Synthetic Dataset; \textsuperscript{3} The number provided in the paper is an approximation.}}
\end{table}

Typical 2D Vision (Images) Datasets have number of data points in Millions (M) and Billions (B) \citep{OpenImages, ImageNet, CC12M, LAION400M, LAION5B, CLIP, RedCaps}, but on the other hand as can be seen from Table \ref{tab:3d_datasets} that 3D datasets are much smaller in size as compared to 2D datasets with number of datapoints being in Thousands (T/K). Due to this, the current state of the art (SOTA) for 2D VLMs is way ahead of current 3D VLMs in terms of accuracy and precision. This problem of scarcity of available 3D Data for training models is further discussed in Section \ref{sec:challenges}.

There might exist a distributional bias in the dataset, such as imbalances between \linebreak urban/rural or day/night scenes that degrade 3D-LLM generalizability, so to mitigate these, a combination of dataset-level and model-level strategies is essential. At the dataset level, this involves not only curating diverse real-world data but also enriching training distributions with rare scenarios through synthetic data generation via text-to-3D \mbox{models \citep{DreamFusion,GPT4Point}} and applying specialized point cloud augmentations. At the model level, domain adaptation techniques are used to align feature distributions and minimize discrepancies between different conditions \citep{RevisitingDomainAdaptive3DObjectDetection, MaximumClassifierDiscrepancy}, while the fusion of complementary sensor modalities like LiDAR, camera and thermal data enhances perceptual robustness against environmental shifts \citep{FogFusion, MultimodalEndToEndLearning}. Crucially, leveraging the broad contextual knowledge from large-scale pretraining for open-vocabulary and zero-shot learning enables generalization to unseen conditions, directly countering the effects of biased training data, as exemplified by foundational models like CLIP \citep{CLIP} and unified frameworks such as ULIP \citep{ULIP}.

\section{Evaluation Metrics} \label{sec:evaluation_metrics}

Evaluation Metrics play a crucial role in training any deep learning/machine learning model. It can be a loss function or reward function, which essentially acts as a feedback for the model on its performance, based on which the model weights are updated after each iteration while training. In this section, we discuss a few of those evaluation metrics used in the literature while training their models.

To evaluate the sentence similarity at the word level, we can use metrics like \citep{METEOR, ROUGE, CIDEr} as used in \citep{OmniDrive}. Ref. \citep{METEOR} is an automatic metric for Machine Translation (MT) evaluation, which has been demonstrated to have high levels of correlation with human judgments of translation quality, outperforming the then SOTA IBM's Bleu \citep{Bleu} metric, can be used as a target function in parameter optimization training procedures \citep{METEOR}. Metric for Evaluation of Translation with Explicit ORdering or METEOR \citep{METEOR} evaluates machine translation by comparing a candidate with reference translations and a sequence of word-mapping modules incrementally produces this alignment. It computes unigram precision $P$ and recall $R$ as $P = \frac{\text{matched words}}{\text{candidate words}}$ and $R = \frac{\text{matched words}}{\text{reference words}}$, combining them into a weighted harmonic mean given by Equation (\ref{eq:meteor}).
\begin{equation}
    F_{\text{mean}} = \frac{(1 + \alpha) \cdot P \cdot R}{\alpha \cdot P + R} \text{ with } \alpha = 0.9
    \label{eq:meteor}
\end{equation}

\noindent A fragmentation penalty $P_{\text{frag}} = \gamma \cdot \text{frag}^\beta$ and $$ \text{frag =} \frac{\text{chunks}}{\text{matched words}}$$

\textls[-15]{$0 < \gamma < 1$, accounts for maximum penalty and $\beta$ determines the functional relationship between fragmentation and penalty. The final score is $\text{METEOR} = F_{\text{mean}} \cdot (1 - P_{\text{frag}})$.} It incorporates linguistic features like stemming, synonyms and penalizes \mbox{disordered alignments.}

The CIDEr \citep{CIDEr} (Consensus-based Image Description Evaluation) metric evaluates how well a candidate sentence describes an image by comparing it to a set of reference sentences. It focuses on n-grams and incorporates a Term Frequency-Inverse Document Frequency (TF-IDF) weighting scheme. Each sentence (candidate or reference) is represented using a set of n-grams, where an n-gram is an ordered sequence of n words. For this metric, n-grams up to a length of 4 are used (i.e., unigrams, bigrams, trigrams and 4-g). The weight of an n-gram $ \omega_k $ is computed using TF-IDF, which accounts for its importance within the dataset. The TF-IDF vector is calculated using the Equation (\ref{eq:gk}).
\begin{equation}
    g_k\left(s_{i j}\right)= \frac{h_k\left(s_{i j}\right)}{\sum_{\omega_l \in \Omega} h_l\left(s_{i j}\right)} \log \left(\frac{I}{\sum_{I_p \in I} \min \left(1, \sum_q h_k\left(s_{p q}\right)\right)}\right)
    \label{eq:gk}
\end{equation}

\noindent where $ h_k(s_{ij}) $ is the frequency of the n-gram $ \omega_k $ in sentence $ s_{ij} $, $ I $ is the total number of images in the dataset, $ \Omega $ is the vocabulary of all n-grams and $ \min(1, \sum h_k(s_{pq})) $ indicates whether $ \omega_k $ is present in the reference set. The similarity for n-grams of length $ n $ is calculated using the average cosine similarity between the candidate and reference sentences. The cosine similarity is calculated using Equation (\ref{eq:cosine}).
\begin{equation}
    \text{CIDEr}_n(c_i, S_i) = \frac{1}{m} \sum_j \frac{g^n(c_i) \cdot g^n(s_{ij})}{\|g^n(c_i)\| \|g^n(s_{ij})\|}
    \label{eq:cosine}
\end{equation}

\noindent where $ g^n(c_i) $ is the TF-IDF vector for n-grams of the candidate sentence $ c_i $, $ g^n(s_{ij}) $ is the TF-IDF vector for n-grams of the reference sentence $ s_{ij} $ and $ m $ is the number of reference sentences. The final CIDEr score, which combines scores from different n-gram lengths, is given by Equation (\ref{eq:cider}).
\begin{equation}
    \text{CIDEr}(c_i, S_i) = \sum_{n=1}^N w_n \cdot \text{CIDEr}_n(c_i, S_i)
    \label{eq:cider}
\end{equation}

\noindent where $ w_n $ is the weight for each n-gram length, typically $ w_n = 1/N $ (i.e., uniform weighting) and $ N = 4 $ is the maximum n-gram length. The TF component captures how frequently an n-gram appears in the candidate/reference sentence, while the IDF discounts n-grams that are common across all images, emphasizing visually descriptive and rare words. Cosine similarity measures alignment between the candidate and reference sentences in terms of n-gram usage. This metric is particularly effective for evaluating image descriptions as it balances precision and recall while penalizing generic terms.

To evaluate the performance of the counterfactual reasoning, we can use any general-purpose LLM like \citep{Radford2018ImprovingLU, devlin2018bert, achiam2023gpt} to extract keywords based on the predictions. Then the extracted keywords can be compared with the ground truth to calculate the Precision and Recall for each category. This method has been implemented in OmniDrive \citep{OmniDrive}, where they use it to predict the category of accident (safety, collision, running a red light, out of drivable area, etc.).

The Top-k metric is a common evaluation measure used in machine learning tasks, such as those used in \citep{JM3D}, particularly in classification and ranking problems, to assess the performance of a model. It determines how often the correct label or item appears within the top $ k $ predictions made by the model. Normally, for a given test sample, the model produces a ranked list of predictions and the Top-k metric checks if the ground truth label is among the top $ k $ predictions. The Top-k metric is particularly useful in scenarios where multiple predictions are acceptable or when the model's top choice may not always be correct, but the correct answer is still ranked highly. By adjusting $ k $, the metric can provide insights into the model's ability to rank relevant items near the top of its predictions, making it a flexible and informative evaluation tool.

Cross-entropy loss is a widely used loss function for classification tasks like \citep{PLA, EQAInPhotorealisticEnvironments}, particularly in scenarios involving probabilistic models. It measures the dissimilarity between the true label distribution and the predicted probability distribution produced by the model. Given a dataset with $ N $ samples, the cross-entropy loss is computed by Equation (\ref{eq:ce}).
\begin{equation}
    \mathcal{L}_{\text{CE}} = -\frac{1}{N} \sum_{i=1}^N \sum_{c=1}^C y_{i,c} \log(\hat{y}_{i,c})
    \label{eq:ce}
\end{equation}

\noindent where $ y_{i,c} $ is a binary indicator (0 or 1) that specifies whether class $ c $ is the correct class for sample $ i $, $ \hat{y}_{i,c} $ is the predicted probability of class $ c $for sample $ i $ and $ C $ is the total number of classes. The loss penalizes incorrect predictions by assigning a higher penalty when the model assigns a low probability to the correct class. Intuitively, the cross-entropy loss quantifies how well the predicted probability distribution aligns with the true labels. It is particularly effective when combined with softmax activation in the final layer of a neural network, as the softmax ensures that the predicted probabilities form a valid distribution. The loss function drives the model to maximize the predicted probability for the correct class while minimizing probabilities for incorrect classes, thus promoting accurate classification. Gradient-based optimization methods like stochastic gradient descent make use of the differentiability of the cross-entropy loss function to arrive at the \mbox{optimum value.}

Ref. \citep{EQAInPhotorealisticEnvironments} uses smooth L1 loss, also referred to as Huber loss, which is a combination of L1 and L2 loss functions, designed to be less sensitive to outliers while retaining the benefits of both. It applies a piecewise definition: small errors are treated with a squared error (L2), while large errors are treated with an absolute error (L1). The smooth L1 loss for a single prediction is defined as given in Equation (\ref{eq:smoothl1}).
\begin{equation}
    \mathcal{L}_{\text{SmoothL1}}(x) = \begin{cases} 
    0.5x^2 & \text{if } |x| < \delta, \\
    \delta(|x| - 0.5\delta) & \text{otherwise}
    \end{cases}
    \label{eq:smoothl1}
\end{equation}

\noindent where $ x = y_i - \hat{y}_i $ is the prediction error and $ \delta $ is a threshold that determines the transition point between L1 and L2 behavior. For small errors ($ |x| < \delta $), the loss behaves like L2 loss, providing smooth gradients that are beneficial for optimization. For large errors ($ |x| \geq \delta $), the loss transitions to L1 loss, reducing the influence of outliers. The smooth L1 loss is widely used in computer vision tasks, such as object detection, because it provides a balance between robustness to outliers and smooth optimization.

The Earth Mover's Distance (EMD) \citep{EMD} is a metric used to quantify the similarity between two probability distributions or sets by solving a transportation problem as used in \citep{EQAInPhotorealisticEnvironments, LearningRepresentationsAndGenerativeModels}. It measures the minimum cost of transforming one distribution into the other. For two equally sized subsets $ S_1 \subseteq \mathbb{R}^3 $ and $ S_2 \subseteq \mathbb{R}^3 $, the EMD is defined by Equation (\ref{eq:emd}).
\begin{equation}
    d_{\text{EMD}}(S_1, S_2) = \min_{\phi: S_1 \to S_2} \sum_{x \in S_1} \|x - \phi(x)\|_2
    \label{eq:emd}
\end{equation}

\noindent where $ \phi $ is a bijection mapping each element in $ S_1 $ to an element in $ S_2 $. Intuitively, EMD represents the minimum ``work'' required to match all elements of $ S_1 $ with those in $ S_2 $, where work is defined as the product of the distance between matched elements and the amount of ``mass'' being moved. EMD is differentiable almost everywhere, making it suitable for gradient-based optimization. This property has made EMD popular as a loss function in applications such as image retrieval, 3D shape comparison and generative modeling, where preserving structural similarity between distributions or sets is critical. Its robustness and interpretability make it a valuable tool in tasks requiring alignment between complex data representations.

Embodied agent not only involves the task of grounding in the 3D scene, but subsequent tasks involve navigating to the object and performing the task, for which some special metrics are used.

Ref. \citep{LearningFromUnlabeled3DEnvironments} uses the SPL (Success weighted by Path Length) metric, which evaluates how efficiently an agent completes a task in navigation scenarios by comparing the agent's path length $ P $ to the shortest path length $ L $. It is defined in Equation (\ref{eq:spl}).
\begin{equation}
    \text{SPL} = S \cdot \frac{L}{\max(P, L)}
    \label{eq:spl}
\end{equation}

\noindent where $ S $ is 1 if the agent successfully completes the task and 0 otherwise, $ P $ is the agent's path length and $ L $ is the shortest path length from the start to the goal. While SPL rewards agents for completing tasks with shorter paths, it does not account for energy efficiency or dynamics, potentially favoring point-turn behaviors that may be suboptimal. Additionally, SPL does not penalize redundant or inefficient actions like pausing or pivoting if they do not increase path length. 

To address these limitations, the Success weighted by Completion Time (SCT) metric is proposed \citep{SCT}, which incorporates time efficiency. SCT is defined as given by Equation (\ref{eq:sct}).
\begin{equation}
    \text{SCT} = S \cdot \frac{T}{\max(C, T)}
    \label{eq:sct}
\end{equation}

\noindent where $ C $ is the agent's completion time and $ T $ is the shortest possible time to reach the goal based on the agent's dynamics. Unlike SPL, SCT accounts for energy-efficient and time-optimal behaviors, providing a more holistic evaluation of agent performance. It incentivizes agents to minimize completion time while adhering to task requirements, making it suitable for scenarios where dynamics and efficiency are critical.

In evaluating language-enabled robotic systems, a critical distinction must be made between task-driven outcomes and generic natural language processing (NLP) scores. While NLP metrics such as BLEU and CIDEr are useful for assessing the linguistic quality of generated text, they often fail to correlate with the functional success of a robotic agent. These scores primarily measure n-gram overlap and language similarity, which can be misleading as a robot might generate a linguistically fluent response that leads to incorrect or unsafe physical actions \citep{SurveyOnChatbots,SurveyEvaluationMethodsForDialogueSystems}. Therefore, for robotics applications where physical interaction and safety are paramount, task-driven metrics must be prioritized. Metrics like success rate, completion time and energy consumption provide direct, objective measures of a robot's operational effectiveness, efficiency and safety. A comprehensive evaluation framework should thus integrate both types of metrics, with task-driven outcomes serving as the primary indicators of performance and NLP scores acting as complementary diagnostics for the language interface, especially in safety-critical domains where functional reliability is non-negotiable \citep{MetricsForRobotProficiency,VirtualAgentsToRobotTeams}. Future research should focus on developing novel, integrated metrics that holistically capture both semantic understanding and successful task execution.

When evaluating the transfer of knowledge from 2D Vision-Language Models (VLMs) to 3D Vision-Language Models (LVMs), the choice of knowledge distillation (KD) architecture is a critical factor influencing performance on downstream tasks. Empirical evidence suggests that hybrid KD architectures, which synergistically combine teacher-student frameworks with contrastive loss objectives, are most effective. This dominant approach utilizes a pre-trained 2D-VLM as a teacher to guide a 3D-LVM student, aligning the student's 3D spatial features with the teacher's rich semantic priors, a technique successfully employed in models such as ULIP \citep{ULIP} and for spatial-semantic alignment in methods like \mbox{CSA \& M3LM \citep{ContextAwareAlignmentAndMutualMaskingFor3DLanguagePreTraining}.} The integration of a contrastive loss, foundational to modern multimodal alignment \citep{CLIP} and also central to ULIP \citep{ULIP}, further enhances transfer by explicitly forcing the alignment of 2D and 3D embeddings in a shared latent space to improve modality-agnostic representation and open-vocabulary generalization. As a computationally economical alternative, adapter-based methods inject lightweight, trainable modules into frozen 2D-VLM backbones to adapt them for 3D inputs, minimizing parameter overhead while effectively bridging the domain gap \citep{PointBindPointLLM}. Collectively, while these methods are powerful, significant challenges remain in mitigating the modality gap between 2D pixels and 3D spatial data and overcoming data scarcity. Future directions point towards multi-stage distillation and synthetic data augmentation to further enhance the robustness and zero-shot capabilities of 3D-LVMs.

\section{Challenges and Limitations}\label{sec:challenges}

Working with 3D data and LLMs in the context of embodied agent understanding poses significant challenges, particularly in terms of computational resources, data availability, interpretability and real-time performance. Processing 3D data, often derived from sensors such as LiDAR or depth cameras, requires handling high-dimensional information that is computationally intensive. Tasks such as point cloud generation, depth estimation and segmentation demand substantial computational power, especially when real-time processing is required. Simultaneously, LLMs, with their high parameterization and extensive resource requirements for training and inference, further exacerbate the computational load. The intersection of these two domains creates a bottleneck for practical applications, particularly in environments such as robotics and augmented reality, where efficiency and scalability are crucial.

This scale limit of 3D data, caused by the high cost of 3D data collection and \linebreak \mbox{annotation \citep{ShapeNet, 3DShapeNets, PointBERT, RevisitingPointCloudShapeClassification},} has been hindering the generalization of 3D recognition models and their real-world applications. Another critical limitation is the scarcity of high-quality, annotated 3D datasets. Unlike 2D image datasets, which are widely available, 3D data remains underrepresented due to the challenges associated with its collection and annotation. Creating labeled 3D datasets involves resource-intensive processes, including capturing point clouds or depth images and annotating spatial relationships or object categories. Annotation in 3D requires specialized expertise and tools, making it both time-consuming and expensive. Some researchers have attempted to mitigate this issue through automatic annotation pipelines or synthetic data generation, leveraging simulation environments to produce large-scale datasets. For instance, models like PointLLM have been utilized to generate annotated 3D data automatically, thus reducing the dependency on manual efforts and accelerating dataset creation \citep{PointLLM}.

The interpretability of both 3D models and LLMs poses another significant challenge, particularly in applications that require high reliability, such as autonomous driving or healthcare. Deep learning models for 3D tasks, including object recognition and scene understanding, often lack inherent transparency, making it difficult to understand how spatial features are interpreted. Similarly, LLMs operate as black boxes, making the reasoning behind their decisions opaque. This lack of explainability is a major limitation in ensuring safety and trust in embodied agents. While techniques such as attention mechanisms and post-hoc analysis tools have been proposed to enhance interpretability, these approaches remain nascent and require further development \citep{ExplainabilityDeepModels}.

Integrating 3D data with LLMs introduces multimodal challenges, particularly in aligning and synchronizing diverse input modalities. For example, an agent may need to process textual commands provided by an LLM while simultaneously interpreting its 3D environment to execute actions. Natural language instructions are often ambiguous or context-dependent, making alignment with 3D spatial data complex. Effective multimodal integration is crucial for ensuring that embodied agents can reason and act coherently, yet achieving this remains a significant technical hurdle. Misalignment between textual and spatial representations can result in poor performance, undermining the agent’s ability to execute tasks effectively \citep{MultimodalAlignmentAndFusionSurvey}.

A significant challenge arises when fusing data from multiple sensors, especially when integrating this information with Large Language Models (LLMs). Each sensor, such as a camera, microphone or thermal imager, has its own unique type of noise and statistical profile, making it challenging to combine their data accurately. To address this, two primary uncertainty quantification frameworks are particularly well-suited: Bayesian fusion and factor graphs. Bayesian fusion is a foundational approach that treats each sensor's measurement as a probability. It combines prior knowledge with new sensor data using Bayes' rule to produce a final, unified estimate that also includes a clear measure of its own confidence or uncertainty. This method is highly flexible, can be updated dynamically as new data arrives and is capable of handling sensor failures or missing signals, which is essential for real-world robotics applications \citep{ManagingUncertainity}. For more complex, high-dimensional and non-linear fusion tasks, factor graphs provide a more advanced graphical modeling solution. In this framework, system variables and sensor measurements are represented as nodes in a graph and the probabilistic relationships between them are encoded as "factors." This structure enables the joint optimization of all sensor inputs simultaneously and supports both batch and incremental updates, making it highly efficient. Factor graphs have proven to be more stable and consistent than traditional filters, especially in challenging navigation and perception tasks where they can recover from errors caused by sensor dropouts or misalignments \citep{SensorFusionApplication, FactorGraphForINS}. The latest frontier involves integrating LLMs into these frameworks by feeding them structured representations of the fused, uncertain data, such as probability maps. The LLM then adds a layer of semantic reasoning to interpret this information, correct errors and make more robust decisions in dynamic, open-world environments. Researchers are now exploring how LLMs can enhance fusion for specific goals like target detection through transfer learning \citep{SensorFusionForTargetDetection} and are developing sophisticated ensemble and causal methods to better decompose and mitigate sources of uncertainty within the LLM's reasoning process itself \citep{UncertaintyAwareFusion, UncertaintyQuantification}.

Point cloud data augmentation is a critical strategy to address overfitting and poor generalization, challenges commonly observed in 3D deep learning models due to limited diversity in training datasets and inherent variations in real-world 3D data, such as occlusions, noise and changes in object pose or scale. As shown in Figure \ref{fig:augmentation_methods}, augmentation methods are broadly divided into two main categories: basic and specialized point cloud augmentation \citep{AdvancementsInPointCloudDataAugmentation}. Basic augmentation techniques, such as affine transformations (translation, rotation, flipping and scaling), introduce geometric variations to improve robustness to changes in orientation, position and size. Methods like point dropping and jittering simulate occlusions and sensor noise, while ground truth sampling retains structural integrity while generating diverse samples. Specialized augmentation techniques address domain-specific challenges, including mixup operations that blend multiple point clouds for diversity, domain adaptation to align training and testing data distributions and adversarial deformations to test robustness against distortions. Additional approaches, such as up-sampling and completion, address sparsity and occlusions by densifying or reconstructing point clouds, while generation and multimodal augmentation enrich datasets through synthetic samples and cross-modal integration. These techniques collectively enhance dataset diversity, improve model robustness and mitigate overfitting, enabling better performance in real-world 3D vision tasks.

A primary vulnerability lies in sensor failure and data robustness. LLM-driven systems are semantically sensitive, meaning corrupted or ambiguous sensor data can lead to catastrophic failures. For instance, vision-only systems suffer significant performance degradation in adverse weather conditions like rain and fog \citep{AtmosphericInterferenceLidar}. While multimodal sensor fusion, incorporating thermal imaging to penetrate obscurants \citep{OpenRSS} or tactile sensing for physical property recognition \citep{AtmosphericInterferenceLidar}, is a promising solution, it introduces new challenges. The abstract nature of non-visual data is often difficult for LLMs to interpret without specialized pre-training and sophisticated fusion and alignment \mbox{architectures \citep{MultimodalFusionAndVLMs,MultimodalAlignmentAndFusionSurvey}.} Furthermore, all sensors are susceptible to inherent errors, such as multi-path interference in Time-of-Flight (ToF) cameras, which requires dedicated deep learning models for \mbox{correction \citep{DLForMultiPathErrorRemovalInToFSensors}.}. Current robotic systems often lack real-time monitoring and dynamic recovery mechanisms, highlighting the need for unified frameworks that can detect, identify and correct execution-time failures as they occur.

\begin{figure}
    \centering
    \includegraphics[width=0.99\columnwidth]{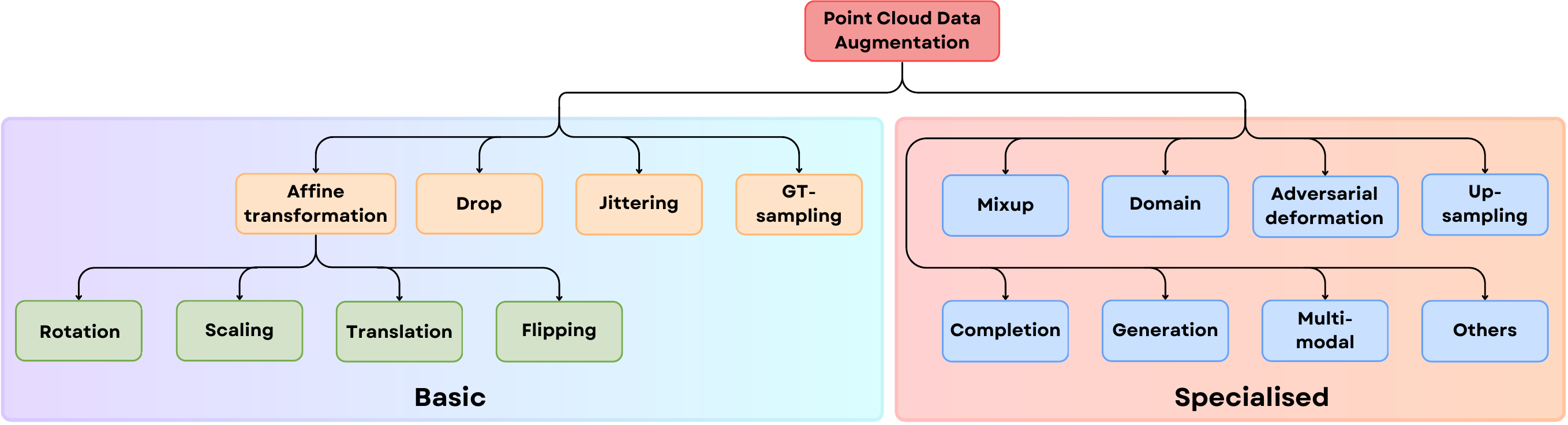}
    \caption{Augmentation Methods: Classification of point cloud data augmentation techniques into basic and specialized methods.}
    \label{fig:augmentation_methods}
\end{figure}

Another significant concern is adversarial robustness and safety. Embodied agents are highly susceptible to adversarial attacks and hazardous instructions, which can exploit model vulnerabilities to cause harm. Research using safety benchmarks has revealed that current agents frequently violate core safety principles when subjected to ``jailbreak'' prompts designed to elicit unsafe behaviors. This is particularly true at the decision-making level, where LLM-based systems exhibit low robustness. For high-stakes applications such as autonomous driving, ensuring the system can reliably interpret its environment and make safe decisions is paramount \citep{Atlas}. Future work must focus on developing comprehensive safety evaluation frameworks, including risk-aware instruction datasets and simulation sandboxes, to systematically test and harden the agent’s entire decision-making pipeline against adversarial manipulation.

The ethical deployment and interpretability of these systems remain major open problems. The risk of LLMs enacting discriminatory or unlawful actions due to biases in their vast training data is a significant ethical concern. This necessitates the development of comprehensive guidelines and evaluation frameworks that address fairness, transparency, accountability and robust safety guardrails. A major barrier to achieving this is the lack of interpretability in both LLMs and 3D deep learning models, which often function as ``black boxes'' \citep{yin2023survey}. This opacity makes it difficult to audit decisions, verify safety and build trust. Consequently, future research must prioritize the development of explainable AI (XAI) techniques, formal verification methods and policy steering mechanisms to ensure that LLM-driven robots are not only capable but also transparent, safe and ethically aligned.

Ensuring real-time performance and scalability is a persistent challenge. Embodied agents must process continuously changing 3D data while generating timely responses through LLMs, a combination that places immense computational demands on the system. For applications such as robotics, where delays can compromise functionality or safety, achieving the necessary computational efficiency is critical. Furthermore, scaling these systems to handle larger environments or more complex scenarios exacerbates the difficulty, as both 3D data processing and language models are inherently resource-intensive. Addressing these limitations is essential for advancing the practical deployment of embodied agents in real-world scenarios.

Large Language Models (LLMs) face significant challenges when working with non-textual modalities such as thermal, tactile and auditory data. For thermal data, the abstract nature of thermal signatures (such as temperature gradients and emissivity variations) makes it challenging for LLMs, primarily trained on text and visual data, to interpret and correlate these patterns with rich semantic descriptions without extensive, specialized pre-training. The typically low spatial resolution of thermal imagery further hampers the ability of LLMs to ground fine-grained textual descriptions or infer detailed object geometry. Moreover, the scarcity of large-scale, annotated thermal-language datasets limits the development of robust, thermal-aware LLMs, restricting their zero-shot or few-shot learning \mbox{capabilities \citep{LimitationsLLMs1, LimitationsLLMs2}.}

Grounding abstract human commands, such as ``tidy the room,'' into reliable physical 3D affordances presents a significant challenge, requiring the integration of advanced perception, reasoning and interaction capabilities. The process begins with scene understanding, where robots utilize 3D sensors, such as LiDAR and RGB-D cameras, to construct detailed spatial models and localize objects within them. To bridge the semantic gap, language-3D alignment models are crucial; systems like $\text{D}^3\text{Net}$ \citep{D3Net} and Transcrib3D \citep{Transcrib3D} leverage Large Language Models (LLMs) to convert natural language commands into spatial queries, effectively identifying task-relevant objects within the 3D representation. This is often coupled with task decomposition, where the LLM reasons over the scene to break down the abstract goal into a sequence of concrete, goal-directed actions. A key step is affordance extraction, where physical interaction possibilities, like graspability, movability or switchability, are inferred from an object's 3D geometry and semantic category. For instance, frameworks like PolarNet \citep{PolarNet} demonstrate how point cloud data can be encoded with language to guide robotic manipulation based on such inferred affordances. To handle ambiguity and dynamic environments, this pipeline is enhanced by multimodal sensor fusion and continuous adaptation. Despite significant progress, challenges persist in resolving linguistic ambiguity, aligning complex 3D data and the general scarcity of large paired language-3D datasets. Future work aims to improve zero-shot generalization, enhance LLM spatial reasoning as explored in ConceptGraphs \citep{ConceptGraphs} and enable richer affordance detection through sophisticated multimodal integration.

Despite its promise, mandating formal verification for entire large-scale language models is currently infeasible due to fundamental challenges in scale, complexity and specification. Modern LLMs, with their billions of parameters, present a state space that is intractably vast for exhaustive analysis by current formal methods, a problem often termed the ``curse of dimensionality''. A more profound obstacle is the semantic or abstraction gap that arises when attempting to formalize high-level, context-dependent safety properties, such as "harmlessness" or "honesty," into precise, machine-verifiable logical formulas. This difficulty is rooted in the inherent limitations of LLMs in deeply understanding human language and cognition. \citep{LimitationsLLMs1}. This challenge is significantly amplified in multimodal systems, such as those used in autonomous driving \citep{OmniDrive}, where verifying the correct interpretation and fusion of diverse data streams like 3D point clouds \citep{AdvancementsInPointCloudDataAugmentation} and visual data is an unsolved problem \citep{MultimodalAlignmentAndFusionSurvey,EMLMs}. Consequently, guarantees proven for a simplified, abstract model may not hold for the complex, often opaque behavior of the actual deployed \mbox{system \citep{ExplainabilityDeepModels},} particularly given that LLMs face significant challenges in non-textual modality \mbox{understanding \citep{LimitationsLLMs2},} making a universal mandate for formal verification impractical \mbox{at present.}

A significant challenge in the field is precisely defining the scope of adaptive architectures, as they do not exclusively imply any single methodology but rather represent a synergistic integration of multiple advanced techniques. For instance, modular transformers provide structural flexibility, allowing for dynamic expansion and task-specific customization without complete model retraining, as demonstrated by the Task Attentive Multimodal Continual Learning (TAM-CL) framework, which adds task-attention layers to a base transformer for efficient scaling \citep{DynamicTransformer}. This modularity is intrinsically linked to continual learning (CL), where techniques like knowledge distillation and experience replay are employed within the same framework to mitigate catastrophic forgetting and enable positive knowledge transfer across sequential tasks \citep{DynamicTransformer}. Furthermore, reinforcement-based controllers are integrated to enhance computational efficiency through mechanisms like adaptive attention spans, where each attention head learns a variable-length context window via L1 regularization, effectively pruning unnecessary computations in memory-intensive reinforcement learning tasks \citep{AdaptiveTransformersInRL}. Therefore, the primary challenge lies not in choosing between these paradigms but in developing architectures that effectively combine their strengths to achieve robust, scalable and efficient lifelong learning in complex, real-world environments.

Similarly, tactile data presents high dimensionality and lacks intuitive linguistic equivalents, complicating its representation and interpretation for LLMs. Concepts such as "slippery" or "fragile" are abstract and require explicit mapping from raw sensor data, which is often sparse and context-dependent, making scene-level understanding challenging. For auditory data, LLMs struggle with the ambiguity and noise inherent in real-world auditory scenes, particularly when distinguishing and interpreting non-speech environmental sounds. The temporal complexity of sound, where meaningful events unfold over time, adds another layer of difficulty, as synchronizing these patterns with static or asynchronous textual descriptions remains a significant hurdle \citep{LimitationsLLMs1, LimitationsLLMs2}. Research in multimodal AI and dimensionality reduction highlights these challenges, noting the interpretability gap and the need for better frameworks to connect high-dimensional, non-linguistic sensory data with language models.

\section{Conclusions}\label{sec:conclusion}

This review has provided a comprehensive examination of the synergistic integration of Large Language Models (LLMs) with 3D vision, charting the progress toward creating more intelligent and autonomous robotic systems. Our analysis was structured to address pivotal research questions concerning the architectural, multimodal and generalization challenges in this rapidly evolving field.

In response to our first research question on architectural paradigms for spatial grounding, we have detailed the dominant strategies for aligning the symbolic reasoning of LLMs with raw geometric data from 3D sensors. We explored a spectrum of methodologies, from supervised techniques that rely on dense annotations to more flexible, programmatic frameworks that leverage LLMs to decompose linguistic commands into executable spatial queries. These approaches are fundamental to achieving robust object referencing and localization, enabling robots to connect abstract language to physical entities in \mbox{their environment.}

Addressing our second question regarding the integration of non-visual sensor feedback, this paper highlighted how LLMs serve as a central reasoning engine for fusing heterogeneous data streams. By interpreting tactile force distributions for nuanced manipulation, thermal signatures for human detection in perceptually degraded conditions like fog or darkness and acoustic cues for event localization, LLMs construct a more holistic and robust 3D world model. This capability moves robotic perception beyond vision-centric limitations, significantly enhancing situational awareness, safety and the capacity for complex physical interaction.

Finally, to answer our third question concerning the challenges of 3D data scarcity and the modality gap, we identified and analyzed key emerging solutions. The paper surveyed open-vocabulary pre-training methods, which enable generalization to unseen objects without task-specific fine-tuning. Furthermore, we examined procedural text-to-3D generation systems, which mitigate data bottlenecks by programmatically synthesizing rich, annotated 3D environments from language prompts. The fusion of diverse sensory modalities, as explored throughout the paper, also serves as a critical strategy to build more robust and generalizable perception systems that are less dependent on any single \mbox{data source.}

Our key contributions include a thorough analysis of 3D-language alignment strategies, a resource catalog of benchmark datasets and evaluation metrics to standardize progress and a forward-looking discussion on multimodal integration. Future research must prioritize the development of real-time adaptive architectures, advance cross-modal distillation techniques to bridge the semantic gap between 2D and 3D modalities and refine evaluation protocols to better capture task-driven success over purely linguistic metrics. By continuing to merge the cognitive power of language with the spatial acuity of 3D perception, LLM-driven robotic systems are poised to achieve unprecedented levels of autonomy and awareness, unlocking new frontiers in manufacturing, healthcare and embodied AI.

\vspace{6pt}


\begin{thebibliography}{999}

\bibitem[Achiam et~al.(2023)Achiam, Adler, Agarwal, Ahmad, Akkaya, Aleman, Almeida, Altenschmidt, Altman, Anadkat, et~al.]{achiam2023gpt}
Achiam, J.; Adler, S.; Agarwal, S.; Ahmad, L.; Akkaya, I.; Aleman, F.L.; Almeida, D.; Altenschmidt, J.; Altman, S.; Anadkat, S.;  et~al.
\newblock Gpt-4 technical report.
\newblock {\em arXiv} {\bf 2023}, arXiv:2303.08774.

\bibitem[Touvron et~al.(2023)Touvron, Lavril, Izacard, Martinet, Lachaux, Lacroix, Rozi{\`e}re, Goyal, Hambro, Azhar, et~al.]{touvron2023llama}
Touvron, H.; Lavril, T.; Izacard, G.; Martinet, X.; Lachaux, M.A.; Lacroix, T.; Rozi{\`e}re, B.; Goyal, N.; Hambro, E.; Azhar, F.;  et~al.
\newblock Llama: Open and efficient foundation language models.
\newblock {\em arXiv} {\bf 2023}, arXiv:2302.13971.

\bibitem[Chiang et~al.(2023)Chiang, Li, Lin, Sheng, Wu, Zhang, Zheng, Zhuang, Zhuang, Gonzalez, Stoica and Xing]{vicuna2023}
Chiang, W.L.; Li, Z.; Lin, Z.; Sheng, Y.; Wu, Z.; Zhang, H.; Zheng, L.; Zhuang, S.; Zhuang, Y.; Gonzalez, J.E.;  et~al.
\newblock Vicuna: \mbox{An Open-Source} Chatbot Impressing GPT-4 with 90\%* ChatGPT Quality. 2023. Available online: \url{https://vicuna.lmsys.org} (accessed on 10 July 2025).

\bibitem[Zhu et~al.(2023)Zhu, Chen, Shen, Li and Elhoseiny]{zhu2023minigpt}
Zhu, D.; Chen, J.; Shen, X.; Li, X.; Elhoseiny, M.
\newblock Minigpt-4: Enhancing vision-language understanding with advanced large language models.
\newblock {\em arXiv} {\bf 2023}, arXiv:2304.10592.

\bibitem[Guo et~al.(2025)Guo, Yang, Zhang, Song, Zhang, Xu, Zhu, Ma, Wang, Bi, et~al.]{DeepSeekR1}
Guo, D.; Yang, D.; Zhang, H.; Song, J.; Zhang, R.; Xu, R.; Zhu, Q.; Ma, S.; Wang, P.; Bi, X.;  et~al.
\newblock Deepseek-r1: Incentivizing reasoning capability in llms via reinforcement learning.
\newblock {\em arXiv} {\bf 2025}, arXiv:2501.12948.

\bibitem[Devlin(2018)]{devlin2018bert}
Devlin, J.
\newblock Bert: Pre-training of deep bidirectional transformers for language understanding.
\newblock {\em arXiv} {\bf 2018}, arXiv:1810.04805.

\bibitem[Lan(2019)]{lan2019albert}
Lan, Z.
\newblock Albert: A lite bert for self-supervised learning of language representations.
\newblock {\em arXiv} {\bf 2019}, arXiv:1909.11942.

\bibitem[Sanh(2019)]{sanh2019distilbert}
Sanh, V.
\newblock DistilBERT, a distilled version of BERT: Smaller, faster, cheaper and lighter.
\newblock {\em arXiv} {\bf 2019}, arXiv:1910.01108.

\bibitem[Liu(2019)]{liu2019roberta}
Liu, Y.
\newblock Roberta: A robustly optimized bert pretraining approach.
\newblock {\em arXiv} {\bf 2019} arXiv:1907.11692.

\bibitem[Joshi et~al.(2020)Joshi, Chen, Liu, Weld, Zettlemoyer and Levy]{joshi2020spanbert}
Joshi, M.; Chen, D.; Liu, Y.; Weld, D.S.; Zettlemoyer, L.; Levy, O.
\newblock Spanbert: Improving pre-training by representing and predicting spans.
\newblock {\em Trans. Assoc. Comput. Linguist.} {\bf 2020}, {\em 8}, 64--77.

\bibitem[Radford and Narasimhan(2018)]{Radford2018ImprovingLU}
Radford, A.; Narasimhan, K.
\newblock Improving Language Understanding by Generative Pre-Training.
\newblock  2018. Available online: \url{https://cdn.openai.com/research-covers/language-unsupervised/language_understanding_paper.pdf} (accessed on 15 April 2025).

\bibitem[Radford et~al.(2019)Radford, Wu, Child, Luan, Amodei and Sutskever]{Radford2019LanguageMA}
Radford, A.; Wu, J.; Child, R.; Luan, D.; Amodei, D.; Sutskever, I.
\newblock Language Models are Unsupervised Multitask Learners.
\newblock \textit{OpenAI Blog} \textbf{2019}, \textit{1}, 9.

\bibitem[Brown et~al.(2020)Brown, Mann, Ryder, Subbiah, Kaplan, Dhariwal, Neelakantan, Shyam, Sastry, Askell, et~al.]{brown2020language}
Brown, T.; Mann, B.; Ryder, N.; Subbiah, M.; Kaplan, J.D.; Dhariwal, P.; Neelakantan, A.; Shyam, P.; Sastry, G.; Askell, A.;  et~al.
\newblock Language models are few-shot learners.
\newblock In \textit{Advances in Neural Information Processing Systems}; 	
Curran Associates Inc.: Red Hook, NY, USA, 2020; Volume 33, pp. 1877--1901.

\bibitem[Meta()]{Llama3.1}
Meta.
\newblock Llama 3.1.
\newblock  Available online: \url{https://ai.meta.com/blog/meta-llama-3-1/} (accessed on 5 February 2025).

\bibitem[Ouyang et~al.(2022)Ouyang, Wu, Jiang, Almeida, Wainwright, Mishkin, Zhang, Agarwal, Slama, Ray, et~al.]{InstructGPT}
Ouyang, L.; Wu, J.; Jiang, X.; Almeida, D.; Wainwright, C.; Mishkin, P.; Zhang, C.; Agarwal, S.; Slama, K.; Ray, A.;  et~al.
\newblock Training language models to follow instructions with human feedback.
\newblock In \textit{Advances in Neural Information Processing Systems}; Curran Associates Inc.: Red Hook, NY, USA, 2022; Volume 35, pp. 27730--27744.

\bibitem[Lewis(2019)]{BART}
Lewis, M.
\newblock Bart: Denoising sequence-to-sequence pre-training for natural language generation, translation and comprehension.
\newblock {\em arXiv} {\bf 2019}, arXiv:1910.13461.

\bibitem[Wikipedia()]{GPT4Wikipedia}
Wikipedia.
\newblock GPT-4.
\newblock  Available online: \url{https://en.wikipedia.org/wiki/GPT-4} (accessed on 19 December 2024).

\bibitem[Liu et~al.(2017)Liu, Wang, Zhuang and Hu]{NovelTrailDetection}
Liu, Y.; Wang, Q.; Zhuang, Y.; Hu, H.
\newblock A Novel Trail Detection and Scene Understanding Framework for a Quadrotor UAV with Monocular Vision.
\newblock {\em IEEE Sens. J.} {\bf 2017}, {\em 17}, 6778--6787.
\newblock {\url{https://doi.org/10.1109/JSEN.2017.2746184}}.

\bibitem[Gu et~al.(2019)Gu, Wang and Kamijo]{DDRInSmartphone}
Gu, Y.; Wang, Q.; Kamijo, S.
\newblock Intelligent Driving Data Recorder in Smartphone Using Deep Neural Network-Based Speedometer and Scene Understanding.
\newblock {\em IEEE Sens. J.} {\bf 2019}, {\em 19}, 287--296.
\newblock {\url{https://doi.org/10.1109/JSEN.2018.2874665}}.

\bibitem[Huang et~al.(2021)Huang, Lv, Xing and
  Wu]{AutonomousDrivingWithSceneUnderstanding}
Huang, Z.; Lv, C.; Xing, Y.; Wu, J.
\newblock Multi-Modal Sensor Fusion-Based Deep Neural Network for End-to-End Autonomous Driving with Scene Understanding.
\newblock {\em IEEE Sens. J.} {\bf 2021}, {\em 21}, 11781--11790.
\newblock {\url{https://doi.org/10.1109/JSEN.2020.3003121}}.

\bibitem[Ni et~al.(2024)Ni, Ren, Tang, Cao and Shi]{FastPanopticSegmentation}
Ni, J.; Ren, S.; Tang, G.; Cao, W.; Shi, P.
\newblock An Improved Shared Encoder-Based Model for Fast Panoptic Segmentation.
\newblock {\em IEEE Sens. J.} {\bf 2024}, {\em 24}, 22070--22083.
\newblock {\url{https://doi.org/10.1109/JSEN.2023.3332354}}.

\bibitem[Wu et~al.(2025)Wu, Tian, Swamy and Bajcsy]{FOREWARN}
Wu, Y.; Tian, R.; Swamy, G.; Bajcsy, A.
\newblock From Foresight to Forethought: VLM-In-the-Loop Policy Steering via Latent Alignment.
\newblock {\em arXiv} {\bf 2025}, arXiv:2502.01828.

\bibitem[Radford et~al.(2021)Radford, Kim, Hallacy, Ramesh, Goh, Agarwal, Sastry, Askell, Mishkin, Clark, et~al.]{CLIP}
Radford, A.; Kim, J.W.; Hallacy, C.; Ramesh, A.; Goh, G.; Agarwal, S.; Sastry, G.; Askell, A.; Mishkin, P.; Clark, J.;  et~al.
\newblock Learning transferable visual models from natural language supervision.
\newblock In Proceedings of the International Conference on Machine Learning, Virtual, 18--24 July 2021; pp. 8748--8763.

\bibitem[Lu et~al.(2024)Lu, Liu, Zhang, Wang, Dong, Liu, Sun, Ren, Li, Yang, et~al.]{DeepSeekVLM}
Lu, H.; Liu, W.; Zhang, B.; Wang, B.; Dong, K.; Liu, B.; Sun, J.; Ren, T.; Li, Z.; Yang, H.;  et~al.
\newblock Deepseek-vl: Towards real-world vision-language understanding.
\newblock {\em arXiv} {\bf 2024}, arXiv:2403.05525.

\bibitem[Jia et~al.(2021)Jia, Yang, Xia, Chen, Parekh, Pham, Le, Sung, Li and Duerig]{ALIGN}
Jia, C.; Yang, Y.; Xia, Y.; Chen, Y.T.; Parekh, Z.; Pham, H.; Le, Q.; Sung, Y.H.; Li, Z.; Duerig, T.
\newblock Scaling up visual and vision-language representation learning with noisy text supervision.
\newblock In Proceedings of the International Conference on Machine Learning, online, 18 July 2021; pp. 4904--4916.

\bibitem[Kim et~al.(2021)Kim, Son and Kim]{ViLT}
Kim, W.; Son, B.; Kim, I.
\newblock Vilt: Vision-and-language transformer without convolution or region supervision.
\newblock In Proceedings of the International Conference on Machine Learning, online, 18 July 2021; pp. 5583--5594.

\bibitem[Chen et~al.(2025)Chen, Wu, Zhang, Li, Zhang, Ma, Yu and Li]{EMLMs}
Chen, S.; Wu, Z.; Zhang, K.; Li, C.; Zhang, B.; Ma, F.; Yu, F.R.; Li, Q.
\newblock Exploring embodied multimodal large models: Development, datasets and future directions.
\newblock {\em Inf. Fusion} {\bf 2025}, \textit{122}, 103198.

\bibitem[Zhang et~al.(2024)Zhang, Yu, Dong, Li, Su, Chu and Yu]{zhang2024mm}
Zhang, D.; Yu, Y.; Dong, J.; Li, C.; Su, D.; Chu, C.; Yu, D.
\newblock Mm-llms: Recent advances in multimodal large language models.
\newblock {\em arXiv} {\bf 2024}, arXiv:2401.13601.

\bibitem[Yin et~al.(2023)Yin, Fu, Zhao, Li, Sun, Xu and Chen]{yin2023survey}
Yin, S.; Fu, C.; Zhao, S.; Li, K.; Sun, X.; Xu, T.; Chen, E.
\newblock A survey on multimodal large language models. \newblock {\em arXiv} {\bf 2023}, arXiv:2306.13549.

\bibitem[Kamath et~al.(2024)Kamath, Keenan, Somers and Sorenson]{Kamath2024}
Kamath, U.; Keenan, K.; Somers, G.; Sorenson, S. Multimodal LLMs.
\newblock In {\em Large Language Models: A Deep Dive: Bridging Theory and Practice}; Springer Nature: Cham, Switzerland, 2024; pp. 375--421.
\newblock {\url{https://doi.org/10.1007/978-3-031-65647-7_9}}.

\bibitem[Wang et~al.(2024)Wang, Chen, Han, Lin, Zhao, Liu, Zhai, Yuan, You and Yang]{wang2024exploring}
Wang, Y.; Chen, W.; Han, X.; Lin, X.; Zhao, H.; Liu, Y.; Zhai, B.; Yuan, J.; You, Q.; Yang, H.
\newblock Exploring the reasoning abilities of multimodal large language models (mllms): A comprehensive survey on emerging trends in multimodal reasoning.
\newblock {\em arXiv} {\bf 2024}, arXiv:2401.06805.

\bibitem[Ahn et~al.(2022)Ahn, Brohan, Brown, Chebotar, Cortes, David, Finn, Fu, Gopalakrishnan, Hausman, et~al.]{DoAsICanNotAsISay}
Ahn, M.; Brohan, A.; Brown, N.; Chebotar, Y.; Cortes, O.; David, B.; Finn, C.; Fu, C.; Gopalakrishnan, K.; Hausman, K.;  et~al.
\newblock Do as i can, not as i say: Grounding language in robotic affordances.
\newblock {\em arXiv} {\bf 2022}, arXiv:2204.01691.

\bibitem[Huang et~al.(2022)Huang, Xia, Xiao, Chan, Liang, Florence, Zeng, Tompson, Mordatch, Chebotar, et~al.]{InnerMonologue}
Huang, W.; Xia, F.; Xiao, T.; Chan, H.; Liang, J.; Florence, P.; Zeng, A.; Tompson, J.; Mordatch, I.; Chebotar, Y.;  et~al.
\newblock Inner monologue: Embodied reasoning through planning with language models.
\newblock {\em arXiv} {\bf 2022}, arXiv:2207.05608.

\bibitem[Zeng et~al.(2022)Zeng, Attarian, Ichter, Choromanski, Wong, Welker, Tombari, Purohit, Ryoo, Sindhwani, et~al.]{SocraticModels}
Zeng, A.; Attarian, M.; Ichter, B.; Choromanski, K.; Wong, A.; Welker, S.; Tombari, F.; Purohit, A.; Ryoo, M.; Sindhwani, V.;  et~al.
\newblock Socratic models: Composing zero-shot multimodal reasoning with language.
\newblock {\em arXiv} {\bf 2022}, arXiv:2204.00598.

\bibitem[Raffel et~al.(2020)Raffel, Shazeer, Roberts, Lee, Narang, Matena, Zhou, Li and Liu]{ExploringTheLimitsOfTransferLearning}
Raffel, C.; Shazeer, N.; Roberts, A.; Lee, K.; Narang, S.; Matena, M.; Zhou, Y.; Li, W.; Liu, P.J.
\newblock Exploring the limits of transfer learning with a unified text-to-text transformer.
\newblock {\em J. Mach. Learn. Res.} {\bf 2020}, {\em 21}, 1--67.
  
\bibitem[Vaswani(2017)]{AttentionIsAllYouNeed}
Vaswani, A.; Vaswani, A.; Shazeer, N.; Parmar, N.; Uszkoreit, J.; Jones, L.; Gomez, A.N.; Kaisek, Ł.; Polosukhin, I. 
\newblock Attention is all you need.
\newblock In \textit{Advances in Neural Information Processing Systems}; Curran Associates Inc.: Red Hook, NY, USA, 2017.  

\bibitem[Alexey(2020)]{ViT}
Alexey, D.
\newblock An image is worth 16x16 words: Transformers for image recognition at scale.
\newblock {\em arXiv} {\bf 2020}, arXiv: 2010.11929.

\bibitem[Ye et~al.(2024)Ye, Dong, Xia, Sun, Zhu, Huang and Wei]{DiffTransformer}
Ye, T.; Dong, L.; Xia, Y.; Sun, Y.; Zhu, Y.; Huang, G.; Wei, F.
\newblock Differential transformer.
\newblock {\em arXiv} {\bf 2024}, arXiv:2410.05258.

\bibitem[Ahmed et~al.(2018)Ahmed, Saint, Shabayek, Cherenkova, Das, Gusev, Aouada and Ottersten]{3dDataRepresentationsSurvey}
Ahmed, E.; Saint, A.; Shabayek, A.E.R.; Cherenkova, K.; Das, R.; Gusev, G.; Aouada, D.; Ottersten, B.
\newblock A survey on deep learning advances on different 3D data representations.
\newblock {\em arXiv} {\bf 2018}, arXiv:1808.01462.

\bibitem[Li et~al.(2023)Li, Zhang, Cho, Waghwase, Lee, Rameau, Yang, Bae and Hong]{GenerativeAIMeets3d}
Li, C.; Zhang, C.; Cho, J.; Waghwase, A.; Lee, L.H.; Rameau, F.; Yang, Y.; Bae, S.H.; Hong, C.S.
\newblock Generative ai meets 3d: A survey on text-to-3d in aigc era.
\newblock {\em arXiv} {\bf 2023}, arXiv:2305.06131.

\bibitem[Bronstein et~al.(2017)Bronstein, Bruna, LeCun, Szlam and Vandergheynst]{GoingBeyondEuclideanData}
Bronstein, M.M.; Bruna, J.; LeCun, Y.; Szlam, A.; Vandergheynst, P.
\newblock Geometric deep learning: Going beyond euclidean data.
\newblock {\em IEEE Signal Process. Mag.} {\bf 2017}, {\em 34}, 18--42.

\bibitem[Lin et~al.(2024)Lin, Peng, Cong, Zheng, Sun, Hou, Zhu, Yang and Ma]{WildRefer}
Lin, Z.; Peng, X.; Cong, P.; Zheng, G.; Sun, Y.; Hou, Y.; Zhu, X.; Yang, S.; Ma, Y.
\newblock Wildrefer: 3d object localization in large-scale dynamic scenes with multi-modal visual data and natural language.
\newblock In Proceedings of the European Conference on Computer Vision, 2024; pp. 456--473.

\bibitem[Yan et~al.(2024)Yan, Zeng, Xiao, Tong, Tan, Fang, Cao and Zhou]{CrossGLG}
Yan, T.; Zeng, W.; Xiao, Y.; Tong, X.; Tan, B.; Fang, Z.; Cao, Z.; Zhou, J.T.
\newblock Crossglg: Llm guides one-shot skeleton-based 3d action recognition in a cross-level manner.
\newblock In Proceedings of the European Conference on Computer Vision, 2024; pp. 113--131.

\bibitem[Li et~al.(2024)Li, Zhang, Wang, Ren, Xu, Ma, Liu and Wei]{3DMIT}
Li, Z.; Zhang, C.; Wang, X.; Ren, R.; Xu, Y.; Ma, R.; Liu, X.; Wei, R.
\newblock 3dmit: 3d multi-modal instruction tuning for scene understanding.
\newblock In Proceedings of the 2024 IEEE International Conference on Multimedia and Expo Workshops (ICMEW), 2024; pp. 1--5.

\bibitem[Yang et~al.(2023)Yang, Liu, Zhang, Pan, Guo, Li, Chen, Gao, Guo and Zhang]{LiDARLLM}
Yang, S.; Liu, J.; Zhang, R.; Pan, M.; Guo, Z.; Li, X.; Chen, Z.; Gao, P.; Guo, Y.; Zhang, S.
\newblock Lidar-llm: Exploring the potential of large language models for 3d lidar understanding.
\newblock {\em arXiv} {\bf 2023}, arXiv:2312.14074.

\bibitem[Jia et~al.(2024)Jia, Chen, Yu, Wang, Niu, Liu, Li and Huang]{SceneVerse}
Jia, B.; Chen, Y.; Yu, H.; Wang, Y.; Niu, X.; Liu, T.; Li, Q.; Huang, S.
\newblock Sceneverse: Scaling 3d vision-language learning for grounded scene understanding.
\newblock In Proceedings of the European Conference on Computer Vision, 2024; pp. 289--310.

\bibitem[Zhang et~al.(2024)Zhang, Huang, Deng, Tang, Ouyang, He and Zhang]{Agent3DZero}
Zhang, S.; Huang, D.; Deng, J.; Tang, S.; Ouyang, W.; He, T.; Zhang, Y.
\newblock Agent3d-zero: An agent for zero-shot 3d understanding.
\newblock In Proceedings of the European Conference on Computer Vision, 2024; pp. 186--202.

\bibitem[Xu et~al.(2024)Xu, Wang, Wang, Chen, Pang and Lin]{PointLLM}
Xu, R.; Wang, X.; Wang, T.; Chen, Y.; Pang, J.; Lin, D.
\newblock Pointllm: Empowering large language models to understand point clouds.
\newblock In Proceedings of the European Conference on Computer Vision, 2024; pp. 131--147.

\bibitem[Mehan et~al.(2024)Mehan, Gupta, Jayanti, Govil, Garg and Krishna]{QueSTMaps}
Mehan, Y.; Gupta, K.; Jayanti, R.; Govil, A.; Garg, S.; Krishna, M.
\newblock QueSTMaps: Queryable Semantic Topological Maps for 3D Scene Understanding.
\newblock {\em arXiv} {\bf 2024}, arXiv:2404.06442.

\bibitem[Wang et~al.(2023)Wang, Huang, Zhao, Zhang and Zhao]{Chat3D}
Wang, Z.; Huang, H.; Zhao, Y.; Zhang, Z.; Zhao, Z.
\newblock Chat-3d: Data-efficiently tuning large language model for universal dialogue of 3d scenes.
\newblock {\em arXiv} {\bf 2023}, arXiv:2308.08769.

\bibitem[Jatavallabhula et~al.(2023)Jatavallabhula, Kuwajerwala, Gu, Omama, Chen, Maalouf, Li, Iyer, Saryazdi, Keetha, et~al.]{ConceptFusion}
Jatavallabhula, K.M.; Kuwajerwala, A.; Gu, Q.; Omama, M.; Chen, T.; Maalouf, A.; Li, S.; Iyer, G.; Saryazdi, S.; Keetha, N.;  et~al.
\newblock Conceptfusion: Open-set multimodal 3d mapping.
\newblock {\em arXiv} {\bf 2023}, arXiv:2302.07241.

\bibitem[Sigurdsson et~al.(2023)Sigurdsson, Thomason, Sukhatme and Piramuthu]{RRExBoT}
Sigurdsson, G.A.; Thomason, J.; Sukhatme, G.S.; Piramuthu, R.
\newblock Rrex-bot: Remote referring expressions with a bag of tricks.
\newblock In Proceedings of the 2023 IEEE/RSJ International Conference on Intelligent Robots and Systems (IROS), 2023; pp. 5203--5210.

\bibitem[Ding et~al.(2023)Ding, Yang, Xue, Zhang, Bai and Qi]{PLA}
Ding, R.; Yang, J.; Xue, C.; Zhang, W.; Bai, S.; Qi, X.
\newblock Pla: Language-driven open-vocabulary 3d scene understanding.
\newblock In Proceedings of the IEEE/CVF Conference on Computer Vision and Pattern Recognition, 2023; pp. 7010--7019.

\bibitem[Peng et~al.(2023)Peng, Genova, Jiang, Tagliasacchi, Pollefeys, Funkhouser, et~al.]{OpenScene}
Peng, S.; Genova, K.; Jiang, C.; Tagliasacchi, A.; Pollefeys, M.; Funkhouser, T.
\newblock Openscene: 3d scene understanding with open vocabularies.
\newblock In Proceedings of the IEEE/CVF Conference on Computer Vision and Pattern Recognition, 2023; pp. 815--824.

\bibitem[Xue et~al.(2023)Xue, Gao, Xing, Mart{\'\i}n-Mart{\'\i}n, Wu, Xiong, Xu, Niebles and Savarese]{ULIP}
Xue, L.; Gao, M.; Xing, C.; Mart{\'\i}n-Mart{\'\i}n, R.; Wu, J.; Xiong, C.; Xu, R.; Niebles, J.C.; Savarese, S.
\newblock Ulip: Learning a unified representation of language, images and point clouds for 3d understanding.
\newblock In Proceedings of the IEEE/CVF Conference on Computer Vision and Pattern Recognition, 2023; pp. 1179--1189.

\bibitem[Chen et~al.(2023)Chen, Garcia, Schmid and Laptev]{PolarNet}
Chen, S.; Garcia, R.; Schmid, C.; Laptev, I.
\newblock Polarnet: 3d point clouds for language-guided robotic manipulation.
\newblock {\em arXiv} {\bf 2023}, arXiv:2309.15596.

\bibitem[Giuili et~al.(2025)Giuili, Atari and Sintov]{OracleGrasp}
Giuili, A.; Atari, R.; Sintov, A.
\newblock ORACLE-Grasp: Zero-Shot Task-Oriented Robotic Grasping using Large Multimodal Models.
\newblock {\em arXiv} {\bf 2025}, arXiv:2505.08417.

\bibitem[Zheng et~al.(2025)Zheng, Huang, Li and Wang]{EnhancingMLLMsWith3DVision}
Zheng, D.; Huang, S.; Li, Y.; Wang, L.
\newblock Learning from Videos for 3D World: Enhancing MLLMs with 3D Vision Geometry Priors.
\newblock {\em arXiv} {\bf 2025}, arXiv:2505.24625.

\bibitem[Wang et~al.(2024)Wang, Yu, Jiang, Lan, Shi, Chang, Kautz, Li and Alvarez]{OmniDrive}
Wang, S.; Yu, Z.; Jiang, X.; Lan, S.; Shi, M.; Chang, N.; Kautz, J.; Li, Y.; Alvarez, J.M.
\newblock OmniDrive: A Holistic LLM-Agent Framework for Autonomous Driving with 3D Perception, Reasoning and Planning.
\newblock {\em arXiv} {\bf 2024}, arXiv:2405.01533.

\bibitem[Bai et~al.(2024)Bai, Wu, Liu, Jia, Mao, Zhang, Zhao, Shen, Wei, Wang, et~al.]{Atlas}
Bai, Y.; Wu, D.; Liu, Y.; Jia, F.; Mao, W.; Zhang, Z.; Zhao, Y.; Shen, J.; Wei, X.; Wang, T.;  et~al.
\newblock Is a 3d-tokenized llm the key to reliable autonomous driving?
\newblock {\em arXiv} {\bf 2024}, arXiv:2405.18361.

\bibitem[Yang et~al.(2025)Yang, Chen, Madaan, Iyengar, Qian, Fouhey and Chai]{3DGrand}
Yang, J.; Chen, X.; Madaan, N.; Iyengar, M.; Qian, S.; Fouhey, D.F.; Chai, J.
\newblock 3d-grand: A million-scale dataset for 3d-llms with better grounding and less hallucination.
\newblock In Proceedings of the Computer Vision and Pattern Recognition Conference, 2025; \mbox{pp. 29501--29512.}

\bibitem[Qian et~al.(2022)Qian, Li, Peng, Mai, Hammoud, Elhoseiny and Ghanem]{PointNext}
Qian, G.; Li, Y.; Peng, H.; Mai, J.; Hammoud, H.; Elhoseiny, M.; Ghanem, B.
\newblock Pointnext: Revisiting pointnet++ with improved training and scaling strategies.
\newblock In \textit{Advances in Neural Information Processing Systems}; Curran Associates Inc.: Red Hook, NY, USA, 2022; Volume 35, pp. 23192--23204.

\bibitem[Liu et~al.(2022)Liu, Li, Ling, Li and Su]{FrameMining}
Liu, M.; Li, X.; Ling, Z.; Li, Y.; Su, H.
\newblock Frame mining: A free lunch for learning robotic manipulation from 3d point clouds.
\newblock {\em arXiv} {\bf 2022}, arXiv:2210.07442.

\bibitem[Sun et~al.(2023)Sun, Han, Deng, et~al.]{3DGPT}
Sun, C.; Han, J.; Deng, W.; Wang, X.; Qin, Z.; Gould, S.
\newblock 3D-GPT: Procedural 3D Modeling with Large Language Models.
\newblock {\em arXiv} {\bf 2023}, arXiv:2310.12945.

\bibitem[Kuka()]{MultimodalFoundationModels}
Kuka, V.
\newblock Multimodal Foundation Models: 2024's Surveys to Understand the Future of AI.
\newblock  Available online: \url{https://www.turingpost.com/p/multimodal-resources} (accessed on 1 October 2025).

\bibitem[Han et~al.(2025)Han, Chen, Fu, Feng, Fan, An, Wang, Guo, Meng, Zhang, et~al.]{MultimodalFusionAndVLMs}
Han, X.; Chen, S.; Fu, Z.; Feng, Z.; Fan, L.; An, D.; Wang, C.; Guo, L.; Meng, W.; Zhang, X.;  et~al.
\newblock Multimodal fusion and vision-language models: A survey for robot vision.
\newblock {\em Inf. Fusion} \textbf{2026}, \textit{126}, 103652.

\bibitem[Li and Tang(2024)]{MultimodalAlignmentAndFusionSurvey}
Li, S.; Tang, H.
\newblock Multimodal alignment and fusion: A survey.
\newblock {\em arXiv} {\bf 2024}, arXiv:2411.17040.

\bibitem[Lin et~al.(2016)Lin, Hsieh and Chang]{DSPBasedVisualServoing}
Lin, C.Y.; Hsieh, P.J.; Chang, F.A.
\newblock Dsp based uncalibrated visual servoing for a 3-dof robot manipulator.
\newblock In Proceedings of the 2016 IEEE International Conference on Industrial Technology (ICIT), 2016; pp. 1618--1621.

\bibitem[Martinez González et~al.(2016)Martinez González, Castelán and Arechavaleta]{VisionBasedLocalization}
Martinez González, P.; Castelán, M.; Arechavaleta, G.
\newblock Vision Based Persistent Localization of a Humanoid Robot for Locomotion Tasks.
\newblock {\em Int. J. Appl. Math. Comput. Sci.} {\bf 2016}, {\em 26}, 669--682.
\newblock {\url{https://doi.org/10.1515/amcs-2016-0046}}.

\bibitem[Garcia-Garcia et~al.(2017)Garcia-Garcia orts, Oprea, Rodríguez, Azorin-Lopez, Saval-Calvo and Cazorla]{FullPoseEstimation}
Garcia-Garcia, A.; Orts, S.; Oprea, S.; Rodríguez, J.; Azorin-Lopez, J.; Saval-Calvo, M.; Cazorla, M.
\newblock Multi-sensor 3D Object Dataset for Object Recognition with Full Pose Estimation.
\newblock {\em Neural Comput. Appl.} {\bf 2017}, {\em 28}, 941–952.
\newblock {\url{https://doi.org/10.1007/s00521-016-2224-9}}.

\bibitem[Zhang et~al.(2016)Zhang, Guo, Chen and Shuai]{CornerBased3DObjectPoseEstimation}
Zhang, L.; Guo, Z.; Chen, H.; Shuai, L.
\newblock Corner-Based 3D Object Pose Estimation in Robot Vision.
\newblock  In Proceedings of the 2016 8th International Conference on Intelligent Human-Machine Systems and Cybernetics (IHMSC), Hangzhou, China,  27--28 August 2016; pp. 363--368.
\newblock {\url{https://doi.org/10.1109/IHMSC.2016.200}}.

\bibitem[Wen et~al.(2016)Wen, Wu, Kong and Liu]{KalmanFilterForRobotEgoMotionEstimation}
Wen, K.; Wu, W.; Kong, X.; Liu, K.
\newblock A Comparative Study of the Multi-state Constraint and the Multi-view Geometry Constraint Kalman Filter for Robot Ego-Motion Estimation. In Proceedings of the 2016 8th International Conference on Intelligent Human-Machine Systems and Cybernetics (IHMSC), Hangzhou, China, 27--28 August 2016; pp. 466--471.

\bibitem[Vicente et~al.(2016)Vicente, Jamone and Bernardino]{RobotHandPoseEstimation}
Vicente, P.; Jamone, L.; Bernardino, A.
\newblock Robotic Hand Pose Estimation Based on Stereo Vision and GPU-enabled Internal Graphical Simulation.
\newblock {\em J. Intell. Robot. Syst.} {\bf 2016}, {\em 83}, 339--358.

\bibitem[Masuta et~al.(2016)Masuta, Motoyoshi, Koyanagi, Oshima and Lim]{DirectPerceptionObjectGrasping}
Masuta, H.; Motoyoshi, T.; Koyanagi, K.; Oshima, T.; Lim, H.o.
\newblock Direct Perception of Easily Visible Information for Unknown Object Grasping. In Proceedings of the 9th International Conference on Intelligent Robotics and Applications, ICIRA 2016, Tokyo, Japan, 22--24 August 2016; Volume 9835, pp. 78--89.
\newblock {\url{https://doi.org/10.1007/978-3-319-43518-3_8}}.

\bibitem[Kent et~al.(2015)Kent, Behrooz and Chernova]{ConstructionOfA3DObject}
Kent, D.; Behrooz, M.; Chernova, S.
\newblock Construction of a 3D object recognition and manipulation database from grasp demonstrations.
\newblock {\em Auton. Robot.} {\bf 2015}, {\em 40}, 175--192.
\newblock {\url{https://doi.org/10.1007/s10514-015-9451-2}}.

\bibitem[Gunatilake et~al.(2021)Gunatilake, Piyathilaka, Tran, Vishwanathan, Thiyagarajan and Kodagoda]{Gunatilake2021}
Gunatilake, A.; Piyathilaka, L.; Tran, A.; Vishwanathan, V.K.; Thiyagarajan, K.; Kodagoda, S.
\newblock Stereo Vision Combined with Laser Profiling for Mapping of Pipeline Internal Defects.
\newblock {\em IEEE Sens. J.} {\bf 2021}, {\em 21}, 11926–11934.
\newblock {\url{https://doi.org/10.1109/jsen.2020.3040396}}.

\bibitem[Du et~al.(2020)Du, Muslikhin, Hsieh and Wang]{CloseRangeManipulationStereoVision}
Du, Y.C.; Muslikhin, M.; Hsieh, T.H.; Wang, M.S.
\newblock Stereo Vision-Based Object Recognition and Manipulation by Regions with Convolutional Neural Network.
\newblock {\em Electronics} {\bf 2020}, {\em 9}, 210.
\newblock {\url{https://doi.org/10.3390/electronics9020210}}.

\bibitem[Haque and Nejadpak(2017)]{ObstacleAvoidanceStereoVision1}
Haque, A.U.; Nejadpak, A.
\newblock Obstacle avoidance using stereo camera.
\newblock {\em arXiv} {\bf 2017}, arXiv:1705.04114.

\bibitem[Kumano et~al.(2000)Kumano, Ohya and Yuta]{ObstacleAvoidanceStereoVision2}
Kumano, M.; Ohya, A.; Yuta, S.
\newblock Obstacle avoidance of autonomous mobile robot using stereo vision sensor.
\newblock {In Proceedings of the 2nd International Symposium on Robotics and Automation}, 2000.

\bibitem[Howard(2008)]{OdometryStereoVision1}
Howard, A.
\newblock Real-time stereo visual odometry for autonomous ground vehicles.
\newblock In Proceedings of the 2008 IEEE/RSJ International Conference on Intelligent Robots and Systems, 2008; pp. 3946--3952.
\newblock {\url{https://doi.org/10.1109/IROS.2008.4651147}}.

\bibitem[Zhao et~al.(2019)Zhao, Luo and Zhang]{OdometryStereoVision2}
Zhao, Q.; Luo, B.; Zhang, Y.
\newblock Stereo-based multi-motion visual odometry for mobile robots.
\newblock {\em arXiv} {\bf 2019}, arXiv:1910.06607.

\bibitem[Nguyen et~al.(2017)Nguyen, Park, Cho, Kim and Kim]{3dReconstructionStereoVision}
Nguyen, C.D.T.; Park, J.; Cho, K.Y.; Kim, K.S.; Kim, S.
\newblock Novel Descattering Approach for Stereo Vision in Dense Suspended Scatterer Environments.
\newblock {\em Sensors} {\bf 2017}, {\em 17}, 1425.
\newblock {\url{https://doi.org/10.3390/s17061425}}.

\bibitem[He and Chen(2018)]{AdvancesInSensingAndProcessing}
He, Y.; Chen, S.
\newblock Advances in sensing and processing methods for three-dimensional robot vision.
\newblock {\em Int. J. Adv. Robot. Syst.} {\bf 2018}, {\em 15}, 1729881418760623.
\newblock {\url{https://doi.org/10.1177/1729881418760623}}.

\bibitem[Hebert and Krotkov(1992)]{3DMeasurementsFromImagingLaserRadars}
Hebert, M.; Krotkov, E.
\newblock 3D measurements from imaging laser radars: How good are they?
\newblock {\em Image Vis. Comput.} {\bf 1992}, {\em 10}, 170--178.

\bibitem[Honegger et~al.(2013)Honegger, Meier, Tanskanen and Pollefeys]{CMOSCameraForIndoorAndOutdoor}
Honegger, D.; Meier, L.; Tanskanen, P.; Pollefeys, M.
\newblock An open source and open hardware embedded metric optical flow cmos camera for indoor and outdoor applications.
\newblock In Proceedings of the 2013 IEEE International Conference on Robotics and Automation, 2013; pp. 1736--1741.

\bibitem[Besl(1988)]{ActiveOpticalRangeImagingSensors}
Besl, P.J.
\newblock Active, optical range imaging sensors.
\newblock {\em Mach. Vis. Appl.} {\bf 1988}, {\em 1}, 127--152.

\bibitem[Kim et~al.(2010)Kim, Ahn and Cho]{BayesianSensor}
Kim, M.Y.; Ahn, S.T.; Cho, H.S.
\newblock Bayesian sensor fusion of monocular vision and laser structured light sensor for robust localization of a mobile robot.
\newblock {\em J. Inst. Control Robot. Syst.} {\bf 2010}, {\em 16}, 381--390.

\bibitem[Yang and Chen(2010)]{DesignOfA3DInfrared}
Yang, R.; Chen, Y.
\newblock Design of a 3-D infrared imaging system using structured light.
\newblock {\em IEEE Trans. Instrum. Meas.} {\bf 2010}, {\em 60}, 608--617.

\bibitem[Cui et~al.(2010)Cui, Schuon, Chan, Thrun and Theobalt]{3DShapeScanning}
Cui, Y.; Schuon, S.; Chan, D.; Thrun, S.; Theobalt, C.
\newblock 3D shape scanning with a time-of-flight camera.
\newblock In Proceedings of the 2010 IEEE Computer Society Conference on Computer Vision and Pattern Recognition, 2010; pp. 1173--1180.

\bibitem[Sansoni et~al.(2009)Sansoni, Trebeschi and Docchio]{SOTA3DImagingSensors}
Sansoni, G.; Trebeschi, M.; Docchio, F.
\newblock State-of-The-Art and Applications of 3D Imaging Sensors in Industry, Cultural Heritage, Medicine and Criminal Investigation.
\newblock {\em Sensors} {\bf 2009}, {\em 9}, 568--601.
\newblock {\url{https://doi.org/10.3390/s90100568}}.

\bibitem[Boyer and Kak(1987)]{ColorEncodedStructuredLight}
Boyer, K.L.; Kak, A.C.
\newblock Color-encoded structured light for rapid active ranging.
\newblock {\em IEEE Trans. Pattern Anal. Mach. Intell.} {\bf 1987}, \textit{PAMI-9}, 14--28.

\bibitem[Chen et~al.(2008)Chen, Li and Zhang]{VisionProcessingForRealtime}
Chen, S.; Li, Y.F.; Zhang, J.
\newblock Vision processing for realtime 3-D data acquisition based on coded structured light.
\newblock {\em IEEE Trans. Image Process.} {\bf 2008}, {\em 17}, 167--176.

\bibitem[Wei et~al.(2009)Wei, Gao, Li, Fan, Gao and Gao]{IndoorMobileRobotObstacle}
Wei, B.; Gao, J.; Li, K.; Fan, Y.; Gao, X.; Gao, B.
\newblock Indoor mobile robot obstacle detection based on linear structured light vision system.
\newblock In Proceedings of the 2008 IEEE International Conference on Robotics and Biomimetics, 2009; pp. 834--839.

\bibitem[King and Weiman(1991)]{HelpmateAutonomousMobileRobot}
King, S.J.; Weiman, C.F.
\newblock Helpmate autonomous mobile robot navigation system.
\newblock In \textit{Mobile Robots V, Proceedings of the Advances in Intelligent Robotics Systems, Boston, MA, USA, 4--9 November 1990}; SPIE: Bellingham, WA,  USA, 1991; Volume 1388, pp. 190--198.

\bibitem[Silberman and Fergus(2011)]{IndoorSceneSegmentation}
Silberman, N.; Fergus, R.
\newblock Indoor scene segmentation using a structured light sensor.
\newblock In Proceedings of the 2011 IEEE International Conference on Computer Vision Workshops (ICCV Workshops), 2011; pp. 601--608.

\bibitem[Sarafraz and Haus(2016)]{StructuredLightUnderwaterSurface}
Sarafraz, A.; Haus, B.K.
\newblock A structured light method for underwater surface reconstruction.
\newblock {\em ISPRS J. Photogramm. Remote Sens.} {\bf 2016}, {\em 114}, 40--52.

\bibitem[Johnson-Roberson et~al.(2017)Johnson-Roberson, Bryson, Friedman, Pizarro, Troni, Ozog and Henderson]{HighResolutionUnderwater}
Johnson-Roberson, M.; Bryson, M.; Friedman, A.; Pizarro, O.; Troni, G.; Ozog, P.; Henderson, J.C.
\newblock High-resolution underwater robotic vision-based mapping and three-dimensional reconstruction for archaeology.
\newblock {\em J. Field Robot.} {\bf 2017}, {\em 34}, 625--643.

\bibitem[Zhang et~al.(2002)Zhang, Curless and Seitz]{RapidShapeAcquisition}
Zhang, L.; Curless, B.; Seitz, S.M.
\newblock Rapid shape acquisition using color structured light and multi-pass dynamic programming.
\newblock In Proceedings of the First International Symposium on 3D Data Processing Visualization and Transmission, 2002; \mbox{pp. 24--36.}

\bibitem[Park et~al.(2001)Park, DeSouza and Kak]{DualBeamStructuredLight}
Park, J.; DeSouza, G.N.; Kak, A.C.
\newblock Dual-beam structured-light scanning for 3-D object modeling.
\newblock In Proceedings of the Third International Conference on 3-D Digital Imaging and Modeling, 2001; pp. 65--72.

\bibitem[Park and Kim(2006)]{3DHandEyeRobotVisionSystem}
Park, D.J.; Kim, J.H.
\newblock 3D hand-eye robot vision system using a cone-shaped structured light for the SICE-ICASE International Joint Conference 2006 (SICE-ICCAS 2006).
\newblock In Proceedings of the 2006 SICE-ICASE International Joint Conference, 2006; pp. 2975--2980.

\bibitem[Stuff()]{LidarComparison1}
Stuff, A.
\newblock LiDAR Comparison Chart.
\newblock  Available online: \url{https://autonomoustuff.com/lidar-chart} (accessed on 25 April 2025).

\bibitem[Robots()]{LidarComparison2}
Robots, G.
\newblock How to Select the Right LiDAR.
\newblock
Available online: \url{https://www.generationrobots.com/blog/en/how-to-select-the-right-lidar/} (accessed on 25 April 2025).

\bibitem[Ma et~al.(2024)Ma, Fang, Jiang, Su, Zhang, Wu and Wang]{GestureRecognitionToF}
Ma, Y.; Fang, Z.; Jiang, W.; Su, C.; Zhang, Y.; Wu, J.; Wang, Z.
\newblock Gesture Recognition Based on Time-of-Flight Sensor and Residual Neural Network.
\newblock {\em J. Comput. Commun.} {\bf 2024}, {\em 12}, 103--114.
\newblock {\url{https://doi.org/10.4236/jcc.2024.126007}}.

\bibitem[Farhangian et~al.(2021)Farhangian, Sefidgar and Landry]{IndoorNavigationToF1}
Farhangian, F.; Sefidgar, M.; Landry, R.J.
\newblock Applying a ToF/IMU-Based Multi-Sensor Fusion Architecture in Pedestrian Indoor Navigation Methods.
\newblock {\em Sensors} {\bf 2021}, {\em 21}, 3615.
\newblock {\url{https://doi.org/10.3390/s21113615}}.

\bibitem[Bostelman et~al.(2005)Bostelman, Hong and Madhavan]{IndoorNavigationToF2ObstacleDetectionToF}
Bostelman, R.; Hong, T.; Madhavan, R.
\newblock Obstacle detection using a Time-of-Flight range camera for Automated Guided Vehicle safety and navigation.
\newblock {\em Integr. Comput.-Aided Eng.} {\bf 2005}, {\em 12}, 237--249.
\newblock {\url{https://doi.org/10.3233/ICA-2005-12303}}.

\bibitem[Horio et~al.(2022)Horio, Feng, Kokado, Takasawa, Yasutomi, Kawahito, Komuro, Nagahara and Kagawa]{MultiPathInterferenceToF}
Horio, M.; Feng, Y.; Kokado, T.; Takasawa, T.; Yasutomi, K.; Kawahito, S.; Komuro, T.; Nagahara, H.; Kagawa, K.
\newblock Resolving Multi-Path Interference in Compressive Time-of-Flight Depth Imaging with a Multi-Tap Macro-Pixel Computational CMOS Image Sensor.
\newblock {\em Sensors} {\bf 2022}, {\em 22}, 2442.
\newblock {\url{https://doi.org/10.3390/s22072442}}.

\bibitem[Li et~al.(2017)Li, Chen, Pediredla, Yeh, He, Veeraraghavan and Cossairt]{LowSpatialResolutionToF}
Li, F.; Chen, H.; Pediredla, A.; Yeh, C.; He, K.; Veeraraghavan, A.; Cossairt,
  O.
\newblock CS-ToF: High-resolution compressive time-of-flight imaging.
\newblock {\em Opt. Express} {\bf 2017}, {\em 25}, 31096--31110.
\newblock {\url{https://doi.org/10.1364/OE.25.031096}}.

\bibitem[Lee et~al.(2020)Lee, Yasutomi, Morita, Kawanishi and Kawahito]{ToFSensorUsingFourTapLockInPixels}
Lee, S.; Yasutomi, K.; Morita, M.; Kawanishi, H.; Kawahito, S.
\newblock A Time-of-Flight Range Sensor Using Four-Tap Lock-In Pixels with High near Infrared Sensitivity for LiDAR Applications.
\newblock {\em Sensors} {\bf 2020}, {\em 20}, 116.
\newblock {\url{https://doi.org/10.3390/s20010116}}.

\bibitem[Agresti and Zanuttigh(2018)]{DLForMultiPathErrorRemovalInToFSensors}
Agresti, G.; Zanuttigh, P.
\newblock Deep Learning for Multi-Path Error Removal in ToF Sensors.
\newblock In Proceedings of the European Conference on Computer Vision (ECCV) Workshops,  September 2018.

\bibitem[Son et~al.(2016)Son, Liu and Taguchi]{LearningToRemoveMultipathDistortions}
Son, K.; Liu, M.Y.; Taguchi, Y.
\newblock Learning to remove multipath distortions in Time-of-Flight range images for a robotic arm setup.
\newblock In Proceedings of the 2016 IEEE International Conference on Robotics and Automation (ICRA),  2016; pp. 3390--3397.
\newblock {\url{https://doi.org/10.1109/ICRA.2016.7487515}}.

\bibitem[Su et~al.(2018)Su, Heide, Wetzstein and Heidrich]{DeepEndToEndToFImaging}
Su, S.; Heide, F.; Wetzstein, G.; Heidrich, W.
\newblock Deep End-to-End Time-of-Flight Imaging.
\newblock In Proceedings of the 2018 IEEE/CVF Conference on Computer Vision and Pattern Recognition, 2018; pp. 6383--6392.
\newblock {\url{https://doi.org/10.1109/CVPR.2018.00668}}.

\bibitem[Du et~al.(2025)Du, Deng, Zhou, Jiao, Wang, Xu, Jiang and Guan]{MultipathInterferenceSuppression}
Du, Y.; Deng, Y.; Zhou, Y.; Jiao, F.; Wang, B.; Xu, Z.; Jiang, Z.; Guan, X.
\newblock Multipath interference suppression in indirect time-of-flight imaging via a novel compressed sensing framework.
\newblock {\em arXiv} {\bf 2025}, arXiv:2507.19546.

\bibitem[Buratto et~al.(2021)Buratto, Simonetto, Agresti, Schäfer and Zanuttigh]{DLForTransientImageReconstruction}
Buratto, E.; Simonetto, A.; Agresti, G.; Schäfer, H.; Zanuttigh, P.
\newblock Deep Learning for Transient Image Reconstruction from ToF Data.
\newblock {\em Sensors} {\bf 2021}, {\em 21}, 1962.
\newblock {\url{https://doi.org/10.3390/s21061962}}.

\bibitem[Liu and Zhang(2008)]{OutdoorMappingLidar}
Liu, X.; Zhang, Z.
\newblock LiDAR data reduction for efficient and high quality DEM generation.
\newblock {\em Int. Arch. Photogramm. Remote Sens. Spat. Inf. Sci.} {\bf 2008}, {\em 37}, 173--178.

\bibitem[Yu et~al.()Yu, Zhang, Zhong, Nguyen and Nguyen]{MobileRobotNavigationLidar}
Yu, J.; Zhang, A.; Zhong, Y.; Nguyen, T.T.; Nguyen, T.D.
\newblock An Indoor Mobile Robot 2D Lidar Mapping Based on Cartographer-Slam Algorithm. \textit{Taiwan Ubiquitous Inf.} \textbf{2022}, \textit{7}, 795--804.

\bibitem[Kim et~al.(2023)Kim, Park and Kim]{AtmosphericInterferenceLidar}
Kim, J.; Park, B.j.; Kim, J.
\newblock Empirical Analysis of Autonomous Vehicle’s LiDAR Detection Performance Degradation for Actual Road Driving in Rain and Fog.
\newblock {\em Sensors} {\bf 2023}, {\em 23}, 2972.
\newblock {\url{https://doi.org/10.3390/s23062972}}.

\bibitem[Otgonbayar et~al.(2024)Otgonbayar, Kim, Jekal, Kim, Noh, Oh and Yoon]{MaterialChallengesLidar}
Otgonbayar, Z.; Kim, J.; Jekal, S.; Kim, C.G.; Noh, J.; Oh, W.C.; Yoon, C.M.
\newblock Designing a highly near infrared-reflective black nanoparticles for autonomous driving based on the refractive index and principle.
\newblock {\em J. Colloid Interface Sci.} {\bf 2024}, {\em 667}, 663--678.
\newblock {\url{https://doi.org/10.1016/j.jcis.2024.04.133}}.

\bibitem[IntertialLabs()]{Why905and1550nm}
IntertialLabs.
\newblock Why Have 905 and 1550 nm Become the Standard for LiDAR?
\newblock
Available online:  \url{https://inertiallabs.com/why-have-905-and-1550-nm-become-the-standard-for-lidars/} (accessed on 1 October 2025).

\bibitem[Eureka()]{LidarSystems}
Eureka.
\newblock LIDAR Systems: 905nm vs 1550nm Laser Diode Safety and Performance.
\newblock
Available online: \url{https://eureka.patsnap.com/article/lidar-systems-905nm-vs-1550nm-laser-diode-safety-and-performance} (accessed on 1 October 2025).

\bibitem[Jin et~al.(2023)Jin, Wang, Zhang, Mei and Wang]{TactileSensorsComparison}
Jin, J.; Wang, S.; Zhang, Z.; Mei, D.; Wang, Y.
\newblock Progress on flexible tactile sensors in robotic applications on objects properties recognition, manipulation and human-machine interactions.
\newblock {\em Soft Sci.} {\bf 2023}, {\em 3}, 8.
\newblock {\url{https://doi.org/10.20517/ss.2022.34}}.

\bibitem[Wiki()]{BioTac}
Wiki, R.
\newblock The BioTac, Multimodal Tactile Sensor.
\newblock  Available online: \url{https://wiki.ros.org/BioTac} (accessed on 25 April 2025).

\bibitem[Sight()]{GelSightMini}
Sight, G.
\newblock Gel Sight Datasheet.
\newblock
Available online:  \url{https://www.gelsight.com/wp-content/uploads/2022/09/GelSight_Datasheet_GSMini_9.20.22b.pdf} (accessed on 25 April 2025).

\bibitem[Robots()]{ShadowRobot}
Robots, S.
\newblock Tactile Telerobot, The Story So Far.
\newblock
Available online:  \url{https://www.shadowrobot.com/blog/tactile-telerobot-the-story-so-far/} (accessed on 25 April 2025).

\bibitem[Yu et~al.(2021)Yu, Zhang and Deng]{ProgressInTemperatureTactileSensors}
Yu, J.; Zhang, K.; Deng, Y.
\newblock Recent progress in pressure and temperature tactile sensors: principle, classification, integration and outlook.
\newblock {\em Soft Sci.} {\bf 2021}, {\em 1}, 6.
\newblock {\url{https://doi.org/10.20517/ss.2021.05}}.

\bibitem[Xia et~al.(2025)Xia, Zhou and Oztireli]{RETRO}
Xia, W.; Zhou, C.; Oztireli, C.
\newblock RETRO: REthinking Tactile Representation Learning with Material PriOrs.
\newblock {\em arXiv} {\bf 2025}, arXiv:2505.14319.

\bibitem[Hu et~al.(2025)Hu, Hershcovich and Seifi]{Hapticcap}
Hu, G.; Hershcovich, D.; Seifi, H.
\newblock Hapticcap: A multimodal dataset and task for understanding user experience of vibration haptic signals.
\newblock {\em arXiv} {\bf 2025}, arXiv:2507.13318.

\bibitem[Fu et~al.(2024)Fu, Datta, Huang, Panitch, Drake ortiz, Mukadam, Lambeta, Calandra and Goldberg]{TVL}
Fu, L.; Datta, G.; Huang, H.; Panitch, W.C.H.; Drake, J.; Ortiz, J.; Mukadam, M.; Lambeta, M.; Calandra, R.; Goldberg, K.
\newblock A touch, vision and language dataset for multimodal alignment.
\newblock {\em arXiv} {\bf 2024}, arXiv:2402.13232.

\bibitem[Cao et~al.(2024)Cao, Jiang, Bollegala, Li and Luo]{MultimodalZeroShotLearningForTactileTextureRecognition}
Cao, G.; Jiang, J.; Bollegala, D.; Li, M.; Luo, S.
\newblock Multimodal zero-shot learning for tactile texture recognition.
\newblock {\em Robot. Auton. Syst.} {\bf 2024}, {\em 176}, 104688.
\newblock {\url{https://doi.org/10.1016/j.robot.2024.104688}}.

\bibitem[Tejwani et~al.(2025)Tejwani, Velazquez, Payne, Bonato and Asada]{CrossModalityForceAndLanguageEmbeddings}
Tejwani, R.; Velazquez, K.; Payne, J.; Bonato, P.; Asada, H.
\newblock Cross-modality Force and Language Embeddings for Natural Human-Robot Communication.
\newblock {\em arXiv} {\bf 2025}, arXiv:2502.02772.

\bibitem[Hu et~al.(2009)Hu, Chan, Wang and Wang]{SoundSourceLocalization}
Hu, J.S.; Chan, C.Y.; Wang, C.K.; Wang, C.C.
\newblock Simultaneous localization of mobile robot and multiple sound sources using microphone array.
\newblock In Proceedings of the 2009 IEEE International Conference on Robotics and Automation, 2009; pp. 29--34.
\newblock {\url{https://doi.org/10.1109/ROBOT.2009.5152813}}.

\bibitem[Albustanji et~al.(2023)Albustanji, Elmanaseer and Alkhatib]{SpeechRecognitionMFCC}
Albustanji, R.N.; Elmanaseer, S.; Alkhatib, A.A.A.
\newblock Robotics: Five Senses plus One, An Overview.
\newblock {\em Robotics} {\bf 2023}, {\em 12}, 68.
\newblock {\url{https://doi.org/10.3390/robotics12030068}}.

\bibitem[Guo et~al.(2022)Guo, Ding, ti, Li and Hong]{CocktailPartyProblem}
Guo, X.; Ding, S.; Peng, T.; Li, K.; Hong, X.
\newblock Robot Hearing Through Optical. Channel in a Cocktail Party Environment. \textit{Authorea} {\bf 2022}.
\newblock {\url{https://doi.org/10.22541/au.165556539.94506371/v1}}.

\bibitem[Engineering()]{SearchAndRescueThermal}
Engineering, T.
\newblock How Does Thermal Vision Technology Assist in Search and Rescue.
\newblock
Available online:  \url{https://www.thermal-engineering.org/how-does-thermal-vision-technology-assist-in-search-and-rescue/} (accessed on 25 April 2025).

\bibitem[Castro Jiménez and Martínez-García(2016)]{PlanningAndSearch}
Castro Jiménez, L.E.; Martínez-García, E.A.
\newblock Thermal Image Sensing Model for Robotic Planning and Search.
\newblock {\em Sensors} {\bf 2016}, \mbox{{\em 16}, 1253.}
\newblock {\url{https://doi.org/10.3390/s16081253}}.

\bibitem[Chiu et~al.(2023)Chiu, Tseng and Chen]{LowSpatialResolutionThermal}
Chiu, S.Y.; Tseng, Y.C.; Chen, J.J.
\newblock Low-Resolution Thermal Sensor-Guided Image Synthesis.
\newblock In Proceedings of the 2023 IEEE/CVF Winter Conference on Applications of Computer Vision Workshops (WACVW), 2023; pp. 60--69.
\newblock {\url{https://doi.org/10.1109/WACVW58289.2023.00011}}.

\bibitem[Sun et~al.(2019)Sun, Zuo and Liu]{RTFNet}
Sun, Y.; Zuo, W.; Liu, M.
\newblock RTFNet: RGB-Thermal Fusion Network for Semantic Segmentation of Urban Scenes.
\newblock {\em IEEE Robot. Autom. Lett.} {\bf 2019}, {\em 4}, 2576--2583.
\newblock {\url{https://doi.org/10.1109/LRA.2019.2904733}}.

\bibitem[Zhao et~al.(2024)Zhao, Huang, Yan, Wang, Tang, Ou, Hu and Peng]{OpenRSS}
Zhao, G.; Huang, J.; Yan, X.; Wang, Z.; Tang, J.; Ou, Y.; Hu, X.; Peng, T.
\newblock Open-Vocabulary RGB-Thermal Semantic Segmentation.
\newblock In \textit{Computer Vision, ECCV 2024: 18th European Conference, Milan, Italy, September 29–October 4, 2024, Proceedings, Part LXXIV}; Springer: Berlin/Heidelberg, Germany, 2024; pp. 304–320.
\newblock {\url{https://doi.org/10.1007/978-3-031-72904-1_18}}.

\bibitem[Redmon et~al.(2016)Redmon, Divvala, Girshick and Farhadi]{Yolo}
Redmon, J.; Divvala, S.; Girshick, R.; Farhadi, A.
\newblock You only look once: Unified, real-time object detection.
\newblock In Proceedings of the IEEE Conference on Computer Vision and Pattern Recognition, 2016; pp. 779--788.

\bibitem[Qi et~al.(2017{\natexlab{a}})Qi, Su, Mo and Guibas]{PointNet}
Qi, C.R.; Su, H.; Mo, K.; Guibas, L.J.
\newblock Pointnet: Deep learning on point sets for 3d classification and segmentation.
\newblock In Proceedings of the IEEE Conference on Computer Vision and Pattern Recognition, 2017; pp. 652--660.

\bibitem[Qi et~al.(2017{\natexlab{b}})Qi, Yi, Su and Guibas]{PointNet++}
Qi, C.R.; Yi, L.; Su, H.; Guibas, L.J.
\newblock Pointnet++: Deep hierarchical feature learning on point sets in a metric space. In \textit{Advances in Neural Information Processing Systems}; 	
Curran Associates Inc.: Red Hook, NY, USA, 2017; Volume 30.

\bibitem[Driess et~al.(2023)Driess, Xia, Sajjadi, Lynch, Chowdhery, Ichter, Wahid, Tompson, Vuong, Yu, et~al.]{PALME}
Driess, D.; Xia, F.; Sajjadi, M.S.; Lynch, C.; Chowdhery, A.; Ichter, B.; Wahid, A.; Tompson, J.; Vuong, Q.; Yu, T.;  et~al.
\newblock Palm-e: An embodied multimodal language model.
\newblock {\em arXiv} {\bf 2023}, arXiv:2303.03378.

\bibitem[Lu et~al.(2019)Lu, Batra, Parikh and Lee]{ViLBERT}
Lu, J.; Batra, D.; Parikh, D.; Lee, S.
\newblock Vilbert: Pretraining task-agnostic visiolinguistic representations for vision-and-language tasks. In \textit{Advances in Neural Information Processing Systems}; 
Curran Associates Inc.: Red Hook, NY, USA, 2019; Volume 32.

\bibitem[Chen et~al.(2022)Chen, Wu, Nießner and Chang]{D3Net}
Chen, D.; Wu, Q.; Nießner, M.; Chang, A.X.
\newblock D3 net: A unified speaker-listener architecture for 3d dense captioning and visual grounding.
\newblock In Proceedings of the European Conference on Computer Vision, 2022; pp. 487--505.

\bibitem[Jiang et~al.(2020)Jiang, Zhao, Shi, Liu, Fu and Jia]{Pointgroup}
Jiang, L.; Zhao, H.; Shi, S.; Liu, S.; Fu, C.W.; Jia, J.
\newblock Pointgroup: Dual-set point grouping for 3d instance segmentation.
\newblock In Proceedings of the IEEE/CVF Conference on Computer Vision and Pattern Recognition, 2020; pp. 4867--4876.

\bibitem[Fang et~al.(2024)Fang, Tan, Lin, Vasiljevic, Guizilini, Mei, Ambrus, Shakhnarovich and Walter]{Transcrib3D}
Fang, J.; Tan, X.; Lin, S.; Vasiljevic, I.; Guizilini, V.; Mei, H.; Ambrus, R.; Shakhnarovich, G.; Walter, M.R.
\newblock Transcrib3D: 3D Referring Expression Resolution through Large Language Models.
\newblock {\em arXiv} {\bf 2024}, arXiv:2404.19221.

\bibitem[Schult et~al.(2023)Schult, Engelmann, Hermans, Litany, Tang and Leibe]{Mask3D}
Schult, J.; Engelmann, F.; Hermans, A.; Litany, O.; Tang, S.; Leibe, B.
\newblock Mask3d: Mask transformer for 3d semantic instance segmentation.
\newblock In Proceedings of the 2023 IEEE International Conference on Robotics and Automation (ICRA), 2023; \mbox{pp. 8216--8223.}

\bibitem[Garg et~al.(2019)Garg, V, Dharmasiri, Hausler, Sünderhauf, Kumar, Drummond and Milford]{LookNoDeeper}
Garg, S.; Babu V, M.; Dharmasiri, T.; Hausler, S.; Sünderhauf, N.; Kumar, S.; Drummond, T.; Milford, M.
\newblock Look No Deeper: Recognizing Places from Opposing Viewpoints under Varying Scene Appearance using Single-View Depth Estimation. In Proceedings of the 2019 International Conference on Robotics and Automation (ICRA), Montreal, QC, Canada, 20--24 May 2019; pp. 4916--4923.
\newblock {\url{https://doi.org/10.1109/ICRA.2019.8794178}}.

\bibitem[Pavlova and Birbaumer(2001)]{PointLightMotion}
Pavlova, M.; Krägeloh-Mann, I.; Sokolov, A.; Birbaumer, N.
\newblock Recognition of point-light biological motion displays by young children.
\newblock \textit{Perception} \textbf{2001}, \textit{30}, 925--933.
\newblock {\url{https://doi.org/10.1068/p3157}}.

\bibitem[Gu et~al.(2024)Gu, Kuwajerwala, Morin, Jatavallabhula, Sen, Agarwal, Rivera, Paul, Ellis, Chellappa, et~al.]{ConceptGraphs}
Gu, Q.; Kuwajerwala, A.; Morin, S.; Jatavallabhula, K.M.; Sen, B.; Agarwal, A.; Rivera, C.; Paul, W.; Ellis, K.; Chellappa, R.;  et~al.
\newblock ConceptGraphs: Open-Vocabulary 3D Scene Graphs for Perception and Planning.
\newblock In Proceedings of the 2024 IEEE International Conference on Robotics and Automation (ICRA), 2024; pp. 5021--5028.

\bibitem[Liu et~al.(2023)Liu, Li, Wu and Lee]{LLaVA}
Liu, H.; Li, C.; Wu, Q.; Lee, Y.J.
\newblock Visual instruction tuning. In \textit{Advances in Neural Information Processing Systems}; Curran Associates Inc.: Red Hook, NY, USA, 2023; Volume 36, pp. 34892--34916.

\bibitem[He et~al.(2017)He, Gkioxari, Doll{\'a}r and Girshick]{MaskRCNN}
He, K.; Gkioxari, G.; Doll{\'a}r, P.; Girshick, R.
\newblock Mask r-cnn.
\newblock In Proceedings of the IEEE International Conference on Computer Vision, 2017; pp. 2961--2969.

\bibitem[Chang et~al.(2017)Chang, Dai, Funkhouser, Halber, Niessner, Savva, Song, Zeng and Zhang]{Matterport3D}
Chang, A.; Dai, A.; Funkhouser, T.; Halber, M.; Niessner, M.; Savva, M.; Song, S.; Zeng, A.; Zhang, Y.
\newblock Matterport3d: Learning from rgb-d data in indoor environments.
\newblock {\em arXiv} {\bf 2017}, arXiv:1709.06158.

\bibitem[He et~al.(2016)He, Zhang, Ren and Sun]{Resnet}
He, K.; Zhang, X.; Ren, S.; Sun, J.
\newblock Deep residual learning for image recognition.
\newblock In Proceedings of the IEEE Conference on Computer Vision and Pattern Recognition, 2016; pp. 770--778.

\bibitem[Carion et~al.(2020)Carion, Massa, Synnaeve, Usunier, Kirillov and Zagoruyko]{DETR}
Carion, N.; Massa, F.; Synnaeve, G.; Usunier, N.; Kirillov, A.; Zagoruyko, S.
\newblock End-to-end object detection with transformers.
\newblock In Proceedings of the European Conference on Computer Vision, 2020; pp. 213--229.

\bibitem[Abdelreheem et~al.(2023)Abdelreheem, Eldesokey, Ovsjanikov and Wonka]{ZeroShot3DShapeCorrespondence}
Abdelreheem, A.; Eldesokey, A.; Ovsjanikov, M.; Wonka, P.
\newblock Zero-shot 3d shape correspondence.
\newblock In Proceedings of the SIGGRAPH Asia 2023 Conference Papers, 2023; pp. 1--11.

\bibitem[Azuma et~al.(2022)Azuma, Miyanishi, Kurita and Kawanabe]{ScanQA}
Azuma, D.; Miyanishi, T.; Kurita, S.; Kawanabe, M.
\newblock Scanqa: 3d question answering for spatial scene understanding.
\newblock In Proceedings of the IEEE/CVF Conference on Computer Vision and Pattern Recognition, 2022; pp. 19129--19139.

\bibitem[Deng et~al.(2023)Deng, Gao, Ju and Yu]{SkeletonBasedActionRecognition}
Deng, Z.; Gao, Q.; Ju, Z.; Yu, X.
\newblock Skeleton-Based Multifeatures and Multistream Network for Real-Time Action Recognition.
\newblock {\em IEEE Sens. J.} {\bf 2023}, {\em 23}, 7397--7409.
\newblock {\url{https://doi.org/10.1109/JSEN.2023.3246133}}.

\bibitem[Yuan et~al.(2024)Yuan, Ren, Feng, Zhao, Cui and Li]{VisualProgramming}
Yuan, Z.; Ren, J.; Feng, C.M.; Zhao, H.; Cui, S.; Li, Z.
\newblock Visual Programming for Zero-shot Open-Vocabulary 3D Visual Grounding.
\newblock {\em arXiv} {\bf 2024}, arXiv:2311.15383.

\bibitem[Chen et~al.(2020)Chen, Chang and Nießner]{ScanRefer}
Chen, D.; Chang, A.X.; Nießner, M.
\newblock ScanRefer: 3D object localization in RGB-D scans using natural language.
\newblock In Proceedings of the European Conference on Computer Vision, 2020; pp. 202--221.

\bibitem[Jin et~al.(2023)Jin, Hayat, Yang, Guo and Lei]{ContextAwareAlignmentAndMutualMaskingFor3DLanguagePreTraining}
Jin, Z.; Hayat, M.; Yang, Y.; Guo, Y.; Lei, Y.
\newblock Context-aware alignment and mutual masking for 3d-language pre-training.
\newblock In Proceedings of the IEEE/CVF Conference on Computer Vision and Pattern Recognition,  2023; pp. 10984--10994.

\bibitem[Wu et~al.(2015)Wu, Song, Khosla, Yu, Zhang, Tang and Xiao]{3DShapeNets}
Wu, Z.; Song, S.; Khosla, A.; Yu, F.; Zhang, L.; Tang, X.; Xiao, J.
\newblock 3d shapenets: A deep representation for volumetric shapes.
\newblock In Proceedings of the IEEE Conference on Computer Vision and Pattern Recognition, 2015; pp. 1912--1920.

\bibitem[Uy et~al.(2019)Uy, Pham, Hua, Nguyen and Yeung]{ScanObjectNN}
Uy, M.A.; Pham, Q.H.; Hua, B.S.; Nguyen, T.; Yeung, S.K.
\newblock Revisiting point cloud classification: A new benchmark dataset and classification model on real-world data.
\newblock In Proceedings of the IEEE/CVF International Conference on Computer Vision, 2019; \mbox{pp. 1588--1597.}

\bibitem[Poole et~al.(2022)Poole, Jain, Barron and Mildenhall]{DreamFusion}
Poole, B.; Jain, A.; Barron, J.T.; Mildenhall, B.
\newblock Dreamfusion: Text-to-3d using 2d diffusion.
\newblock {\em arXiv} {\bf 2022}, arXiv:2209.14988.

\bibitem[Mildenhall et~al.(2021)Mildenhall, Srinivasan, Tancik, Barron, Ramamoorthi and Ng]{NeRF}
Mildenhall, B.; Srinivasan, P.P.; Tancik, M.; Barron, J.T.; Ramamoorthi, R.; Ng, R.
\newblock Nerf: Representing scenes as neural radiance fields for view synthesis.
\newblock {\em Commun. ACM} {\bf 2021}, {\em 65}, 99--106.

\bibitem[Zhou et~al.(2024)Zhou, Ran, Xiong, He, Lin, Wang, Sun and Yang]{GALA3D}
Zhou, X.; Ran, X.; Xiong, Y.; He, J.; Lin, Z.; Wang, Y.; Sun, D.; Yang, M.H.
\newblock Gala3d: Towards text-to-3d complex scene generation via layout-guided generative gaussian splatting.
\newblock {\em arXiv} {\bf 2024}, arXiv:2402.07207.

\bibitem[Qi et~al.(2024)Qi, Fang, Sun, Wu, Wu, Wang, Lin and Zhao]{GPT4Point}
Qi, Z.; Fang, Y.; Sun, Z.; Wu, X.; Wu, T.; Wang, J.; Lin, D.; Zhao, H.
\newblock Gpt4point: A unified framework for point-language understanding and generation.
\newblock In Proceedings of the IEEE/CVF Conference on Computer Vision and Pattern Recognition, 2024; pp. 26417--26427.

\bibitem[Zhang et~al.(2024)Zhang, Zhang, Xu and Tao]{QFormer}
Zhang, Q.; Zhang, J.; Xu, Y.; Tao, D.
\newblock Vision transformer with quadrangle attention.
\newblock {\em IEEE Trans. Pattern Anal. Mach. Intell.} {\bf 2024}, {\em 46}, 3608--3624.

\bibitem[Wang et~al.(2024)Wang, Zhong, Chai, et~al.]{Chat2Layout}
Wang, C.; Zhong, H.; Chai, M.; He, M.; Chen, D.; Liao, J.
\newblock Chat2Layout: Interactive 3D Furniture Layout with a Multimodal LLM.
\newblock {\em arXiv} {\bf 2024}, arXiv:2407.21333.

\bibitem[Liu et~al.(2024)Liu, Huang, Hou, Wang, Yin, Gong, Gao and Ouyang]{Uni3DLLM}
Liu, D.; Huang, X.; Hou, Y.; Wang, Z.; Yin, Z.; Gong, Y.; Gao, P.; Ouyang, W.
\newblock Uni3D-LLM: Unifying Point Cloud Perception, Generation and Editing with Large Language Models.
\newblock {\em arXiv} {\bf 2024}, arXiv:2402.03327.

\bibitem[Wang et~al.(2023)Wang, Bao, Dong, Bjorck, Peng, Liu, Aggarwal, Mohammed, Singhal, Som, et~al.]{BEiT}
Wang, W.; Bao, H.; Dong, L.; Bjorck, J.; Peng, Z.; Liu, Q.; Aggarwal, K.; Mohammed, O.K.; Singhal, S.; Som, S.;  et~al.
\newblock Image as a foreign language: Beit pretraining for vision and vision-language tasks.
\newblock In Proceedings of the IEEE/CVF Conference on Computer Vision and Pattern Recognition, 2023; pp. 19175--19186.

\bibitem[Jain et~al.(2024)Jain, Yang and Shi]{Vcoder}
Jain, J.; Yang, J.; Shi, H.
\newblock Vcoder: Versatile vision encoders for multimodal large language models.
\newblock In Proceedings of the IEEE/CVF Conference on Computer Vision and Pattern Recognition,  2024; pp. 27992--28002.

\bibitem[Chen et~al.(2023)Chen, Han, Zhao, Zhang, Shi, Xu and Xu]{CFormer}
Chen, F.; Han, M.; Zhao, H.; Zhang, Q.; Shi, J.; Xu, S.; Xu, B.
\newblock X-llm: Bootstrapping advanced large language models by treating multi-modalities as foreign languages.
\newblock {\em arXiv} {\bf 2023}, arXiv:2305.04160.

\bibitem[Hsu et~al.(2021)Hsu, Bolte, Tsai, Lakhotia, Salakhutdinov and Mohamed]{HuBERT}
Hsu, W.N.; Bolte, B.; Tsai, Y.H.H.; Lakhotia, K.; Salakhutdinov, R.; Mohamed, A.
\newblock Hubert: Self-supervised speech representation learning by masked prediction of hidden units.
\newblock {\em IEEE/ACM Trans. Audio Speech Lang. Process.} {\bf 2021}, {\em 29}, 3451--3460.

\bibitem[Radford et~al.(2023)Radford, Kim, Xu, Brockman, McLeavey and Sutskever]{Whisper}
Radford, A.; Kim, J.W.; Xu, T.; Brockman, G.; McLeavey, C.; Sutskever, I.
\newblock Robust speech recognition via large-scale weak supervision.
\newblock In Proceedings of the International Conference on Machine Learning, 2023; pp. 28492--28518.

\bibitem[Wu et~al.(2023)Wu, Chen, Zhang, Hui, Berg-Kirkpatrick and Dubnov]{CLAP}
Wu, Y.; Chen, K.; Zhang, T.; Hui, Y.; Berg-Kirkpatrick, T.; Dubnov, S.
\newblock Large-scale contrastive language-audio pretraining with feature fusion and keyword-to-caption augmentation.
\newblock In Proceedings of the ICASSP 2023, 2023 IEEE International Conference on Acoustics, Speech and Signal Processing (ICASSP), 2023; pp. 1--5.

\bibitem[Yu et~al.(2022)Yu, Tang, Rao, Huang, Zhou and Lu]{PointBERT}
Yu, X.; Tang, L.; Rao, Y.; Huang, T.; Zhou, J.; Lu, J.
\newblock Point-bert: Pre-training 3d point cloud transformers with masked point modeling.
\newblock In Proceedings of the IEEE/CVF Conference on Computer Vision and Pattern Recognition, 2022; pp. 19313--19322.

\bibitem[Caron et~al.(2021)Caron, Touvron, Misra, J{\'e}gou, Mairal, Bojanowski and Joulin]{DINO}
Caron, M.; Touvron, H.; Misra, I.; J{\'e}gou, H.; Mairal, J.; Bojanowski, P.; Joulin, A.
\newblock Emerging properties in self-supervised vision transformers.
\newblock In Proceedings of the IEEE/CVF International Conference on Computer Vision,  2021; pp. 9650--9660.

\bibitem[Guzhov et~al.(2022)Guzhov, Raue, Hees and Dengel]{AudioCLIP}
Guzhov, A.; Raue, F.; Hees, J.; Dengel, A.
\newblock Audioclip: Extending clip to image, text and audio.
\newblock In Proceedings of the ICASSP 2022, 2022 IEEE International Conference on Acoustics, Speech and Signal Processing (ICASSP), 2022; pp. 976--980.

\bibitem[Hong et~al.(2023)Hong, Zhen, Chen, Zheng, Du, Chen and Gan]{3DLLM}
Hong, Y.; Zhen, H.; Chen, P.; Zheng, S.; Du, Y.; Chen, Z.; Gan, C.
\newblock 3d-llm: Injecting the 3d world into large language models. In \textit{Advances in Neural Information Processing Systems}; Curran Associates Inc.: Red Hook, NY, USA, 2023; Volume 36, pp. 20482--20494.

\bibitem[Chen et~al.(2022)Chen, Guhur, Tapaswi, Schmid and Laptev]{LearningFromUnlabeled3DEnvironments}
Chen, S.; Guhur, P.L.; Tapaswi, M.; Schmid, C.; Laptev, I.
\newblock Learning from unlabeled 3d environments for vision-and-language navigation.
\newblock In Proceedings of the European Conference on Computer Vision, 2022; pp. 638--655.

\bibitem[Seita et~al.(2023)Seita, Wang, Shetty, Li, Erickson and Held]{ToolFlowNet}
Seita, D.; Wang, Y.; Shetty, S.J.; Li, E.Y.; Erickson, Z.; Held, D.
\newblock Toolflownet: Robotic manipulation with tools via predicting tool flow from point clouds.
\newblock In Proceedings of the Conference on Robot Learning, 2023; pp. 1038--1049.

\bibitem[Qin et~al.(2023)Qin, Huang, Yin, Su and Wang]{DexPoint}
Qin, Y.; Huang, B.; Yin, Z.H.; Su, H.; Wang, X.
\newblock Dexpoint: Generalizable point cloud reinforcement learning for sim-to-real dexterous manipulation.
\newblock In Proceedings of the Conference on Robot Learning, 2023; pp. 594--605.

\bibitem[Reed et~al.(2022)Reed, Zolna, Parisotto, Colmenarejo, Novikov, Barth-Maron, Gimenez, Sulsky, Kay, Springenberg, et~al.]{GATO}
Reed, S.; Zolna, K.; Parisotto, E.; Colmenarejo, S.G.; Novikov, A.; Barth-Maron, G.; Gimenez, M.; Sulsky, Y.; Kay, J.; Springenberg, J.T.; et~al.
\newblock A generalist agent.
\newblock {\em arXiv} {\bf 2022}, arXiv:2205.06175.

\bibitem[Guhur et~al.(2023)Guhur, Chen, Pinel, Tapaswi, Laptev and Schmid]{HiveFormer}
Guhur, P.L.; Chen, S.; Pinel, R.G.; Tapaswi, M.; Laptev, I.; Schmid, C.
\newblock Instruction-driven history-aware policies for robotic manipulations.
\newblock In Proceedings of the Conference on Robot Learning,  2023; pp. 175--187.

\bibitem[Jang et~al.(2022)Jang, Irpan, Khansari, Kappler, Ebert, Lynch, Levine and Finn]{BCZ}
Jang, E.; Irpan, A.; Khansari, M.; Kappler, D.; Ebert, F.; Lynch, C.; Levine, S.; Finn, C.
\newblock Bc-z: Zero-shot task generalization with robotic imitation learning.
\newblock In Proceedings of the Conference on Robot Learning,  2022; pp. 991--1002.

\bibitem[Liu et~al.(2022)Liu, James, Davison and Johns]{AutoLambda}
Liu, S.; James, S.; Davison, A.J.; Johns, E.
\newblock Auto-lambda: Disentangling dynamic task relationships.
\newblock {\em arXiv} {\bf 2022}, arXiv:2202.03091.

\bibitem[Hong et~al.(2024)Hong, Zheng, Chen, Wang, Li and Gan]{MultiPLY}
Hong, Y.; Zheng, Z.; Chen, P.; Wang, Y.; Li, J.; Gan, C.
\newblock Multiply: A multisensory object-centric embodied large language model in 3d world.
\newblock In Proceedings of the IEEE/CVF Conference on Computer Vision and Pattern Recognition,  2024; pp. 26406--26416.

\bibitem[Huang et~al.(2023)Huang, Dong, Yang, Huang, Lau, Ouyang and Zuo]{CLIP2Point}
Huang, T.; Dong, B.; Yang, Y.; Huang, X.; Lau, R.W.; Ouyang, W.; Zuo, W.
\newblock Clip2point: Transfer clip to point cloud classification with image-depth pre-training.
\newblock In Proceedings of the IEEE/CVF International Conference on Computer Vision, 2023; \mbox{pp. 22157--22167.}

\bibitem[Guo et~al.(2023)Guo, Zhang, Zhu, Tang, Ma, Han, Chen, Gao, Li, Li, et~al.]{PointBindPointLLM}
Guo, Z.; Zhang, R.; Zhu, X.; Tang, Y.; Ma, X.; Han, J.; Chen, K.; Gao, P.; Li, X.; Li, H.;  et~al.
\newblock Point-bind \& point-llm: Aligning point cloud with multi-modality for 3d understanding, generation and instruction following.
\newblock {\em arXiv} {\bf 2023}, arXiv:2309.00615.

\bibitem[Ji et~al.(2024)Ji, Wang, Wu, Ma, Sun and Ji]{JM3D}
Ji, J.; Wang, H.; Wu, C.; Ma, Y.; Sun, X.; Ji, R.
\newblock JM3D \& JM3D-LLM: Elevating 3D Representation with Joint Multi-modal Cues.
\newblock {\em IEEE Trans. Pattern Anal. Mach. Intell.}
\textbf{2025}, \textit{47}, 2475--2492.

\bibitem[Karnchanachari et~al.(2024)Karnchanachari, Geromichalos, Tan, Li, Eriksen, Yaghoubi, Mehdipour, Bernasconi, Fong, Guo, et~al.]{nuPlan}
Karnchanachari, N.; Geromichalos, D.; Tan, K.S.; Li, N.; Eriksen, C.; Yaghoubi, S.; Mehdipour, N.; Bernasconi, G.; Fong, W.K.; Guo, Y.;  et~al.
\newblock Towards Learning-Based Planning: The nuPlan Benchmark for Real-World Autonomous Driving.
\newblock {\em arXiv} {\bf 2024}, arXiv:2403.04133.

\bibitem[Dai et~al.(2017)Dai, Chang, Savva, Halber, Funkhouser and Nie{\ss}ner]{ScanNet}
Dai, A.; Chang, A.X.; Savva, M.; Halber, M.; Funkhouser, T.; Nie{\ss}ner, M.
\newblock Scannet: Richly-annotated 3d reconstructions of indoor scenes.
\newblock In Proceedings of the IEEE Conference on Computer Vision and Pattern Recognition, 2017; pp. 5828--5839.

\bibitem[Achlioptas et~al.(2020)Achlioptas, Abdelreheem, Xia, Elhoseiny and Guibas]{ReferIt3D}
Achlioptas, P.; Abdelreheem, A.; Xia, F.; Elhoseiny, M.; Guibas, L.
\newblock ReferIt3D: Neural Listeners for Fine-Grained 3D Object Identification in Real-World Scenes.
\newblock {In Proceedings of the 16th European Conference on Computer Vision (ECCV)}, 2020.

\bibitem[Wald et~al.(2019)Wald, Avetisyan, Navab, Tombari and Nie{\ss}ner]{3RScan}
Wald, J.; Avetisyan, A.; Navab, N.; Tombari, F.; Nie{\ss}ner, M.
\newblock Rio: 3d object instance re-localization in changing indoor environments.
\newblock In Proceedings of the IEEE/CVF International Conference on Computer Vision, 2019; pp. 7658--7667.

\bibitem[Zhu et~al.(2023)Zhu, Ma, Chen, Deng, Huang and Li]{ScanScribe}
Zhu, Z.; Ma, X.; Chen, Y.; Deng, Z.; Huang, S.; Li, Q.
\newblock 3d-vista: Pre-trained transformer for 3d vision and text alignment.
\newblock In Proceedings of the IEEE/CVF International Conference on Computer Vision, 2023; pp. 2911--2921.

\bibitem[Liao et~al.(2022)Liao, Xie and Geiger]{KITTI360Pose}
Liao, Y.; Xie, J.; Geiger, A.
\newblock Kitti-360: A novel dataset and benchmarks for urban scene understanding in 2d and 3d.
\newblock {\em IEEE Trans. Pattern Anal. Mach. Intell.} {\bf 2022}, {\em 45}, 3292--3310.

\bibitem[Yeshwanth et~al.(2023)Yeshwanth, Liu, Nie{\ss}ner and Dai]{ScanNet++}
Yeshwanth, C.; Liu, Y.C.; Nie{\ss}ner, M.; Dai, A.
\newblock Scannet++: A high-fidelity dataset of 3d indoor scenes.
\newblock In Proceedings of the IEEE/CVF International Conference on Computer Vision, 2023; pp. 12--22.

\bibitem[Baruch et~al.(2021)Baruch, Chen, Dehghan, Dimry, Feigin, Fu, Gebauer, Joffe, Kurz, Schwartz, et~al.]{ARKitScenes}
Baruch, G.; Chen, Z.; Dehghan, A.; Dimry, T.; Feigin, Y.; Fu, P.; Gebauer, T.; Joffe, B.; Kurz, D.; Schwartz, A.;  et~al.
\newblock Arkitscenes: A diverse real-world dataset for 3d indoor scene
  understanding using mobile rgb-d data.
\newblock {\em arXiv} {\bf 2021}, arXiv:2111.08897.

\bibitem[Ramakrishnan et~al.(2021)Ramakrishnan, Gokaslan, Wijmans, Maksymets, Clegg, Turner, Undersander, Galuba, Westbury, Chang, et~al.]{HM3D}
Ramakrishnan, S.K.; Gokaslan, A.; Wijmans, E.; Maksymets, O.; Clegg, A.; Turner, J.; Undersander, E.; Galuba, W.; Westbury, A.; Chang, A.X.;  et~al.
\newblock Habitat-matterport 3d dataset (hm3d): 1000 large-scale 3d environments for embodied ai.
\newblock {\em arXiv} {\bf 2021}, arXiv:2109.08238.

\bibitem[Mao et~al.(2022)Mao, Zhang, Jiang, Chang and Savva]{MultiScan}
Mao, Y.; Zhang, Y.; Jiang, H.; Chang, A.; Savva, M.
\newblock MultiScan: Scalable RGBD scanning for 3D environments with articulated objects. In \textit{Advances in Neural Information Processing Systems}; Curran Associates Inc.: Red Hook, NY, USA, 2022; Volume 35, \mbox{pp. 9058--9071.}

\bibitem[Zheng et~al.(2020)Zheng, Zhang, Li, Tang, Gao and Zhou]{Structured3D}
Zheng, J.; Zhang, J.; Li, J.; Tang, R.; Gao, S.; Zhou, Z.
\newblock Structured3d: A large photo-realistic dataset for structured 3d modeling.
\newblock In \textit{Computer Vision--ECCV 2020: 16th European Conference, Glasgow, UK, August 23--28, 2020, Proceedings, Part IX}; Springer: Cham, Switzerland,  2020; pp. 519--535.

\bibitem[Deitke et~al.(2022)Deitke, VanderBilt, Herrasti, Weihs, Ehsani, Salvador, Han, Kolve, Kembhavi and Mottaghi]{ProcTHOR}
Deitke, M.; VanderBilt, E.; Herrasti, A.; Weihs, L.; Ehsani, K.; Salvador, J.; Han, W.; Kolve, E.; Kembhavi, A.; Mottaghi, R.
\newblock ProcTHOR: Large-Scale Embodied AI Using Procedural Generation.
\newblock In \textit{Proceedings of the Advances in Neural Information Processing Systems}; Koyejo, S., Mohamed, S., Agarwal, A., Belgrave, D., Cho, K., Oh, A., Eds.; Curran Associates, Inc.: Red Hook, NY, USA, 2022; Volume 35, pp. 5982--5994.

\bibitem[Hwang et~al.(2015)Hwang, Park, Kim, Choi and Kweon]{Kaist}
Hwang, S.; Park, J.; Kim, N.; Choi, Y.; Kweon, I.S.
\newblock Multispectral Pedestrian Detection: Benchmark Dataset and Baselines.
\newblock In Proceedings of the IEEE Conference on Computer Vision and Pattern Recognition (CVPR),  2015.

\bibitem[Ma et~al.(2022)Ma, Yong, Zheng, Li, Liang, Zhu and Huang]{SQA3D}
Ma, X.; Yong, S.; Zheng, Z.; Li, Q.; Liang, Y.; Zhu, S.C.; Huang, S.
\newblock Sqa3d: Situated question answering in 3d scenes.
\newblock {\em arXiv} {\bf 2022}, arXiv:2210.07474.

\bibitem[Song et~al.(2017)Song, Yu, Zeng, Chang, Savva and Funkhouser]{SUNCG}
Song, S.; Yu, F.; Zeng, A.; Chang, A.X.; Savva, M.; Funkhouser, T.
\newblock Semantic scene completion from a single depth image.
\newblock In Proceedings of the IEEE Conference on Computer Vision and Pattern Recognition,  2017; pp. 1746--1754.

\bibitem[Qi et~al.(2020)Qi, Wu anderson, Wang, Wang, Shen and Hengel]{REVERIE}
Qi, Y.; Wu, Q.; Anderson, P.; Wang, X.; Wang, W.Y.; Shen, C.; Hengel, A.v.d.
\newblock Reverie: Remote embodied visual referring expression in real indoor environments.
\newblock In Proceedings of the IEEE/CVF Conference on Computer Vision and Pattern Recognition, 2020; \mbox{pp 9982--9991.}

\bibitem[Zhu et~al.(2021)Zhu, Liang, Zhu, Yu, Chang and Liang]{SOON}
Zhu, F.; Liang, X.; Zhu, Y.; Yu, Q.; Chang, X.; Liang, X.
\newblock Soon: Scenario oriented object navigation with graph-based exploration.
\newblock In Proceedings of the IEEE/CVF Conference on Computer Vision and Pattern Recognition, 2021; pp. 12689--12699.

\bibitem[Liu et~al.(2023)Liu, Liu, Wu, Ma, Liu, Zhong, Luo and Fan]{FMB}
Liu, J.; Liu, Z.; Wu, G.; Ma, L.; Liu, R.; Zhong, W.; Luo, Z.; Fan, X.
\newblock Multi-interactive feature learning and a full-time multi-modality benchmark for image fusion and segmentation.
\newblock In Proceedings of the IEEE/CVF International Conference on Computer Vision, 2023; pp. 8115--8124.

\bibitem[Liu et~al.(2019)Liu, Shahroudy, Perez, Wang, Duan and Kot]{liu2019ntu}
Liu, J.; Shahroudy, A.; Perez, M.; Wang, G.; Duan, L.Y.; Kot, A.C.
\newblock Ntu rgb+ d 120: A large-scale benchmark for 3d human activity understanding.
\newblock {\em IEEE Trans. Pattern Anal. Mach. Intell.} {\bf 2019}, {\em 42}, 2684--2701.

\bibitem[Cong et~al.(2022)Cong, Zhu, Qiao, Ren, Peng, Hou, Xu, Yang, Manocha and Ma]{STCrowd}
Cong, P.; Zhu, X.; Qiao, F.; Ren, Y.; Peng, X.; Hou, Y.; Xu, L.; Yang, R.; Manocha, D.; Ma, Y.
\newblock Stcrowd: A multimodal dataset for pedestrian perception in crowded scenes.
\newblock In Proceedings of the IEEE/CVF Conference on Computer Vision and Pattern Recognition,  2022; pp. 19608--19617.

\bibitem[Flir()]{FlirADAS}
Flir.
\newblock FLIR Thermal Datasets for Algorithm Training.
\newblock  Available online: \url{https://www.flir.in/oem/adas/dataset/} (accessed on 27 April 2025).


\bibitem[Wang et~al.(2024)Wang, Mao, Zhu, Xu, Lyu, Li, Chen, Zhang, Chen, Xue, et~al.]{EmbodiedScan}
Wang, T.; Mao, X.; Zhu, C.; Xu, R.; Lyu, R.; Li, P.; Chen, X.; Zhang, W.; Chen, K.; Xue, T.;  et~al.
\newblock Embodiedscan: A holistic multi-modal 3d perception suite towards embodied ai.
\newblock In Proceedings of the IEEE/CVF Conference on Computer Vision and Pattern Recognition,  2024; pp. 19757--19767.

\bibitem[Li et~al.(2023)Li, Chen, Zhang, Chen, Zhu, Yin, Yu and Chen]{M3DBench}
Li, M.; Chen, X.; Zhang, C.; Chen, S.; Zhu, H.; Yin, F.; Yu, G.; Chen, T.
\newblock M3dbench: Let's instruct large models with multi-modal 3d prompts.
\newblock {\em arXiv} {\bf 2023}, arXiv:2312.10763.

\bibitem[Yuan et~al.(2025)Yuan, Yan, Li, Li, Guo, Cui and Li]{PhraseRefer}
Yuan, Z.; Yan, X.; Li, Z.; Li, X.; Guo, Y.; Cui, S.; Li, Z.
\newblock Toward Fine-Grained 3-D Visual Grounding Through Referring Textual Phrases.
\newblock {\em IEEE Trans. Neural Netw. Learn. Syst.} {\bf 2025}, \textit{36},  19411--19422.
\newblock {\url{https://doi.org/10.1109/TNNLS.2025.3571959}}.

\bibitem[Yan et~al.(2023)Yan, Yuan, Du, Liao, Guo, Cui and Li]{CLEVR3D}
Yan, X.; Yuan, Z.; Du, Y.; Liao, Y.; Guo, Y.; Cui, S.; Li, Z.
\newblock Comprehensive visual question answering on point clouds through compositional scene manipulation.
\newblock {\em IEEE Trans. Vis. Comput. Graph.} {\bf 2023}, {\em 30}, 7473--7485.

\bibitem[Kuznetsova et~al.(2020)Kuznetsova, Rom, Alldrin, Uijlings, Krasin, Pont-Tuset, Kamali, Popov, Malloci, Kolesnikov, et~al.]{OpenImages}
Kuznetsova, A.; Rom, H.; Alldrin, N.; Uijlings, J.; Krasin, I.; Pont-Tuset, J.; Kamali, S.; Popov, S.; Malloci, M.; Kolesnikov, A.;  et~al.
\newblock The open images dataset v4: Unified image classification, object detection and visual relationship detection at scale.
\newblock {\em Int. J. Comput. Vis.} {\bf 2020}, {\em 128}, 1956--1981.

\bibitem[Russakovsky et~al.(2015)Russakovsky, Deng, Su, Krause, Satheesh, Ma, Huang, Karpathy, Khosla, Bernstein, et~al.]{ImageNet}
Russakovsky, O.; Deng, J.; Su, H.; Krause, J.; Satheesh, S.; Ma, S.; Huang, Z.; Karpathy, A.; Khosla, A.; Bernstein, M.;  et~al.
\newblock Imagenet large scale visual recognition challenge.
\newblock {\em Int. J. Comput. Vis.} {\bf 2015}, {\em 115}, 211--252.

\bibitem[Changpinyo et~al.(2021)Changpinyo, Sharma, Ding and Soricut]{CC12M}
Changpinyo, S.; Sharma, P.; Ding, N.; Soricut, R.
\newblock Conceptual 12m: Pushing web-scale image-text pre-training to recognize long-tail visual concepts.
\newblock In Proceedings of the IEEE/CVF Conference on Computer Vision and Pattern Recognition,  2021; \mbox{pp. 3558--3568.}

\bibitem[Schuhmann et~al.(2021)Schuhmann, Vencu, Beaumont, Kaczmarczyk, Mullis, Katta, Coombes, Jitsev and Komatsuzaki]{LAION400M}
Schuhmann, C.; Vencu, R.; Beaumont, R.; Kaczmarczyk, R.; Mullis, C.; Katta, A.; Coombes, T.; Jitsev, J.; Komatsuzaki, A.
\newblock Laion-400m: Open dataset of clip-filtered 400 million image-text pairs.
\newblock {\em arXiv} {\bf 2021}, arXiv:2111.02114.

\bibitem[Schuhmann et~al.(2022)Schuhmann, Beaumont, Vencu, Gordon, Wightman, Cherti, Coombes, Katta, Mullis, Wortsman, et~al.]{LAION5B}
Schuhmann, C.; Beaumont, R.; Vencu, R.; Gordon, C.; Wightman, R.; Cherti, M.; Coombes, T.; Katta, A.; Mullis, C.; Wortsman, M.;  et~al.
\newblock Laion-5b: An open large-scale dataset for training next generation image-text models. In \textit{Advances in Neura Information Processing Systems}; Curran Associates Inc.: Red Hook, NY, USA, 2022; Volume 35, pp. 25278--25294.

\bibitem[Desai et~al.(2021)Desai, Kaul, Aysola and Johnson]{RedCaps}
Desai, K.; Kaul, G.; Aysola, Z.; Johnson, J.
\newblock Redcaps: Web-curated image-text data created by the people, for the people.
\newblock {\em arXiv} {\bf 2021}, arXiv:2111.11431.

\bibitem[Chen et~al.(2023)Chen, Luo, Wang, Baktashmotlagh and Huang]{RevisitingDomainAdaptive3DObjectDetection}
Chen, Z.; Luo, Y.; Wang, Z.; Baktashmotlagh, M.; Huang, Z.
\newblock Revisiting domain-adaptive 3D object detection by reliable, diverse and class-balanced pseudo-labeling.
\newblock In Proceedings of the IEEE/CVF International Conference on Computer Vision,  2023; \mbox{pp. 3714--3726.}

\bibitem[Saito et~al.(2018)Saito, Watanabe, Ushiku and Harada]{MaximumClassifierDiscrepancy}
Saito, K.; Watanabe, K.; Ushiku, Y.; Harada, T.
\newblock Maximum classifier discrepancy for unsupervised domain adaptation.
\newblock In Proceedings of the IEEE Conference on Computer Vision and Pattern Recognition,  2018; pp. 3723--3732.

\bibitem[Zhang et~al.(0)Zhang, Wang, Zhang, Jiang, Luo, Wang, Zhang, Liu, Shen, Ye and Jiang]{FogFusion}
Zhang, B.; Wang, Y.; Zhang, C.; Jiang, J.; Luo, X.; Wang, X.; Zhang, Y.; Liu, Z.; Shen, G.; Ye, Y.;  et~al.
\newblock FogFusion: Robust 3D object detection based on camera-LiDAR fusion for autonomous driving in foggy weather conditions.
\newblock {\em Proc. Inst. Mech. Eng. Part D J. Automob. Eng.} \textbf{2025}, 09544070251327229.
\newblock {\url{https://doi.org/10.1177/09544070251327229}}.

\bibitem[Maanp{\"a}{\"a} et~al.(2021)Maanp{\"a}{\"a}, Taher, Manninen, Pakola, Melekhov and Hyypp{\"a}]{MultimodalEndToEndLearning}
Maanp{\"a}{\"a}, J.; Taher, J.; Manninen, P.; Pakola, L.; Melekhov, I.; Hyypp{\"a}, J.
\newblock Multimodal end-to-end learning for autonomous steering in adverse road and weather conditions.
\newblock In Proceedings of the 2020 25th International Conference on Pattern Recognition (ICPR),  2021; pp. 699--706.

\bibitem[Banerjee and Lavie(2005)]{METEOR}
Banerjee, S.; Lavie, A.
\newblock METEOR: An automatic metric for MT evaluation with improved correlation with human judgments.
\newblock In Proceedings of the ACL Workshop on Intrinsic and Extrinsic Evaluation Measures for Machine Translation and/or Summarization,  2005; pp. 65--72.

\bibitem[Lin(2004)]{ROUGE}
Lin, C.Y.
\newblock Rouge: A package for automatic evaluation of summaries.
\newblock In Proceedings of the Text Summarization Branches out,  2004; pp. 74--81.

\bibitem[Vedantam et~al.(2015)Vedantam, Lawrence Zitnick and Parikh]{CIDEr}
Vedantam, R.; Lawrence Zitnick, C.; Parikh, D.
\newblock Cider: Consensus-based image description evaluation.
\newblock In Proceedings of the IEEE Conference on Computer Vision and Pattern Recognition,  2015; pp. 4566--4575.

\bibitem[Papineni et~al.(2002)Papineni, Roukos, Ward and Zhu]{Bleu}
Papineni, K.; Roukos, S.; Ward, T.; Zhu, W.J.
\newblock Bleu: A method for automatic evaluation of machine translation.
\newblock In Proceedings of the 40th Annual Meeting of the Association for Computational Linguistics,  2002; pp. 311--318.

\bibitem[Wijmans et~al.(2019)Wijmans, Datta, Maksymets, Das, Gkioxari, Lee, Essa, Parikh and Batra]{EQAInPhotorealisticEnvironments}
Wijmans, E.; Datta, S.; Maksymets, O.; Das, A.; Gkioxari, G.; Lee, S.; Essa, I.; Parikh, D.; Batra, D.
\newblock Embodied question answering in photorealistic environments with point cloud perception.
\newblock In Proceedings of the IEEE/CVF Conference on Computer Vision and Pattern Recognition,  2019; pp. 6659--6668.

\bibitem[Rubner et~al.(2000)Rubner, Tomasi and Guibas]{EMD}
Rubner, Y.; Tomasi, C.; Guibas, L.J.
\newblock The earth mover's distance as a metric for image retrieval.
\newblock {\em Int. J. Comput. Vis.} {\bf 2000}, {\em 40}, 99--121.

\bibitem[Achlioptas et~al.(2018)Achlioptas, Diamanti, Mitliagkas and Guibas]{LearningRepresentationsAndGenerativeModels}
Achlioptas, P.; Diamanti, O.; Mitliagkas, I.; Guibas, L.
\newblock Learning representations and generative models for 3d point clouds.
\newblock In Proceedings of the International Conference on Machine Learning,  2018; pp. 40--49.

\bibitem[Yokoyama et~al.(2021)Yokoyama, Ha and Batra]{SCT}
Yokoyama, N.; Ha, S.; Batra, D.
\newblock Success weighted by completion time: A dynamics-aware evaluation criteria for embodied navigation.
\newblock In Proceedings of the 2021 IEEE/RSJ International Conference on Intelligent Robots and Systems (IROS),  2021; \mbox{pp. 1562--1569.}

\bibitem[Singh and Namin(2025)]{SurveyOnChatbots}
Singh, S.U.; Namin, A.S.
\newblock A survey on chatbots and large language models: Testing and evaluation techniques.
\newblock {\em Nat. Lang. Process. J.} {\bf 2025}, {\em 10}, 100128.
\newblock {\url{https://doi.org/10.1016/j.nlp.2025.100128}}.

\bibitem[Deriu et~al.(2021)Deriu, Rodrigo, Otegi, Echegoyen, Rosset, Agirre and Cieliebak]{SurveyEvaluationMethodsForDialogueSystems}
Deriu, J.; Rodrigo, A.; Otegi, A.; Echegoyen, G.; Rosset, S.; Agirre, E.; Cieliebak, M.
\newblock Survey on evaluation methods for dialogue systems.
\newblock {\em Artif. Intell. Rev.} {\bf 2021}, {\em 54}, 755--810.

\bibitem[Norton et~al.(2022)Norton, Admoni, Crandall, Fitzgerald, Gautam, Goodrich, Saretsky, Scheutz, Simmons, Steinfeld and Yanco]{MetricsForRobotProficiency}
Norton, A.; Admoni, H.; Crandall, J.; Fitzgerald, T.; Gautam, A.; Goodrich, M.; Saretsky, A.; Scheutz, M.; Simmons, R.; Steinfeld, A.;  et~al.
\newblock Metrics for Robot Proficiency Self-Assessment and Communication of
  Proficiency in Human-Robot Teams.
\newblock {\em ACM Trans. Hum.-Robot Interact.} {\bf 2022}, {\em
  11}, 29.
\newblock {\url{https://doi.org/10.1145/3522579}}.

\bibitem[Bai et~al.(2025)Bai, Ding and Taylor]{VirtualAgentsToRobotTeams}
Bai, Y.; Ding, Z.; Taylor, A.
\newblock From Virtual Agents to Robot Teams: A Multi-Robot Framework Evaluation in High-Stakes Healthcare Context.
\newblock {\em arXiv} {\bf 2025}, arXiv:2506.03546.

\bibitem[Chang et~al.(2015)Chang, Funkhouser, Guibas, Hanrahan, Huang, Li, Savarese, Savva, Song, Su, et~al.]{ShapeNet}
Chang, A.X.; Funkhouser, T.; Guibas, L.; Hanrahan, P.; Huang, Q.; Li, Z.; Savarese, S.; Savva, M.; Song, S.; Su, H.;  et~al.
\newblock Shapenet: An information-rich 3d model repository.
\newblock {\em arXiv} {\bf 2015}, arXiv:1512.03012.

\bibitem[Goyal et~al.(2021)Goyal, Law, Liu, Newell and Deng]{RevisitingPointCloudShapeClassification}
Goyal, A.; Law, H.; Liu, B.; Newell, A.; Deng, J.
\newblock Revisiting point cloud shape classification with a simple and effective baseline.
\newblock In Proceedings of the International Conference on Machine Learning, 2021; pp. 3809--3820.

\bibitem[Zini and Awad(2022)]{ExplainabilityDeepModels}
Zini, J.E.; Awad, M.
\newblock On the explainability of natural language processing deep models.
\newblock {\em ACM Comput. Surv.} {\bf 2022}, {\em 55}, 103.

\bibitem[iMerit()]{ManagingUncertainity}
iMerit.
\newblock Managing Uncertainty in Multi-Sensor Fusion: Bayesian Approaches for Robust Object Detection and Localization.
\newblock
Available online:  \url{https://imerit.net/resources/blog/managing-uncertainty-in-multi-sensor-fusion-bayesian-approaches-for-robust-object-detection-and-localization/} (accessed on 1 October 2025).

\bibitem[Pfeifer et~al.(2016)Pfeifer, Weissig, Lange and Protzel]{SensorFusionApplication}
Pfeifer, T.; Weissig, P.; Lange, S.; Protzel, P.
\newblock Robust factor graph optimization, A comparison for sensor fusion applications.
\newblock In Proceedings of the 2016 IEEE 21st International Conference on Emerging Technologies and Factory Automation (ETFA), 2016; \mbox{pp. 1--4.}
\newblock {\url{https://doi.org/10.1109/ETFA.2016.7733598}}.

\bibitem[Xu et~al.(2021)Xu, Yang, Sun and Picek]{FactorGraphForINS}
Xu, J.; Yang, G.; Sun, Y.; Picek, S.
\newblock A Multi-Sensor Information Fusion Method Based on Factor Graph for Integrated Navigation System.
\newblock {\em IEEE Access} {\bf 2021}, {\em 9}, 12044--12054.
\newblock {\url{https://doi.org/10.1109/ACCESS.2021.3051715}}.

\bibitem[Ziv et~al.(2025)Ziv, Matzliach and Ben-Gal]{SensorFusionForTargetDetection}
Ziv, Y.; Matzliach, B.; Ben-Gal, I.
\newblock Sensor Fusion for Target Detection Using LLM-Based Transfer Learning Approach.
\newblock {\em Entropy} {\bf 2025}, {\em 27}, 928.
\newblock {\url{https://doi.org/10.3390/e27090928}}.

\bibitem[Dey et~al.(2025)Dey, Merugu and Kaveri]{UncertaintyAwareFusion}
Dey, P.; Merugu, S.; Kaveri, S.
\newblock Uncertainty-aware fusion: An ensemble framework for mitigating hallucinations in large language models.
\newblock In Proceedings of the Companion Proceedings of the ACM on Web Conference 2025, 2025; pp. 947--951.

\bibitem[KWEON et~al.(2025)KWEON, Jang, Kang and Yu]{UncertaintyQuantification}
Kweon, W.; Jang, S.; Kang, S.; Yu, H.
\newblock Uncertainty Quantification and Decomposition for {LLM}-based Recommendation.
\newblock In Proceedings of the ACM on Web Conference 2025, 2025.

\bibitem[Zhu et~al.(2024)Zhu, Fan and Weng]{AdvancementsInPointCloudDataAugmentation}
Zhu, Q.; Fan, L.; Weng, N.
\newblock Advancements in point cloud data augmentation for deep learning: A survey.
\newblock {\em Pattern Recognit.} {\bf 2024}, \textit{153}, 110532.

\bibitem[Cuskley et~al.(2024)Cuskley, Woods and Flaherty]{LimitationsLLMs1}
Cuskley, C.; Woods, R.; Flaherty, M.
\newblock The Limitations of Large Language Models for Understanding Human Language and Cognition.
\newblock {\em Open Mind} {\bf 2024}, {\em 8}, 1058--1083.
\newblock {\url{https://doi.org/10.1162/opmi_a_00160}}.

\bibitem[Mikhail Burtsev and Job()]{LimitationsLLMs2}
Burtsev, M.; Reeves, M.; Job, A.
\newblock The Working Limitations of Large Language Models.
\newblock
Available online: \url{https://sloanreview.mit.edu/article/the-working-limitations-of-large-language-models/} (accessed on 27 April 2025).

\bibitem[Cai and Rostami(2024)]{DynamicTransformer}
Cai, Y.; Rostami, M.
\newblock Dynamic transformer architecture for continual learning of multimodal tasks.
\newblock {\em arXiv} {\bf 2024}, arXiv:2401.15275.

\bibitem[Kumar et~al.(2020)Kumar, Parker and Naderian]{AdaptiveTransformersInRL}
Kumar, S.; Parker, J.; Naderian, P.
\newblock Adaptive transformers in RL.
\newblock {\em arXiv} {\bf 2020}, arXiv:2004.03761.

\end{thebibliography}
\end{document}